\documentclass[10pt,twocolumn,letterpaper]{article}

\usepackage[pagenumbers]{cvpr} % To force page numbers, e.g. for an arXiv version

\definecolor{cvprblue}{rgb}{0.21,0.49,0.74}
\usepackage[pagebackref,breaklinks,colorlinks,allcolors=cvprblue]{hyperref}
\NewDocumentCommand{\TODO}{o}{
    \IfNoValueTF{#1}
    {
        \textbf{\color{red}[TODO]}
    }
    {
        \textbf{\color{red}[TODO: #1]}
    }
}
\usepackage{xcolor}
\newcommand{\rgb}{\textcolor{red}{R}\textcolor{teal}{G}\textcolor{blue}{B}}
\newcommand{\hsv}{\textcolor{cyan}{H}\textcolor{red}{S}\textcolor{darkgray}{V}}
\newcommand{\rnd}{\textcolor{teal}{R}\textcolor{magenta}{N}\textcolor{violet}{D}}

\newcommand{\ourModel}{Icarus}

\usepackage[ruled]{algorithm2e} % For algorithms

\SetAlFnt{\small}
\SetAlCapFnt{\small}
\SetAlCapNameFnt{\small}
\SetAlCapHSkip{0pt}

\usepackage{booktabs} 

\usepackage{caption}
\usepackage{subcaption}
\usepackage{float}
\usepackage{esint} 
\usepackage{comment}
\usepackage{graphicx}
\usepackage{arydshln}
\usepackage{multirow}
\usepackage{rotating}
\usepackage{lscape}
\usepackage{float}
\usepackage{tikz}
\usetikzlibrary{
    arrows,
    arrows.meta,
    positioning,
    quotes
}
\usepackage{amsmath,mathtools}
\usepackage[inline]{enumitem}
\usepackage{placeins}
\usepackage[inkscapelatex=false]{svg}
\usepackage{draftwatermark}
\SetWatermarkText{}
\SetWatermarkScale{1} % Adjust size
\SetWatermarkFontSize{1cm} 
\SetWatermarkAngle{45} % Adjust angle
\SetWatermarkColor{red} % Adjust color

\DeclareRobustCommand{\rchi}{{\mathpalette\irchi\relax}}
\newcommand{\irchi}[2]{\raisebox{\depth}{$#1\chi$}} % inner command, used by \rchi

\usepackage{xr}
\usepackage{pgffor}
\usepackage{xstring}
\usepackage{etoolbox}

\begin{document}
\crefformat{footnote}{#2\footnotemark[#1]#3}

\title{Full Dynamic Range Sky-Modelling For Image Based Lighting}

%%%%%%%%% AUTHORS - PLEASE UPDATE
\author{Ian J.~Maquignaz \\
    Université Laval \\
    Québec, Québec \\
    Canada \\
    {\tt\small ian.maquignaz.1@ulaval.ca}
}

%%%%%%%%% TITLE & TEASER

% Teaser
\twocolumn[{%
\renewcommand\twocolumn[1][]{#1}%
\vspace{-1.5cm}
\maketitle

\begin{center}
    \centering
    \vspace{-1cm}
    \captionsetup{type=figure}
    \includegraphics[width=1.0\textwidth,keepaspectratio]{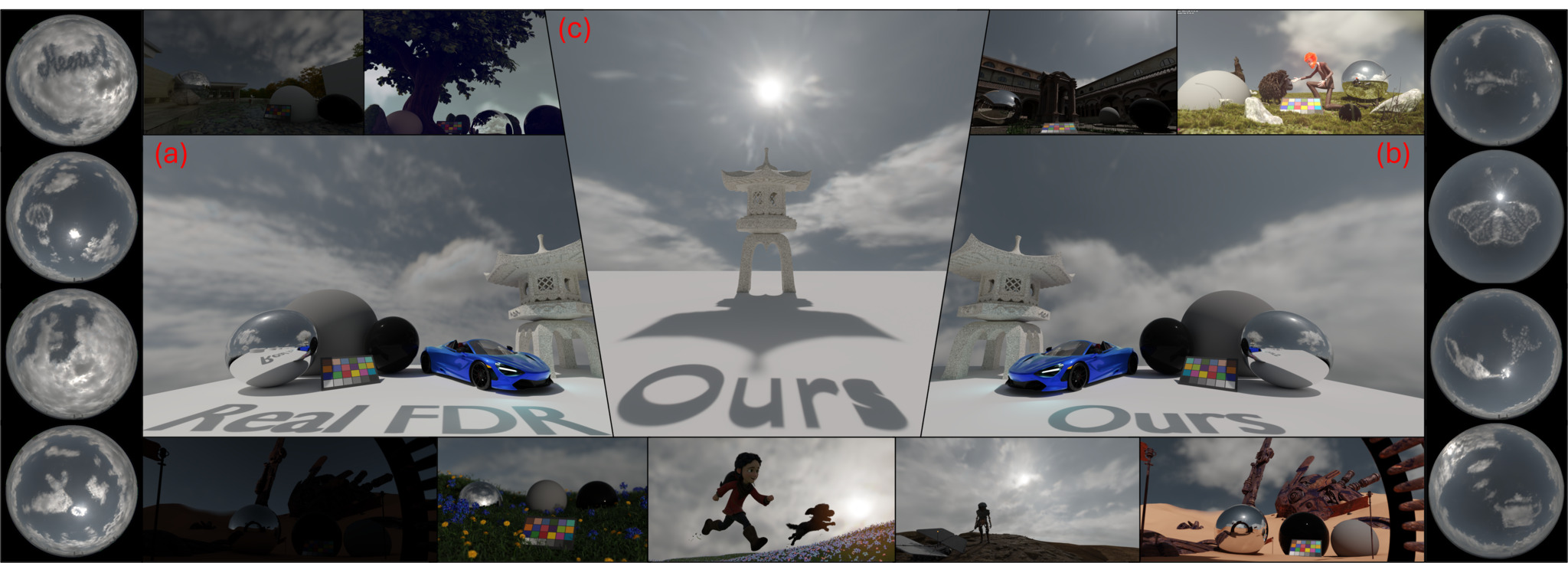}
    \captionof{figure}{
    Scenes rendered per $512^2$ environment maps.
    [Centre] Our DNN sky-model (Fig.1\textcolor{red}{b}, per \rgb-\ourModel{} with $f_{\text{\tiny Roberson}}$ fusion) recreates the illumination, tones and light transmission of real-world Full Dynamic Range imagery (Fig.1\textcolor{red}{a}, \textit{FDR} ground truth). \ourModel{} accurately models solar illumination for unprecedented lighting directionality (shadows, Fig.1\textcolor{red}{c}). [Border] \ourModel{} enables intuitive user-control over positioning and styling of solar and atmospheric formations.
    }
    \label{fig:teaser}
\end{center}%
}]

%%%%%%%%% ABSTRACT

\begin{abstract}
Accurate environment maps are a key component to modelling real-world outdoor scenes.
They enable captivating visual arts, immersive virtual reality and a wide range of scientific and engineering applications.
To alleviate the burden of physical-capture, physically-simulation and volumetric rendering, sky-models have been proposed as fast, flexible, and cost-saving alternatives.
In recent years, sky-models have been extended through deep learning to be more comprehensive and inclusive of cloud formations, but recent work has demonstrated these models fall short in faithfully recreating accurate and photorealistic natural skies.
Particularly at higher resolutions, DNN sky-models struggle to accurately model the 14EV+ class-imbalanced solar region, resulting in poor visual quality and scenes illuminated with skewed light transmission, shadows and tones.
In this work, we propose \ourModel{}, an all-weather sky-model capable of learning the exposure range of Full Dynamic Range (FDR) physically captured outdoor imagery.
Our model allows conditional generation of environment maps with intuitive user-positioning of solar and cloud formations, and extends on current state-of-the-art to enable user-controlled texturing of atmospheric formations.
Through our evaluation, we demonstrate \ourModel{} is interchangeable with FDR physically captured outdoor imagery or parametric sky-models, and illuminates scenes with unprecedented accuracy, photorealism, lighting directionality (shadows), and tones in Image Based Lightning (IBL).
\end{abstract}

%%%%%%%%% BODY
%%%%%%%%% BODY %%%%%%%%%
\section{Introduction}
\label{sec:introduction}

Modelling real-world illumination has been a long-standing area of study reflecting human perception of physical spaces and the visual quality of media and film \cite{lighting_lalonde24,lighting_lalonde21,Säks_2024}. 
Early works modelled illumination data combined from various sources as parametric sky-models to enable a wide range of engineering and scientific applications \cite{MOON_1940,PEREZ_1993}. 
With the advent of the digital age, parametric sky-models were extended to colour \cite{NISHITA_1993}, and subsequent numerical and parametric models
\cite{PEREZ_1993,NISHITA_1996,PREETHAM_1999,ONEAL_2005,HABER_2005,BRUNETON_2008,ELEK_2010,HOSEK_13,HOSEK_13Sun,LM_2014,PSM_21} generated clear and overcast skies with selective inclusion of a solar disk to enable Image-Based Lighting (IBL) techniques \cite{IBL} and facilitate the rendering of synthetic objects into real and virtual scenes.

Recent advancements have proposed all-encompassing DNN sky-models capable of generating weathered skies \cite{YANNICK_2019_SKYNET,SKYGAN_2022,DEEPCLOUDS_22,LM-GAN_2023,Ian_towardsSkyModels}, but such models offer variable performance in downstream applications.
Recent work has demonstrated that these model are not evaluated to the same standard as aforementioned numerical and parametric models, and commonly reported metrics including $L_1$, PSNR, HDR-VDP-3 inadequately quantify a sky-model's illumination and exposure range \cite{Ian_towardsSkyModels}. 
As a result, DNN sky-models often gravitate towards either accurate illumination with indiscernible atmospheric formations (e.g. clouds), or photorealistic atmospheric formation with poor illumination \cite{Ian_towardsSkyModels}.
Though a balance between photorealism and accurate illumination can be achieved, performance is dependent on image resolution, input-modalities, tone mapping operator(s), and the characteristics of the physically-captured dataset. 

Among other factors, the visual characteristics of the sky are a reflection of geographic/temporal locality, atmospheric aerosol/particle composition, and current weather systems \cite{BRUNETON_2017_clearSky_eval}. 
Though skies have been captured with a range of apparatus (e.g.\ spectroradiometers, pyranometers, and cameras \cite{8659184,KIDER_captureFramework,STUMPFEL_HDR_Sky_Capture}) and modelled in complex physical simulations and path tracers (libRadtran~\cite{libRadtran}, A.R.T \cite{ART}), no apparatus or model fully-encompasses the myriad of complex systems required to accurately and photorealistically recreate natural skies.
Current all-encompassing sky-models offer a generalization of weathered skies often insufficient for independent use, and proposed partial-sky-models which augment clear-skies to weathered environment maps \cite{DEEPCLOUDS_22,LM-GAN_2023,SKYGAN_2022} have not been demonstrated to offer improvement.
Alternatives to sky-models include labour- and computationally-intensive multi-step volumetric cloud rendering \cite{DeepScatering,aasberg2024real} and cloud simulations \cite{BRUNETON_2008_clouds,SIGGRAPH_course_2020}.
Though such alternative approaches have their merit (e.g.\ extraterrestrial worlds), physically captured skies offer a level of accuracy and photorealism which is often unsurpassed and preferable despite being labour-intensive, inflexible and of fixed temporal- and geo-locality  \cite{KIDER_captureFramework,STUMPFEL_HDR_Sky_Capture}. 

In this work, we propose \ourModel{}, the first all-encompassing DNN sky-model capable of generating photorealistic weathered environment maps with the full exposure range of natural outdoor illumination.
We achieve this by proposing a method for decomposing High Dynamic Range (HDR) imagery to Low Dynamic Range (LDR) brackets, enabling DNNs with our novel decoder architecture to model arbitrary exposure-ranges as brackets which can be merged (fused) to High Dynamic Range Imagery (HDRI) post generation. 
We demonstrate that LDR brackets can be accurately fused with our proposed DNN fusion model, or through various established methods from HDR literature per the desired representation.
Combined, our model \ourModel{} enables user-placement of atmospheric and solar formations and extends to offer user-configurability of cloud-textures through image-to-image style selection and stochastic style generation. 

\section{Background}
\label{sec:background}

\begin{figure}[tb]
    \centering
    \includegraphics[width=\linewidth,keepaspectratio]{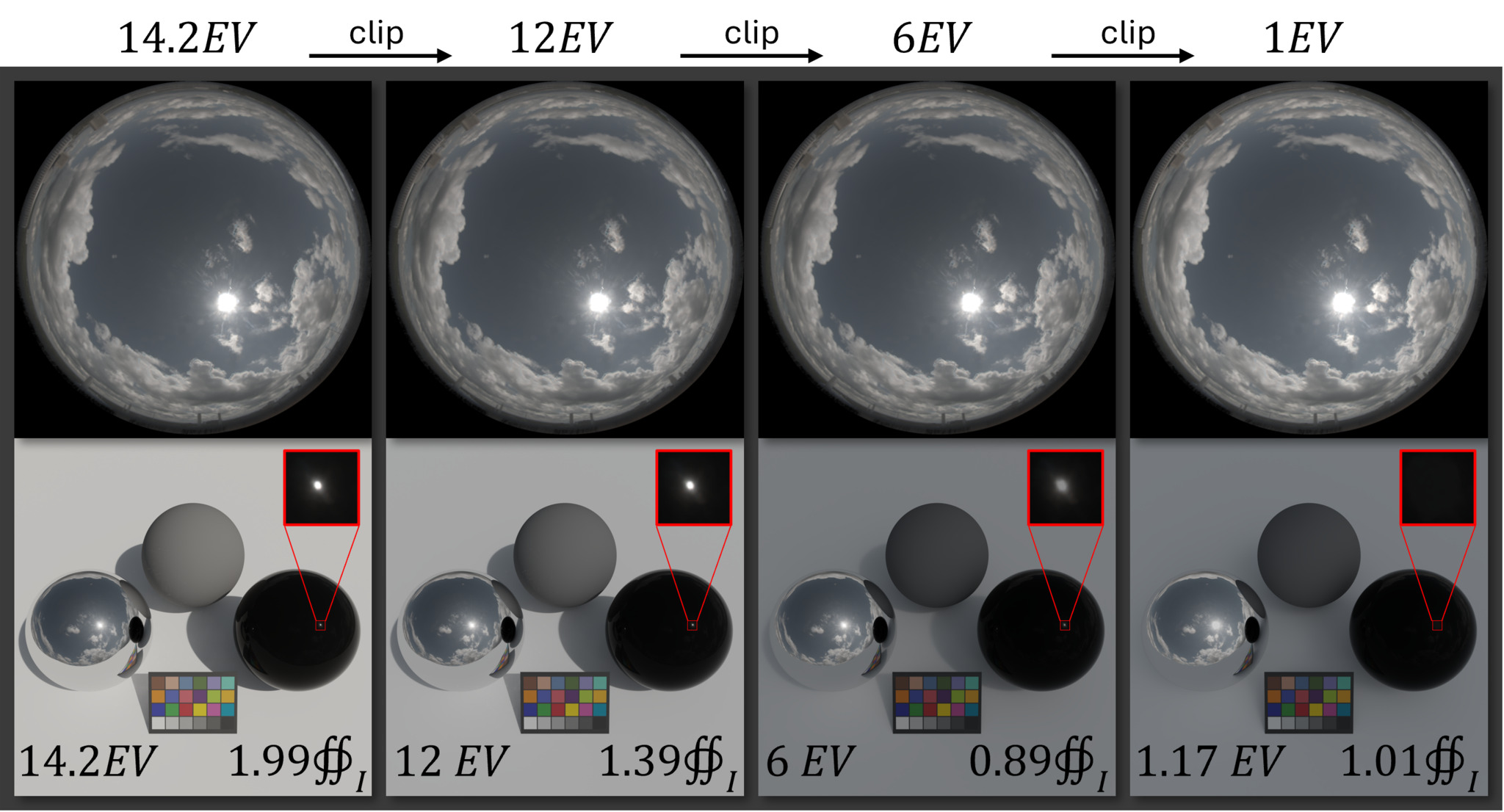}
    \caption{
    Each column illustrates the impact of an incremental clipping intensity with exposure equalization to the 14.2EV FDR ground truth.
    Though these \textit{partial}-capture environment maps are visually unaltered when clipped from FDR 14.2EV to LDR 1EV, renderings exhibit lost illuminance ($\oiint_I$), altering tones, shadows and light transmission (\textcolor{red}{sun transmission}; black glass orb).
    See \cref{sec:methodology} for definition of EV and $\oiint_i$.
    }
    \label{fig:grid_demo_EV}
\end{figure}

Recent work has demonstrated visually imperceptible inaccuracies in sky-model illumination can result in pronounced inaccuracies in downstream applications.
Though conventional LDR imagery is suitable for many scientific \cite{8659184} and rendering applications, HDR \cite{HDR_IBL_BOOK} imagery is integral to capturing the estimated $22$ f-stops of an average real-world outdoor scene.
Many conventional cameras today support HDR, but colloquial use of the term has rendered its definition dubious as a majority of HDRI only \textit{partially}-capture the exposure range of outdoor scenes.
As demonstrated by \cref{fig:grid_demo_EV}, incrementally clipping the exposure range of an HDRI to emulate partial-capture results in visually indiscernible alterations to the environment maps (\cref{fig:grid_demo_EV}, top), but pronounced alteration to illumination in IBL scenes through softer tones, shadows, and light transmission (\cref{fig:grid_demo_EV}, bottom).
Therefore, though all clipped environment maps in \cref{fig:grid_demo_EV} are by definition HDR, sky-modelling requires distinct HDRI which \textit{fully}-capture the exposure range of outdoor scenes to accurately model illumination. To distinguish between HDRI, we define the following:
\begin{enumerate}
    \item \textbf{Low Dynamic Range (LDR)} Imagery: Display-referenced imagery with compressed dynamic range which can be clipped and displayed in 8-bit colour (24-bit RGB) precision.

    \item \textbf{High Dynamic Range (HDR)} Imagery: Scene-referenced measures of illumination with uncompressed dynamic range and precision greater than LDR 8-bit colour (24-bit RGB) for later adaptation to LDR displays. This includes imagery captured by conventional cameras in 12-bit RAW.

    \item \textbf{Extended Dynamic Range (EDR)} Imagery: HDR images captured using techniques such as LDR bracketing for greater exposure range than a single image from a conventional camera.

    \item \textbf{Full Dynamic Range (FDR)} Imagery / Physically-Captured Imagery: HDR images that \textit{fully}-capture the dynamic range of a reference scene without truncation (saturation) of the exposure range.
\end{enumerate}
This distinction between HDRIs is necessary given the predominance of LDR/HDR/EDR sky datasets, and the limited applicability of HDR literature developed for conventional HDRI exposure ranges \cite{Zhang_2017_ICCV,HDR_from_LDR,nerfInTheDark,glowGAN}.
To \textit{fully}-capture the exposure range of an outdoor scene as FDR HDRI, specialized and labour-intensive physical-capture techniques are required \cite{JENSEN_NightSky,HDR_IBL_BOOK,STUMPFEL_HDR_Sky_Capture}.

\begin{figure*}[htb]
    \centering
    \begin{subfigure}{.48\linewidth}
        \centering
        \includegraphics[width=\linewidth,keepaspectratio=true]{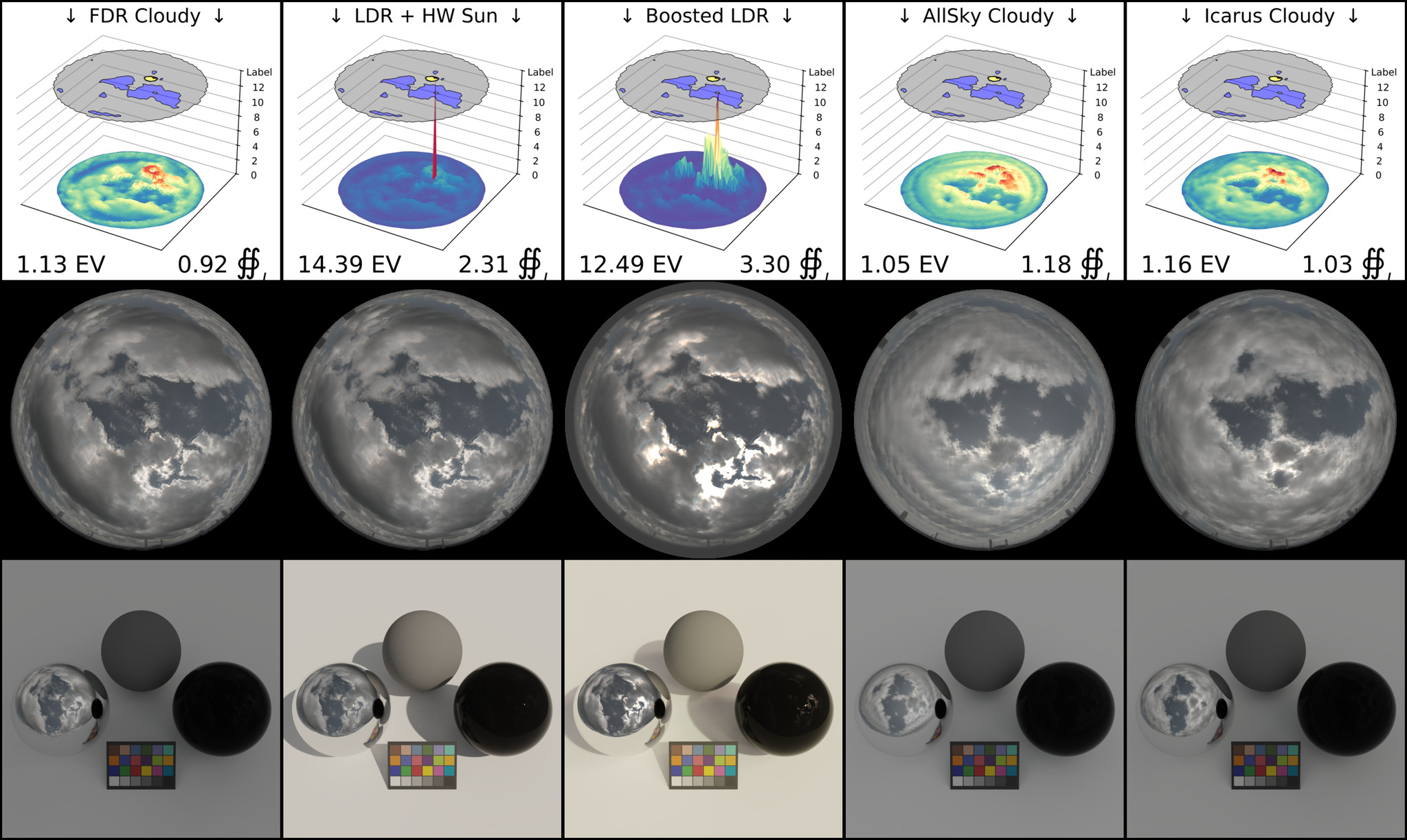}
        \caption{Cloudy sky with obfuscated sun}
        \label{fig::quickFix_cloudySky_ObfuscatedSun}
    \end{subfigure}
    \begin{subfigure}{.48\linewidth}
        \centering
        \includegraphics[width=\linewidth,keepaspectratio=true]{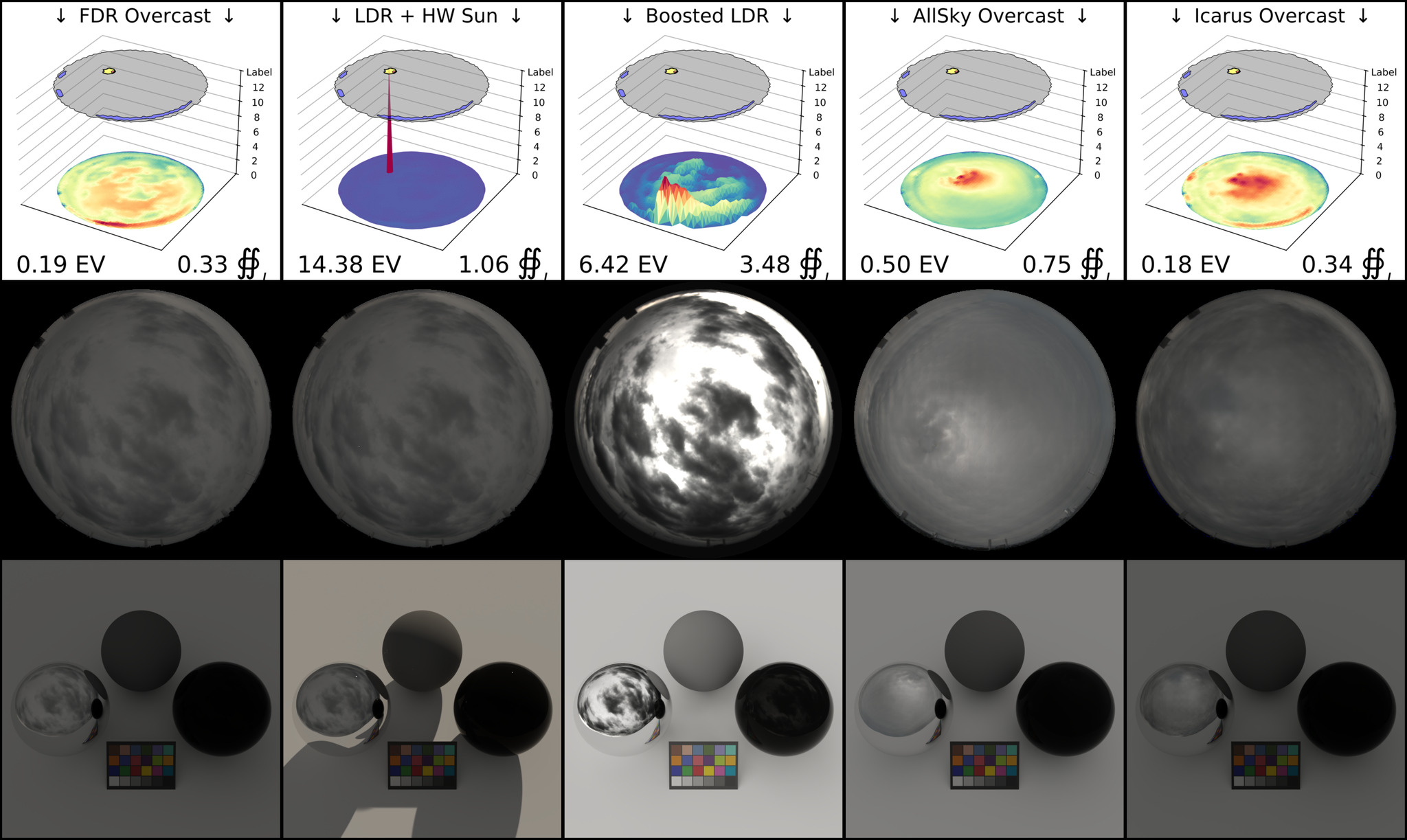}
        \caption{Overcast sky}
        \label{fig::quickFix_overcastSky}
    \end{subfigure}
    \caption{
        Comparison of AllSky and our model (\ourModel{}) to mitigation of solar modelling by substitution of a parametric Hošek-Wilkie sun (HW \cite{HOSEK_13Sun,DEEPCLOUDS_22}) and manual parametric boosting of the HDR environment map (Boosted \cite{text2light}; $\gamma$=0.5, $\beta$=2, $\rho$=6).
        Adding an HW sun (\textit{LDR+HW Sun}) skips atmospheric attenuation and allows the sun to `pierce' through clouds to create strong shadows.
        Boosting an HDR image (\textit{Boosted LDR}) is subject to weather-dependent parameter selection.
        If the sun is obstructed, boosting is prone to over-exposing and producing unpredictable shadows.
        Both mitigation strategies alter the perceived tones in IBL renderings (lambertian planar surface), but AllSky and our model (\ourModel{} with $f_{\text{\tiny Robertson}}$ fusion, right column) accurately model real-world FDR illumination for photorealistic IBL renderings.
        \cref{fn:mitagation_complete_grids}
    }
    \label{fig::quickFix_substitution_and_boosting}
\end{figure*}

To enable DNN sky-models to uniformly generate weathered skies and solar luminous intensity, previous works have proposed a range of tone mapping operators to compress HDRI to favourable visible or latent color-spaces \cite{YANNICK_2019_SKYNET,text2light,Ian_towardsSkyModels,SKYGAN_2022,DEEPCLOUDS_22}\footnote{Comparison of tone mapping operators in \cref{app:background::tonemapping}}.
Recent work has demonstrated that aggressive tone mapping operators such as $\mu\text{-lawLog}_2$ \cite{Ian_towardsSkyModels} are required to support solar exposure ranges but introduce a non-linear relationship between compressed LDR and uncompressed HDR spaces.
This results in exponentially large errors at high exposures and fluctuating illumination in generated environment maps.
Irrespective of tone mapping operator selection, the intrinsic class-imbalance of solar-pixels has been demonstrated to be increasingly prohibitive at higher resolution, with adversarial training heavily favouring solar features in  differentiation \cite{styleGAN3_2021,SKYGAN_2022,Ian_towardsSkyModels}.
Similarly, natural stochasticity in the textures of cloud formations and in the intensity of the solar disk have been shown to inhibit supervised training, where the application of $L_1$ and LPIPS \cite{LPIPS} on models with insufficient tractability for image-reconstruction results in smooth/blurred cloud formations and suppressed illumination \cite{YANNICK_2019_SKYNET,DEEPCLOUDS_22,Ian_towardsSkyModels}.

To mitigate modeling solar luminance, prior works have proposed the substitution of the solar disk with a parametrically generated disk \cite{DEEPCLOUDS_22}, manual parametric boosting \cite{text2light} and composite shading.
As demonstrated in \cref{fig::quickFix_substitution_and_boosting}, performance is varied with strategies exhibiting limited generalization to the four primary sky configurations:
clear, cloudy, cloudy with overcast sun and overcast skies \footnote{Complete grids for clear, cloudy, and obstructed sun skies in \cref{app:background::mitigation_strategies}\label{fn:mitagation_complete_grids}}.
Though results can be visually appealing, emulating atmospheric attenuation for reliable luminance in downstream applications remains a challenge.
Augmenting LDR environment maps to FDR via an Inverse Tone Mapping Operator (iTMO) Multilayer Perceptron (MLP) has been attempted by Text2Light \cite{text2light}, but visual quality is compromised \cite{Ian_towardsSkyModels}.

As a result, the modelling of outdoor FDR imagery remains a challenge without a clear solution \cite{Apple_Paper}.
Regardless of the approach, the current limitations result in many situations where physical capture is the preferred and unsurpassed method to provide both photorealism and weather variations \cite{HOSEK_13Extrasolar, FORZA}.

\section{Methodology}
\label{sec:methodology}

In this work, we propose a novel approach to DNN sky-modelling enabling the generation of high resolution environment maps with arbitrary exposure ranges and real-world atmospheric stochasticity. 
We quantize dynamic range in terms of the Exposure Value (EV) of an image $I_{c,ij}$ given $EV=\log_2\left(\max\left({\left|I\right|}_{ij}\right) - \min\left({\left|I\right|}_{ij}\right) + 1\right)$, where $|I|$ is grayscale intensity. 
This enables the comparison of environment map peak intensities (e.g. solar disks), providing insight into both luminous flux and luminous intensity (casting of shadows and diffused illumination) \cite{Ian_towardsSkyModels}.
We pair EV with Integrated Illumination ($\oiint_I$) to quantize luminous flux given an environment map's solid angles ($\Omega$, pixel-wise angular field-of-view) as: 

\vspace{-\baselineskip}
\begin{equation}
\oiint_I(I) = \sum{\Omega\odot|I|}
\label{eq:oiint_I}
\end{equation}
\vspace{-1.5\baselineskip}

This cumulative measure characterizes an environment map's illuminance of a scene, inclusive of the solar disk and the multiple scatterings within the atmosphere which respectively represent two thirds and one third of the sky's luminous flux \cite{BRUNETON_2017_clearSky_eval}.
As demonstrated in \cref{fig:grid_demo_EV}, these measures offer sensitivity to compare otherwise visually indiscernible differences between environment maps and are key to distinguishing HDR from our FDR environment maps. 

In this work, we propose a novel approach to DNN sky-modeling by first decomposing HDRI to LDR exposure brackets (bracketing) and thus alleviate the requirement for tone mapping operators. 
These LDR brackets enable unsupervised training of our proposed style-aware generator with novel decoder architecture and methodology for post generation fusion to FDR environment maps.  
Lastly, we define the novel discriminators which enable training and ensure the coherence of LDR exposure brackets by accounting for per-exposure feature variability and continuity between exposures.

\subsection{High Dynamic Range Imaging: Bracketing}
\label{sec:methodology::ldr_bracketing}

The predominant method to train HDR DNNs is to compress HDRI $\hat{I}$ using a bijective tone mapping operator $T_m$ to create displayable and/or training-favourable LDR imagery  $\Check{I}=T_m(\hat{I})$. 
Recent work has shown this approach to introduce a non-linear relationship between LDR- and HDR-spaces and to have a limited compatibility with class-imbalanced exposure ranges \cite{Ian_towardsSkyModels}. 
Given the exposure range required of sky-models (i.e. blue sky vs. solar disk), inverse-tone mapping $\hat{I}'=T_m^{-1}(\Check{I}')$ exacerbates small LDR errors to exponentially large HDR errors, and retention of the complete HDR spectrum skews training such that underrepresented classes are disproportionately ignored or overly emphasized \cite{Ian_towardsSkyModels}.

To overcome these limitations, we propose a novel bracketing approach which truncates class-imbalances and alleviates the requirement for tone mapping. 
We achieve this by defining a pseudo-inverse function to HDR fusion \cite{mertens2007exposure,HDR_IBL_BOOK} as $\left\{\Check{I}_{n} : f^{-1}\left(\hat{I}, \Delta{t}_n \right)\right\}^N$,  decomposing HDRI $\hat{I}_{c,ij}$ to brackets (sets) of $N$-LDR exposures $\left\{\Check{I}_{n,c,ij}\right\}^N$ with respect to exposure time $\Delta{t}_n$, a weighting $W_{n,c,ij}$ of exposures and optional tone mapper $T_m$:

\vspace*{-\baselineskip}
\begin{equation}
    \begin{aligned}
    \Check{I}_{n,c,ij} & = f^{-1}\left(\hat{I}_{c,ij}, \Delta{t}_n\right) \\
    & = \begin{dcases}
         0 & \text{if } T_{m}\left(\Delta{t}_n\hat{I}_{c,ij}\right) < \underline{\epsilon}  \\
         T_{m}\left(\Delta{t}_n\hat{I}_{c,ij}\right) & \text{otherwise} \\%\text{if } \epsilon \leq T_{m}(\hat{I}_{ij}/\Delta{t}_n) \leq \overline{\epsilon} \\
         0 & \text{if } \overline{\epsilon} < T_{m}\left(\Delta{t}_n\hat{I}_{c,ij}\right) \\
    \end{dcases}
    \end{aligned}
    \label{eq:HDR_decomposition}
\end{equation}
\vspace{-\baselineskip}
\begin{equation}
    \hat{I}_{n} = T_m^{-1}\left( \Check{I}_{n} \right) \Delta{t}^{-1}_n 
    \label{eq:normalize_ldr_exposure}
\end{equation}

\begin{figure}[htb]
\centering
\includegraphics[width=0.8\columnwidth,keepaspectratio]{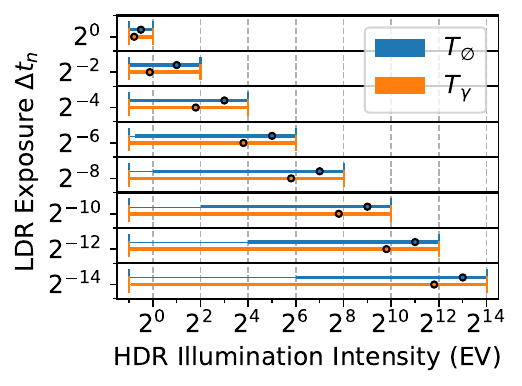}
\caption{
    Visual representation an LDR exposure bracket normalized to HDR-space by \cref{eq:normalize_ldr_exposure}. 
    Each LDR exposure is a `candle-stick' where
    upper- and lower-limits illustrate min/max HDR intensities for $\Check{I}$ clipped to $[0,1]$ and the body min/max HDR intensities for $\Check{I}$ clipped to $[\epsilon, 1-\epsilon]$ for $\epsilon=\underline{\epsilon}=\overline{\epsilon}=\frac{1}{255}$. 
    Markers ($\circ$) indicate the HDR illumination intensity of an LDR exposure value of 0.5. 
    Insufficient overlap and/or gaps between LDR exposures should be avoided during exposure selection.
}
\label{fig:diagram_LDR_bracket}
\end{figure}

Individual exposures can be visualized in HDR-space by partial-inversion of the decomposition as shown in \cref{eq:normalize_ldr_exposure}.
Tone mapping operator $T_{m}$ can be the identity operator $I = T_\varnothing(I)$, or a bijective tone mapping operator (e.g.\ $T_\gamma$). 
Though not explicitly required, as illustrated in \cref{fig:diagram_LDR_bracket} a tone mapping operator can reduce the number of exposures required for coverage of an exposure range. 
As shown by candle-stick lower shadow (tail between 0 and $\epsilon$), it may be necessary to overlap LDR exposures to compensate for poor granularity at low intensities (or high intensity when tone mapping).

\subsection{Model Architecture}

\begin{figure*}[htb]
  \centering
  \includesvg[width=\linewidth, height=\columnwidth,keepaspectratio]{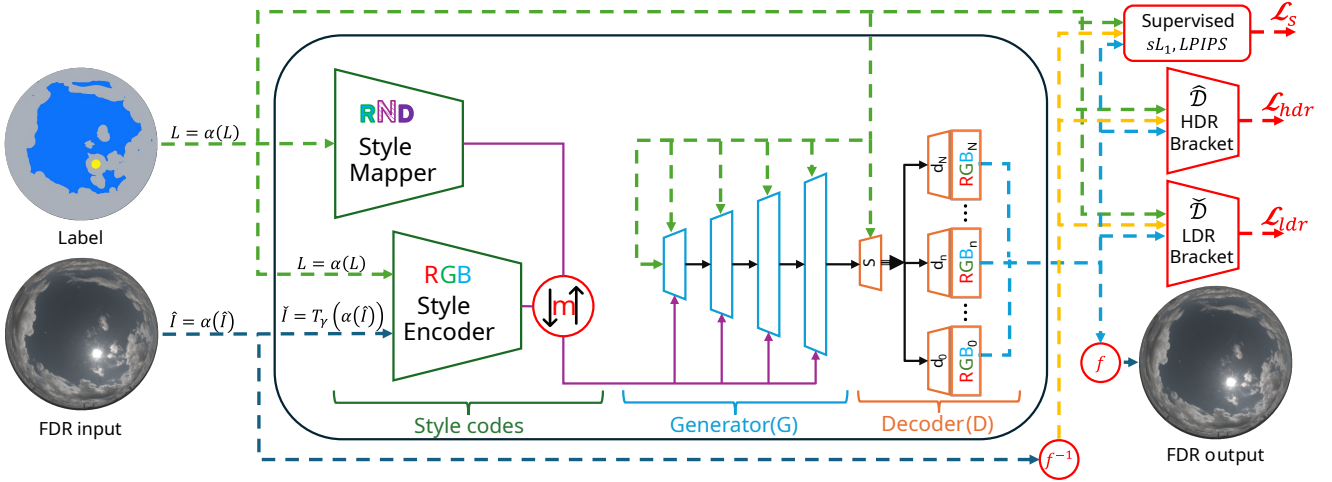}
    \caption{
    \ourModel{} training architecture. 
    Style-mixer (\textcolor{red}{m}) allows for selective training of the \rgb-style encoder and \rnd-style mapper. 
    The Generator (\textcolor{cyan}{G}) and Decoder (\textcolor{orange}{D}) develop an affinity to the selected style-code source.
    The decoder outputs an LDR bracket $\{\check{\mathcal{I}}_n\}^N$ which is evaluated per-exposure ($\check{\mathcal{I}}_n$) by the LDR-discriminator $\mathcal{\check{D}}$ to produce loss $\mathcal{L}_{ldr}$ and as bracket by the HDR-discriminator $\mathcal{\hat{D}}$ to produce loss $\mathcal{L}_{hdr}$.
    Supervised loss $\mathcal{L}_s$ amalgamates per-exposure class-selective losses ($sL_1$) applied to low-variability classes (i.e. border, skydome) and optional per-exposure LPIPS with \rgb-style reconstruction tasks. 
    }
\label{fig:dia_AllSky_SEAN}
\end{figure*}

With the objective of enabling both supervised and adversarial training, we propose \ourModel{} as illustrated in \cref{fig:dia_AllSky_SEAN}. Inspired from the style-aware generative pipeline of Semantic Region-Adaptive Normalization (SEAN, \cite{SEAN_2020}), \ourModel{} is all-encompassing sky-model with novel handling of style stochasticity, LDR brackets and fusion.

\subsubsection{Style Encoder \& Mapper}
 
In the absence of labelled datasets and tools to classify cloud-formations, current DNN sky-models unconditionally generate cloud formations with little diversity.
Where segmentation structure is deterministic of cloud formations, some models achieve a greater diversity by overfitting to input masks \cite{Ian_towardsSkyModels} but overall, the capacity for conditional or unconditional generation of all 27 categories of variably-textured and altitude-specific cloud formations \cite{NOAA_clouds} is undemonstrated. 
Similarly, from an extraterrestrial Point-Of-View (POV) the sun is a $0.5^{\circ}$ angular-diameter disk with near-constant luminance \cite{seeds_astronomy}, but from a terrestrial POV its size and luminance are dependent on attenuation by a stochastic weathered atmosphere and control over this luminance is also undemonstrated.

Through the set of style codes $\left\{\rchi_k\right\}^\mathcal{K}$ for classes $\mathcal{K}$, we propose a per-class latent representations of textures which can be distilled from a segmented image or stochastically produced.
This enables user-control over the texture and luminance of sky, clouds and solar formations without an explicitly labelled dataset. 
The \rgb-Style Encoder distills $\left\{\rchi_k\right\}^\mathcal{K}$ from an input image and segmentation label ($L$), providing the generator ($G$) deterministic tractability in supervised texture reconstruction. 
For versatility, images input to the encoder are Gamma ($T_{\gamma=2.2}$; clipped to $[0,1]$) tone mapped such that any LDR image can be used for style encoding.
Inspired from StyleGAN3's Mapping Network \cite{styleGAN3_2021}, the \rnd-Style Mapper generates style codes $\left\{\rchi_k\right\}^\mathcal{K}$ matching an input segmentation label, thus guiding style generation towards textures which will coherently integrate into the environment map's structure (e.g. mitigating the generation of incompatible clear-sky sun styles with overcast cloud styles).
Styles codes from either or both the \rgb-Style Encoder and \rnd-Style Mapper can be user-selected from one or multiple runs to generate a latent image $\xi=G\left(L,\left\{\rchi_k\right\}^\mathcal{K}\right)$.\footnote{See limitations in \cref{app:X_models::AllSky_SEAN}}

\subsubsection{Decoder}
\label{sec:methodology::model::decoder}

Previous works have focused on uniform generation of HDR environment maps, but recent work has shown significant limitations in generating imagery with class-imbalanced exposure ranges \cite{Ian_towardsSkyModels}. 
To overcome this challenge, we propose a novel decoder $D$ configurable to a desired exposure range through an arbitrary set of $N$-decoding heads, 
transforming the generator's latent image to a bracket of LDR imagery as $\left\{\check{\mathcal{I}}_n\right\}^N=D\left(\xi,L\right)$.

\begin{figure}[htb]
    \begin{subfigure}{.55\linewidth}
    \resizebox{\linewidth}{!}{%
        \begin{tikzpicture}[
    ->,>=stealth',
    shorten >=1pt,
    node distance=1.8cm, thick,
    main node/.style={circle,draw},
    every edge quotes/.style = {inner sep=3pt,
                            font=\footnotesize, sloped, anchor=center},
    pin distance=5mm,
    ]
    
    % Generator
    \node[main node] (S) [
    pin={[pin edge={->,thick,red,shorten <=1pt,transform canvas={xshift=0pt,yshift=2pt}}]left:$\xi$},
    pin={[pin edge={<-,thick,black,shorten <=1pt,transform canvas=  {xshift=0pt,yshift=-2pt}}]left:$\xi$}
    ] {$S$};
    % z-encoder + Mapper
    % \node[main node] (E) [above left of=S] {$E$}
    % \node[main node] (M) [below left of=S] {$M$};
    % Decoder
    \node[main node] (Dk) [right of=S,label={[yshift=5pt]$\vdots$}] {$d_n$};
    % \node[txt node] (Ck_1) [above of=Dk] {$\vdots$};
    % \node[txt node] (Ck_2) [below of=Dk] {$\vdots$};
    \node[main node] (d0) [above of=Dk] {$d_0$};
    \node[main node] (DN) [below of=Dk,label={[yshift=8pt]$\vdots$}] {$d_N$};
    % Fusion
    \node[main node] (F) [
    right of=Dk, 
    pin={[pin edge={<-,thick,red,shorten <=1pt,transform canvas={xshift=0pt,yshift=2pt}}]right:$\hat{I}$},
    pin={[pin edge={->,thick,black,shorten <=1pt,transform canvas=  {xshift=0pt,yshift=-2pt}}]right:$\hat{I}$}
    ] {$f$};

    % \node[main node, draw=none] (wtr) [
    % right of=Dk, 
    % pin={[pin edge={<-,thick,red,shorten <=1pt,transform canvas={xshift=0pt,yshift=2pt}}]right:$\hat{I}$},
    % pin={[pin edge={->,thick,black,shorten <=1pt,transform canvas=  {xshift=0pt,yshift=-2pt}}]right:$\hat{I}$}
    % ] {$f$};
    
    % Path Latent
    \path[every node/.style={
        font=\sffamily\small},transform canvas={xshift=2pt,yshift=-2pt}
    ] 
        (S) edge node [] {} (d0)
        % (M) edge node [] {} (S)
        % (DN) edge node [] {} (F)
    ;
    \path[every node/.style={
        font=\sffamily\small},transform canvas={xshift=2pt,yshift=2pt}
    ] 
        (S) edge node [] {} (DN)
        % (E) edge node [] {} (S)
        % (d0) edge node [] {} (F)
    ;    
    \path[every node/.style={
        font=\sffamily\small},transform canvas={yshift=2pt}
    ] 
        (S) edge node [] {} (Dk)
        % (Dk) edge node [] {} (F)
    ;
    % Path backpropagation
    \path[every node/.style={font=\sffamily\small},
        color=red, shorten <=1pt, transform canvas={xshift=-2pt,yshift=2pt}
    ] 
        (d0) edge node [] {} (S)
        % (S) edge node [] {} (M)
        (F) edge node [] {} (DN)
    ;
    \path[every node/.style={font=\sffamily\small},
        color=red,shorten <=1pt, transform canvas={xshift=-2pt,yshift=-2pt}
    ] 
        (DN) edge node [] {} (S)
        % (S) edge node [] {} (E)
        (F) edge node [] {} (d0)
    ;
    \path[every node/.style={font=\sffamily\small},
        color=red, shorten <=1pt, transform canvas={yshift=-2pt}
    ] 
        (Dk) edge node [] {} (S)
        (F) edge node [] {} (Dk)
    ;
    \draw [transform canvas={xshift=2pt,yshift=-2pt}]
        (DN) edge ["w.r.t $\Delta t_N$" below] (F)
    ;
    \draw [transform canvas={xshift=2pt,yshift=2pt}]
        (d0) edge ["w.r.t $\Delta t_0$" above] (F)
    ;
    \path [transform canvas={xshift=0pt,yshift=2pt}]
        (Dk) edge (F)
    ;
    \path [draw=none, transform canvas={xshift=-4pt,yshift=1pt}]
        (Dk)[draw=none] edge [draw=none,"{\tiny w.r.t $\Delta t_n$}" above] (F)[draw=none]
    ;
\end{tikzpicture}
    }
    \caption{\centering Fusion Decoder \\ {\scriptsize $\hat{\mathcal{I}}=f\left(\left\{\Check{\mathcal{I}}_{n} : d_n\left(S\left(\xi,L\right)\right), \Delta t_n\right\}^N\right)$}}
    \label{diag:decoder::wFusion}
    \end{subfigure}
    \begin{subfigure}{.4\linewidth}
    % \hspace{-25mm}
    \resizebox{1.1\linewidth}{!}{%
    \begin{tikzpicture}[
    ->,>=stealth',
    shorten >=1pt,auto,
    node distance=1.8cm,
    thick,
    main node/.style={circle,draw},
    pin distance=5mm,
    ]

    % Generator
    \node[main node] (S) [
        pin={[pin edge={->,thick,red,shorten <=1pt,transform canvas={xshift=0pt,yshift=2pt}}]left:$\xi$},
        pin={[pin edge={<-,thick,black,shorten <=1pt,transform canvas=  {xshift=0pt,yshift=-2pt}}]left:$\xi$}
    ] {$S$};
    % % z-encoder + Mapper
    % \node[main node, draw=none] (fake1) [
    %     left of=S,
    % ] {};
    % \node[main node, draw=none] (fake2) [
    %     left of=fake1,
    % ] {};
    % \node[main node] (M) [below left of=S] {$M$};
    % Decoder
    \node[main node] (dn) [
        right of=S, label={[yshift=5pt]$\vdots$}, 
        pin={[pin edge={<-,thick,red,shorten <=1pt,transform canvas={xshift=0pt,yshift=2pt}}]right:$\Check{I}_n$},
        pin={[pin edge={->,thick,black,shorten <=1pt,transform canvas=  {xshift=0pt,yshift=-2pt}}]right:$\Check{I}_n$}
    ] {$d_n$};
    \node[main node] (d0) [
        above of=dn, 
        pin={[pin edge={<-,thick,red,shorten <=1pt,transform canvas={xshift=0pt,yshift=2pt}}]right:$\Check{I}_0$},
        pin={[pin edge={->,thick,black,shorten <=1pt,transform canvas=  {xshift=0pt,yshift=-2pt}}]right:$\Check{I}_0$}
    ] {$d_0$};
    \node[main node] (dN) [
        below of=dn, label={[yshift=8pt]$\vdots$}, 
        pin={[pin edge={<-,thick,red,shorten <=1pt,transform canvas={xshift=0pt,yshift=2pt}}]right:$\Check{I}_N$},
        pin={[pin edge={->,thick,black,shorten <=1pt,transform canvas=  {xshift=0pt,yshift=-2pt}}]right:$\Check{I}_N$}
    ] {$d_N$};
    
    % Path Latent
    \path[every node/.style={
        font=\sffamily\small},transform canvas={xshift=2pt,yshift=-2pt}
    ] 
        (S) edge node [] {} (d0)
        % (M) edge node [] {} (S)
    ;
    \path[every node/.style={
        font=\sffamily\small},transform canvas={xshift=2pt,yshift=2pt}
    ] 
        (S) edge node [] {} (dN)
        % (E) edge node [] {} (S)
    ;    
    \path[every node/.style={
        font=\sffamily\small},transform canvas={yshift=2pt}
    ] (S) edge node [] {} (dn);
    % Path backpropagation
    \path[every node/.style={font=\sffamily\small},
        color=red, shorten <=1pt, transform canvas={xshift=-2pt,yshift=2pt}
    ] 
        (d0) edge node [] {} (S)
        % (S) edge node [] {} (M)
    ;
    \path[every node/.style={font=\sffamily\small},
        color=red,shorten <=1pt, transform canvas={xshift=-2pt,yshift=-2pt}
    ] 
        (dN) edge node [] {} (S)
        % (S) edge node [] {} (E)
    ;
    \path[every node/.style={font=\sffamily\small},
        color=red, shorten <=1pt, transform canvas={yshift=-2pt}
    ] (dn) edge node [] {} (S);
    \path[every node/.style={font=\sffamily\small},
        color=red, shorten <=1pt, transform canvas={yshift=-2pt}
    ] (dn) edge node [] {} (S);
    %\path[every node/.style={font=\sffamily\small},
    %    color=red, shorten <=1pt, transform canvas={yshift=-2pt}
    %] (S) edge node [] {} (-1cm,0)+(S);
\end{tikzpicture}
    }
    \caption{\centering LDR Bracket Decoder \\ {\scriptsize $\left\{\Check{\mathcal{I}}_{n} : d_n\left(S\left(\xi, L\right)\right)\right\}^N$}}
    \label{diag:decoder::ldr_bracket}
    \end{subfigure}
    \caption{Progression of generation ($\rightarrow$) and gradients (\textcolor{red}{$\leftarrow$}) through the decoder's SPADE ($S$, \cite{SPADE_2019}), $N$ convolutional decoding heads ($d_n$), and (optionally, \cref{diag:decoder::wFusion}) HDR fusion ($f$) modules. 
    }
    \label{diag:decoder}
\end{figure}
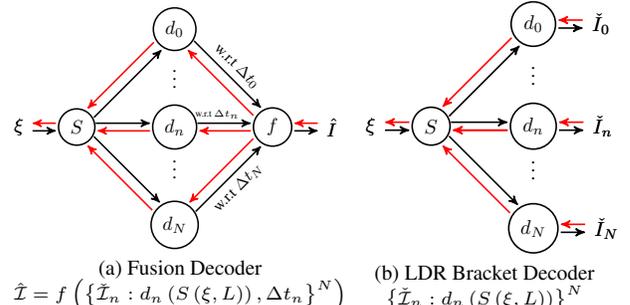

\vspace{-.5\baselineskip}
\begin{equation}
    \frac{\delta \mathcal{L}}{\delta S} = \sum_{n=1}^{N} \frac{\delta \Delta t_n}{\delta d_n}\cdot\frac{\delta f}{\delta \Delta t_n} \cdot \frac{\delta \mathcal{L}}{\delta f}
    \label{eq:decoder_fusion_gradient_at_S}
\end{equation}
\vspace{-.5\baselineskip}
\begin{equation}
    \frac{\delta \mathcal{L}}{\delta S} = \sum_{n=1}^{N} \frac{\delta \mathcal{L}_n}{\delta d_n}
    \label{eq:decoder_bracket_gradient_at_S}
\end{equation}

As illustrated in \cref{diag:decoder::wFusion}, the inclusion of a fusion ($f$) module requires modulation w.r.t exposure time ($\Delta t_n$) such that $\hat{\mathcal{I}}=D\left(\xi,L,\left\{\Delta t_n\right\}^N\right)$.
This implicitly modulates the gradient at SPADE \cite{SPADE_2019} module $S$ to be expressible as \cref{eq:decoder_fusion_gradient_at_S}, resulting in vanishing gradients at higher exposures.
To overcome this limitation, we propose a decoder which outputs an LDR bracket $\left\{\check{\mathcal{I}}_n\right\}^N=D\left(\xi,L\right)$, allowing the decoder to remain exposure agnostic and retain a simplified (non-vanishing) gradient at $S$ per \cref{eq:decoder_bracket_gradient_at_S}.

Where the set of exposures times $\left\{\Delta t_n\right\}^N$ is initialized such that $\Delta t_0\le2^0$ and $T_m\left(\Delta t_N\hat{I}\right)\le1$, distant exposures over wide exposure ranges can exhibit grossly different features. 
As such, a latent image $\xi$ learned from training $\check{\mathcal{I}}_a = d_a\left(S\left(G\left(\left\{\rchi_k\right\}^\mathcal{K},L\right),L\right)\right)$ for $\Delta t_a$ does not guarantee that head $d_b$ can be trained asynchronously by $\check{\mathcal{I}}_b = d_b\left(S\left(\xi,L\right)\right)$ for $\Delta t_b$ with a frozen generator $G$.
As training $d_b$ with an unfrozen generator and without $d_a$ can result in the generator `forgetting' features required by $d_a$, synchronous training of heads $\left\{d_n\right\}^N$ is required, but this is prone to collapse due to seemingly-uncorrelated features at different exposures. 

To mitigate and guide feature correlation-discovery, we propose synchronizing the exposure heads and iteratively decaying exposure times such that the model is incrementally exposed to higher-exposure features.
We achieve this by defining a function $\delta$ to decay from $\Delta t_0$ to $\Delta t_N$ over $i$ epochs, and initializing the set of $N$-exposures times $\left\{\Delta t_n\right\}^N$ to $\Delta t_0$ at epoch $E_0$. 
We then uniformly decay to the target exposures at epoch $E_i$ and continue training for subsequent epochs as $E_{i++}$:

\vspace*{-1.2\baselineskip}
\begin{equation}
\begin{aligned}
 E_0\text{: } \Delta t =& \left\{\Delta t_0,\hdots,\Delta t_n\!=\!\Delta t_0,\hdots,\Delta t_N\!=\!\Delta t_0\right\} \\
 E_1\text{: }\Delta t =& \left\{\Delta t_0,\hdots,\Delta t_n\!=\! \delta(1),\hdots,\Delta t_N\!=\!\delta(1)\right\} \\ &\text{ where } \delta(1) \leq \Delta t_0 \\
 E_{i-1}\text{: } \Delta t =& \left\{\Delta t_0,\hdots, \Delta t_n, \hdots,\Delta t_N\!=\!\delta(i-1)\right\} \\
 & \text{ where } \Delta t_N \leq \delta(i-1) \\  
 E_{i}\text{: } \Delta t =& \left\{\Delta t_0,\hdots,\Delta t_n, \hdots,\Delta t_N\right\} \\
  E_{i++}\text{: } \Delta t =& \left\{\Delta t_0,\hdots,\Delta t_n, \hdots,\Delta t_N\right\}
 \end{aligned} \label{eq:exp_decay}
\end{equation}
\vspace*{-1.5\baselineskip}

Though computationally more expensive than asynchronous training, iterative decay mitigates collapses and enables synchronous training of all decoder heads to fit a desired exposure range.

\subsubsection{LDR Bracket Fusion}
\label{sec:methodology::model::fusion}

With the assumption of a linear camera response function, merging (fusion) of LDR brackets to HDRI can be achieved through $f_{\text{\tiny\rgb}}$ in \cref{eq:LDR_exposure_fusion} and other proven methods from HDR literature including Debevec \cite{merge_Debevec} and Robertson \cite{merge_Robertson} fusion.

\vspace{-1.5\baselineskip}
\begin{equation}
\begin{aligned}
    \hat{I} &= f_{\text{\tiny\rgb}}\left(\left\{\Check{I}_{n},\Delta{t}_n,W_{n}\right\}^N\right) \\ 
    &= \sum_{n=1}^N \Delta{t}_n{\Check{I}_{n}} \odot \left( 1\bigg/\sum_{n=1}^N W_{n} \right)
    \end{aligned}
    \label{eq:LDR_exposure_fusion}
\end{equation}
\vspace{-.5\baselineskip}

Where $W_{n}$ is a mask of values between the lower ($\underline{\epsilon}$) and upper ($\overline{\epsilon}$) saturation points as defined by:

\vspace{-\baselineskip}
\begin{equation}
    W_{n,c,ij} = \begin{dcases}
        1 & \text{if } \underline{\epsilon} \leq \Check{I}_{n,c,ij} \leq \overline{\epsilon} \\
        0 & \text{otherwise}
    \end{dcases}
    \label{eq:LDR_saturation_points}
\end{equation}
\vspace{-\baselineskip}

\begin{figure}[htb]
  \centering
  \includesvg[width=\columnwidth,keepaspectratio]{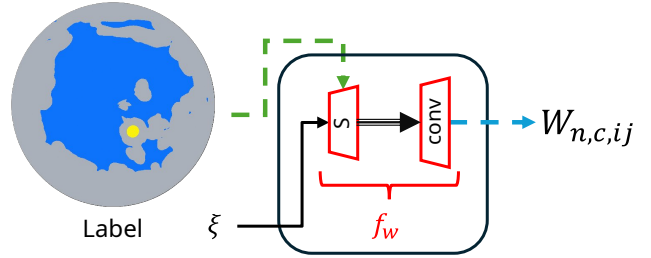}
    \caption{Fusion module $W_{n,c,ij}=f_w(\xi,L)$, learning bracket-adaptive weights $W_{n,c,ij}$ from latent $\xi$ and segmentation $L$.}
\label{fig:dia_AllSky_F}
\end{figure}

We propose separately training a DNN-fusion module to learn bracket-adaptive weights $W_{n,c,ij} = f_w(\xi, L)$ as shown in \cref{fig:dia_AllSky_F}.
As shown in \cref{eq::LDR_exposure_fusion_DNN}, adaptive weights $W_{n,c,ij}$ modulate exposures and exposure times respectively. 
Whereas Robertson fusion \cite{merge_Robertson} weights values per an assumed global Gaussian-like criteria, our approach weights values per localized segmentation- and instance- aware per-exposure criteria. 
% This enables favouring of lucid visual features from lower exposures and the pruning of channel-imbalanced noise from higher-exposures.

\vspace*{-1\baselineskip}
\begin{equation}
\begin{aligned}
    \hat{I} &= f_{\text{DNN}}\left(\left\{\Check{I}_{n},\Delta{t}_n,W_{n}\right\}^N\right) \\ 
    &= \sum_{n=1}^N \Delta{t}_{n} W_{n}\odot{\Check{I}_{n}} \odot \left( 1 \Bigg/ \sum_{n=1}^N {\Delta{t}_n^2} W_{n} \right)
\end{aligned}
    \label{eq::LDR_exposure_fusion_DNN}
\end{equation}
\vspace{-\baselineskip}

We note conventional implementations of fusion are intended for unsigned integer LDR imagery and truncating generated floating-point precision results in lost accuracy and visual quality. 
We overcome this limitation by defining our own implementation of Robertson fusion $f_{\text{\tiny Robertson}}$ in \cref{app:methodology::ldr_brackets:::Robertson} which supports floating-point precision. 

In addition, we note that conventional fusion methods can benefit from per-exposure class-selective masking of exposures $\check{I}_{n,c,ij}$ to a set of classes $M_{n} \subseteq \mathcal{K}$ per \cref{eq:class_selective_mask_per_exposure}. 
Though this can improve accuracy and visual quality, such masking is precarious given the exposure range of atmospheric formations can be dependent on solar intensity (sunrise, daytime, sunset) and thus should be avoided.

\vspace{-.5\baselineskip}
\begin{equation}
\check{I}^{'}_{n,c,ij} =
    \begin{dcases}
         \check{I}_{n,c,ij} & \text{if } {L}_{ij}  \in M_{n}  \\
         0 & \text{otherwise} 
    \end{dcases}
    \label{eq:class_selective_mask_per_exposure}
\end{equation}
\vspace{-\baselineskip}

\vspace{-\baselineskip}
\subsection{Losses}

For each exposure, we implement $L_1$-class-selective losses for border and skydome regions (e.g. blue sky)
Where supervised training is applicable, we optionally include LPIPS \cite{LPIPS} as a per-exposure reconstruction loss.
Adversarial training implements conditional n-Layer discriminator(s) inspired by \cite{Pix2Pix_2016,stablevideodiffusionscaling} with adaptive gradient penalty \cite{arjovsky2017wassersteingan, gulrajani2017improvedtrainingwassersteingans,kodali2017convergencestabilitygans}, adaptive weighting of the adversarial loss \cite{rombach2021highresolution} and a modified hinge loss with $\alpha=1$ and $\beta=0.5$\cite{lim2017geometricgan,Ian_towardsSkyModels}.
Border regions are masked from discriminator input.
For training the generator $\left\{\mathcal{I}_n\right\}^N = D(G(\left\{\rchi_n\right\}^N,L)$, we define an exposure-wise LDR Discriminator and bracket wise HDR Discriminator to guide learning of both individual exposures and ensure their coherence within the exposure bracket. 
We additionally implement a standard \rgb-discriminator for training fusion module $f_w$. 

\subsubsection{LDR Discriminator}
Given diminishing features for differentiation at higher exposures, we propose discriminating each exposure of an LDR bracket independently. 
Through implementation with per-exposure groupings of channels, the LDR discriminator $\check{\mathcal{D}}$ is equivalent to a set of $N$-discriminators such that $\check{\mathcal{D}} \equiv \left\{\check{\mathcal{D}}_n\right\}^N$. 
Given the model $\left\{ \check{\mathcal{I}}_{n}\right\}^N = D\left(G\left(L,\left\{\rchi_k\right\}^\mathcal{K}\right)\right)$, hinge-losses for the discriminator ($\mathcal{L}_{ldr D}$) and generator ($\mathcal{L}_{ldr G}$) are expressible as:  

\vspace{-1.5\baselineskip}
\begin{align}
& \begin{aligned}
    \mathcal{L}_{ldr D} =
    & \frac{1}{2}\sum^N_{n=1}
    \min\left(0,\alpha-\check{\mathcal{D}}_n\left(\check{I}_n,L\right)\right) \\
    & + \frac{1}{2}\sum^N_{n=1}\min\left(0,\alpha+\check{\mathcal{D}}_n\left(\check{\mathcal{I}}_{n},L\right)\right)
\end{aligned} \label{eq:loss_adv_LDR_D} \\ 
& \begin{aligned}
    \mathcal{L}_{ldr G} = 
    % & w_1 \min\left(1,\beta-\check{\mathcal{D}}_1\left(\check{\mathcal{I}}_{1},L\right)\right) \\
    & \frac{1}{N} \sum^N_{n=1}w_n \min\left(0,\beta-\check{\mathcal{D}}_n\left(\check{\mathcal{I}}_{n},L\right)\right)
\end{aligned} 
\label{eq:loss_adv_LDR_G}
\end{align}
\vspace{-\baselineskip}

The LDR discriminator provides a per-exposure loss with weighted bias $w_n$, emphasizing the imperative-to-visual-quality first exposure.  
The weighting coefficient $w_n$ also enables the emulation of exposure decay via \cref{eq:exp_decay} for asynchronous training of either the \rgb-style encoder or \rnd-style mapper with frozen $G$ and $D$.

\subsubsection{HDR Discriminator}
Given the LDR Discriminator guides per-exposure training, we propose an HDR discriminator $\hat{\mathcal{D}}$ to examine the entirety of LDR brackets and ensure continuity between exposures. 
% Examining the entirety of LDR brackets, $\hat{\mathcal{D}}$  mitigates chromatic and other incongruities between exposures.
Given a generator $\{ \check{\mathcal{I}}_{n}\}^N = D(G(L,\rchi))$, the modified hinge-loss can be expressed as:  

\vspace{-1.5\baselineskip}
\begin{align}
& \begin{aligned}
    \mathcal{L}_{hdr D} =
    & \frac{1}{2}
    \min\left(0,\alpha-\hat{\mathcal{D}}\left(\left\{\check{I}_n\right\}^N,L\right)\right) \\
    & + \frac{1}{2}\min\left(0,\alpha+\hat{\mathcal{D}}\left(\left\{\check{\mathcal{I}}_n\right\}^N,L\right)\right)
\end{aligned} \label{eq:loss_adv_HDR_D} \\ 
& \begin{aligned}
    \mathcal{L}_{hdr G} = 
    & w \min\left(0,\beta-\hat{\mathcal{D}}\left(\{\check{\mathcal{I}}_n\}^N,L\right)\right)
\end{aligned} \label{eq:loss_adv_HDR_G}
\end{align}
\vspace{-1\baselineskip}

The HDR discriminator is fractional in size compared to the LDR discriminator, but in its absence a bracket's exposures can exhibit de-synchronization of features. 
This is reflected in fused FDR imagery by artifacts including blurred features and `chromatic aberration'. 

\section{Experimental Configuration}
\label{sec:experiment_config}

For the purpose of comparison, all models were trained against the Laval HDR Sky database (HDRDB \cite{LavalHDRdb}, \cref{app:dataset}) and 
all images are gamma ($T_\gamma; \gamma\!=\!2.2$) tone-mapped.
Illustrations target the four primary skydome configurations: clear, cloudy with unobstructed sun, cloudy with obstructed sun, and overcast skies.

\begin{table*}[htb]
\centering
\caption{
    Comparative of AllSky and \ourModel{} with \rgb- and \rnd-style at $256^2$ and $512^2$ resolution. As shown, \ourModel{} demonstrates the capacity for improved visual quality (FID, MiFID, HDR-VDP3) and accurate illumination (EV, $\oiint_I$, $PL_\Omega$).
    \textbf{Bold} values indicate category best.
    $^\dagger$ \ourModel{} with exposures $\Delta t\!=\!\{2^0,2^{-8}, 2^{-15}\}$ %[1,256,32768]
    $^\ddagger$ \ourModel{} with exposures $\Delta t\!=\!\{2^0,2^{-8}, 2^{-13}, 2^{-14}, 2^{-15}\}$ %[1,256,8192,16384,32768]
}
\label{tab::results}
\begin{tabular}{r|c|ccc|ccccc}
    \multicolumn{2}{c|}{} & \multicolumn{3}{|c|}{\textbf{$T_{\gamma}$-cLDR}} & \multicolumn{5}{|c}{\textbf{HDR}} \\
    \cline{3-10} 
     & Method & LPIPS $\downarrow$ & FID $\downarrow$  & MiFID $\downarrow$ & HDR-VDP3 $\uparrow$ & VIF $\uparrow$ & EV $\leftarrow$ & $\oiint_I \leftarrow$ & $PL_\Omega \leftarrow$ \\ 
    
\toprule

Ground Truth $256^2$ & - & - & - & - & - & - & 8.54 & 1.21 & 0.481 \\
\hdashline
 AllSky & $T_{\mu\log_2}$ & 0.16 & 16.7 & 304 & 8.15 & 0.88 & \textbf{8.58} & 2.01 & 1.268 \\

\hdashline
\ourModel{} \rnd-style$^\dagger$ & $f_{\text{\tiny\rgb}}$ & - & 20.1 & 340 & \textbf{8.21} & \textbf{1.59} & 9.16 & \textbf{1.28} & \textbf{0.425} \\ % MGD 7
\ourModel{} \rnd-style$^\dagger$ & $f_{\text{\tiny Robertson}}$ & - & \textbf{11.0} & \textbf{204} & 8.16 & 1.52 & 9.08 & \textbf{1.28} & \textbf{0.425} \\ % MGD 7; epoch 799
 
\midrule
Ground Truth $512^2$ & - & - & - & - & - & - & 9.44 & 1.22 & 0.209 \\

\hdashline
 AllSky                   &  $T_{\mu\log_2}$ & 0.18 & 24.7 & 411 & 8.15 & \textbf{0.82} & 9.99 & 1.58 & 0.539 \\
\hdashline
% \hdashline
\ourModel{} \rgb-style$^\ddagger$ & $f_{\text{\tiny\rgb}}$ & 0.16 & 27.1 & 460 & \textbf{8.55} & 0.64 & 9.59 & 1.29 & \textbf{0.212} \\ %40
\ourModel{} \rgb-style$^\ddagger$ & $f_{\text{\tiny Robertson}}$ & 0.16 & \textbf{15.1} & \textbf{294} & 8.41 & 0.42 & \textbf{9.38} & \textbf{1.23} & 0.203 \\ %40
\ourModel{} \rgb-style$^\ddagger$ & $f_{\text{\tiny DNN}}$ & 0.16 & 15.3 & 298 & 8.53 & 0.29 & 9.25 & 1.16 & 0.160 \\ %40 F8

\bottomrule
\end{tabular}
\end{table*}

\subsection{HDRI Pre- and Post-processing}

We reproduce the `hand-drawn' segmentation proposed by AllSky \cite{Ian_towardsSkyModels}, expanding on previously identified challenges with environment map format conversions.
As HDRDB provides FDR HDRI in latlong format, models require HDRI in sky-angular format for skydome continuity \cite{Ian_towardsSkyModels}, and IBL rendering in Blender \cite{blender} currently supports spherical or equirectangular (latlong) environment maps, format conversion are inevitable. 
To mitigate instability of HDRI characteristics (see \cref{app:sec::pre_post_processing}), we implement format conversions at $\geq 2\times$ target resolution (inter-area upsampling if required), before inter-area downsampling (average pooling) to a target resolution.  
Note, in order to achieve a real-world solar disk of $0.5^\circ$ angular diameter \cite{seeds_astronomy}, a sky-angular resolution of $512^2$ or greater is required.

\subsection{Models}
We configure \ourModel{} with a fixed architecture of two middle layers and four upsampling layers, reporting independently trained \rgb- and \rnd-style variants. 
We compare to the current state-of-the-art model AllSky \cite{Ian_towardsSkyModels}, retraining per author specifications with our distribution and image processing of HDRDB. 
\footnote{Please see \cref{app:X_models,app:X_experiments::SEAN} for tertiary baselines and experimentation. Our implementation of SEAN \cite{SEAN_2020} trains 50\% faster with 25\% fewer parameters.
}

\subsection{Metrics}
 
To quantize visual quality, we report VIF \cite{VIF,HDR_metric_evaluation}, HDR-VDP3 \cite{HDRVDP3}, FID \cite{FID}, and MiFID \cite{MiFID}. 
To quantize illuminance, we report environment map exposure range (EV) and Integrated Illumination ($\oiint_I$), and propose additionally reporting environment map Peak-Luminance ($PL_\Omega$) as:  

\vspace{-.5\baselineskip}
\begin{equation}
    PL_\Omega(I) = \max\left(\Omega\odot|I|\right)
    \label{eq:PeakRandiantIntensity}
\end{equation} 

Where $\Omega$ is the environment map's solid angles (pixel-wise angular field-of-view). 
$PL_\Omega$ extends EV w.r.t directionality, characterizing peak image intensity as luminance and thus with correlation to luminous flux ($\oiint_I$).

Visual metrics are computed in Gamma ($T_{\gamma=2.2}$; clipped to $[0,1]$) LDR space and illuminance in uncompressed HDR space. 
We do not report $L_1$, $L_2$, and PSNR as recent work has shown these reconstruction metrics penalize natural stochasticity in cloud formations and solar intensity \cite{Ian_towardsSkyModels}.

\section{Experimental Results}
\label{sec:experiment_results}

\begin{figure}[htb]
  \centering
  \includesvg[width=\columnwidth,keepaspectratio]{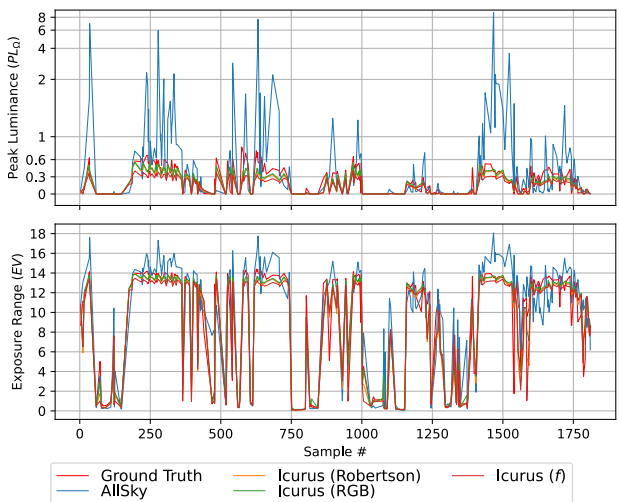}
    \caption{
    Comparison of model illumination ($\oiint_I$) and Peak-Luminance ($PL_\Omega$) across approximately 2k sequential samples. A shown, \ourModel{} $512^2$ with \rgb-styles offers stable and accurate real-world illumination from sunrise-to-sunset. \cref{fn:app_stable_illumination} }
\label{fig:results}
\end{figure}

As shown in \cref{tab::results}, \ourModel{} offers the capacity for improved visual quality and accuracy in illumination, with similar results between $f_{\text{\tiny DNN}}$ and $f_{\text{\tiny Robertson}}$ fusion. 
This demonstrates \ourModel{}'s architecture mitigates class-imbalances to produce both lucid atmospheric formations and real-world illumination and post-generation fusion's mitigation of vanishing gradients at increasing exposure.
Though $f_{\text{\tiny DNN}}$ and $f_{\text{\tiny Robertson}}$ are offer similar performance, the distinction from the visual quality of $f_{\text{\tiny \rgb}}$ highlights the importance of global and/or local per-exposure filtering to remove noise.

Where \cref{tab::results} reports average performance, \cref{fig:results} illustrates per-sample illumination across approximately 2k sequential samples to highlight stability in model performance.\footnote{See \cref{app:X_experiments::Fusion} for an extended comparison of sequential samples \label{fn:app_stable_illumination}}
As shown, AllSky exhibits high variability in peak luminance and exposure, stemming in part from its aggressive $T_{\mu\log_2}$ tone mapping. 
In this regard, overexposure by AllSky from 14EV to 16EV (a 400\% increase in HDR linear-space intensity) stems from a 5\% error in $T_{\mu\log_2}$-space while, in contrast, overexposure by 88\% in $T_{\gamma}$-space of \ourModel{}'s $2^{-15}$ exposure would be required to produce the same error.
\cref{fig:results} demonstrates our novel decomposition of FDR imagery to LDR brackets mitigates the requirement for aggressive tone mapping and stabilizes illumination. 

Visual results from \ourModel{} are included in \cref{fig:teaser,fig::quickFix_substitution_and_boosting}, reflecting stable performance with real-world illumination, shadows, tones and light transmission.
Additional illustrations of \ourModel's capacities for style-transferring \& mixing, cloud editing, artistic creations can be found in \cref{app:figure_pages}.

\section{Discussion}
\label{sec:discussion}

In this work, we propose a novel approach to sky-modelling which overcomes many of the limitation exhibited by current state-of-the-art DNN sky-model models. 
Where previous works have struggled with aggressive tone mappers, clear-sky augmentation, and solar spectrum mitigation techniques, we demonstrate an all-encompassing DNN architecture capable of photorealistically and accurately modeling FDR physically captured outdoor imagery. 

As \ourModel{} enables future sky-models to achieve higher versatility, photorealism and resolution with more advanced generative pipelines, the limitations of existing datasets will become more apparent. 
Few datasets of FDR sky imagery are available to meet the requirements for greater resolution, coverage of geo- and temporal-localities and atmospherics (particles, aerosols, and formations).
In this regard, HDRDB is limited to one geo-locality and a resolution of $512^2$ which, particularly at higher resolutions, is plagued by HDR artifacts including ghosting.
As a side effect of our work, we note that \rgb-style codes attempt to reproduce HDR artifacts while \rnd-Style codes can learn to ignore them.
This may provide future work with an opportunity to extend the life of existing datasets such as HDRDB.

\section{Conclusions}
\label{sec:conclusions}

In this work, we propose \ourModel{}, an all-encompassing sky-model capable of learning the exposure range of Full Dynamic Range (FDR) physically captured outdoor imagery.
We achieve this by solving the key intrinsic problems inhibiting DNN sky-models from achieving higher resolutions and exposure ranges, demonstrating \ourModel{} to provide photorealistic and accurate illumination, lighting directionality (shadows), light transmission, and tones.
Our approach offers intuitive user-control of textures and luminance, generating environment maps suitable for Image Based Lightning (IBL) applications and interchangeable with physically captured HDRI or parametric sky-models.

%%%%%%%%% BIBLIOGRAPHY
{
    \small
    \bibliographystyle{ieeenat_fullname}
    \bibliography{main}
}

%%%%%%%%% APPENDIX
\clearpage
\appendix
\raggedbottom
\clearpage
\section{Figure Pages}
\label{app:figure_pages}

\begin{figure*}[htbp]
    \centering
    \includegraphics[height=\textheight,width=\linewidth,keepaspectratio]{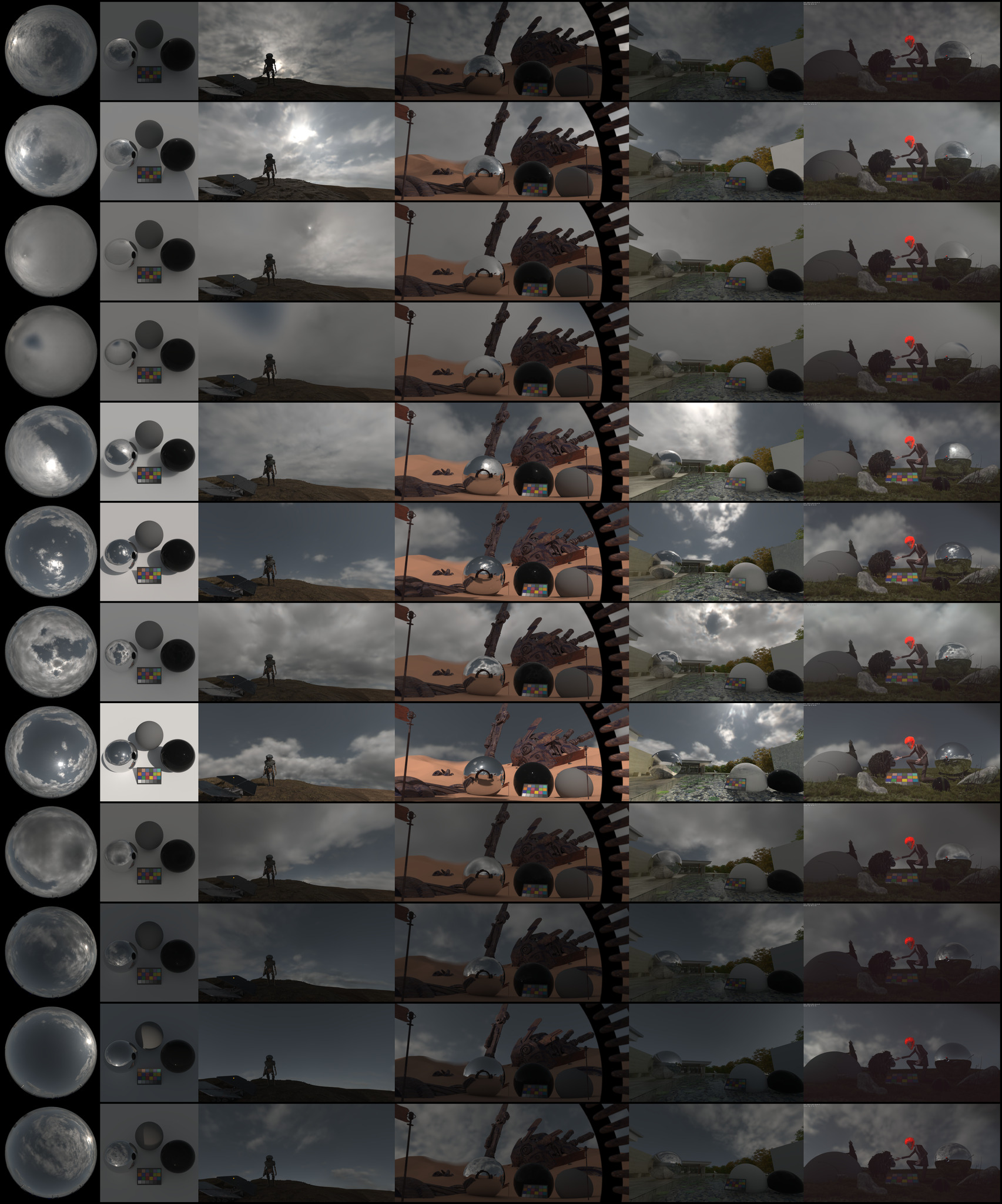}
    \caption{Collection of environment maps generated by \ourModel{} with \rgb-style encoder and $f_{\text{\tiny Robertson}}$, demonstrating IBL rendering performance from sunrise to sunset. }
    \label{app:fig::grid_AllSky_SEAN_ZGD_512}
\end{figure*}

\begin{figure*}[htbp]
    \centering
    \includegraphics[height=\textheight,width=\linewidth,keepaspectratio]{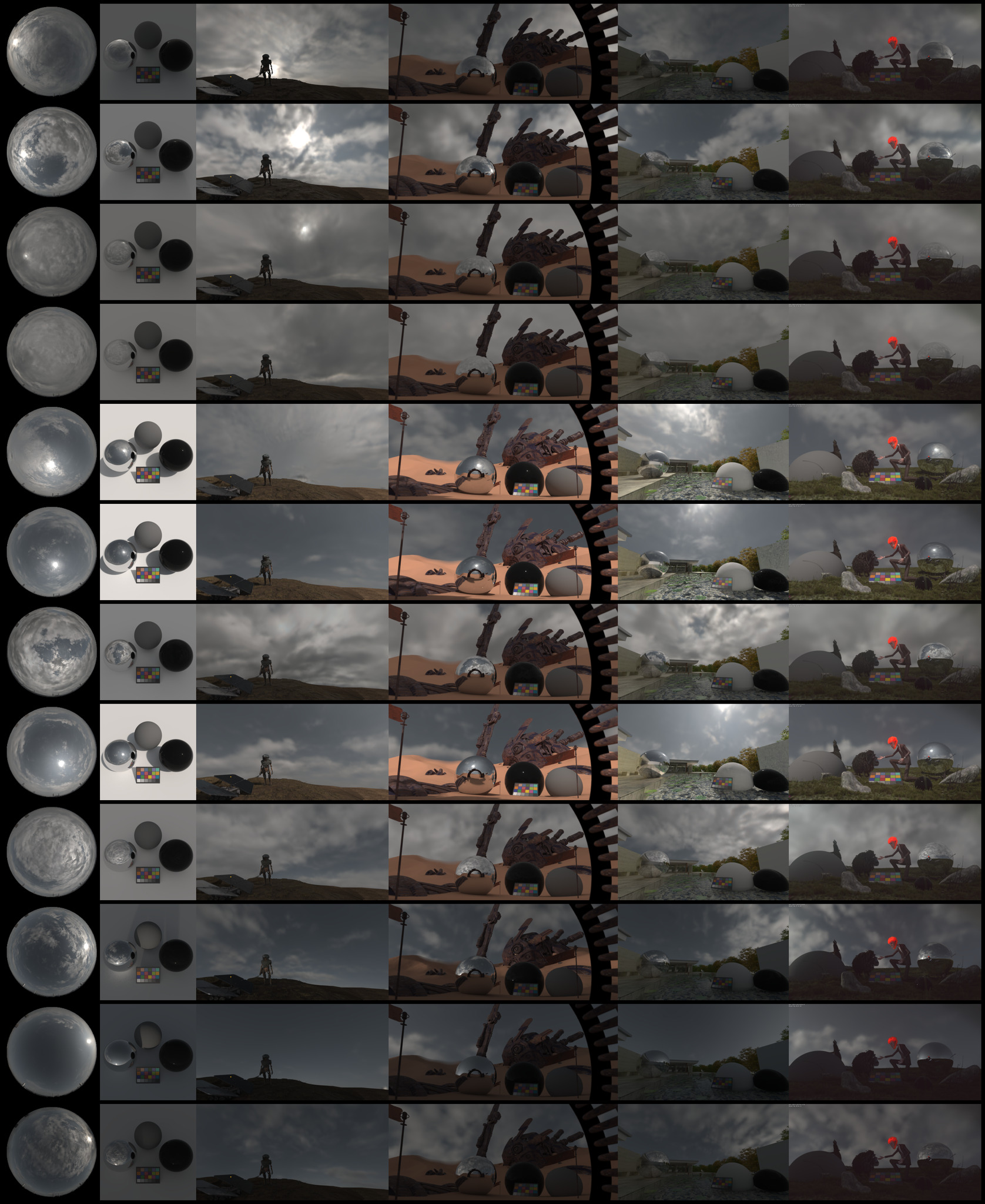}
    \caption{Collection of environment maps generated by \ourModel{} with \rnd-style mapper and $f_{\text{\tiny Robertson}}$, demonstrating IBL rendering performance from sunrise to sunset. }
    \label{app:fig::grid_AllSky_SEAN_MGD_256}
\end{figure*}

\begin{figure*}[htbp]
    \centering
    \includegraphics[height=\textheight,width=\linewidth,keepaspectratio]{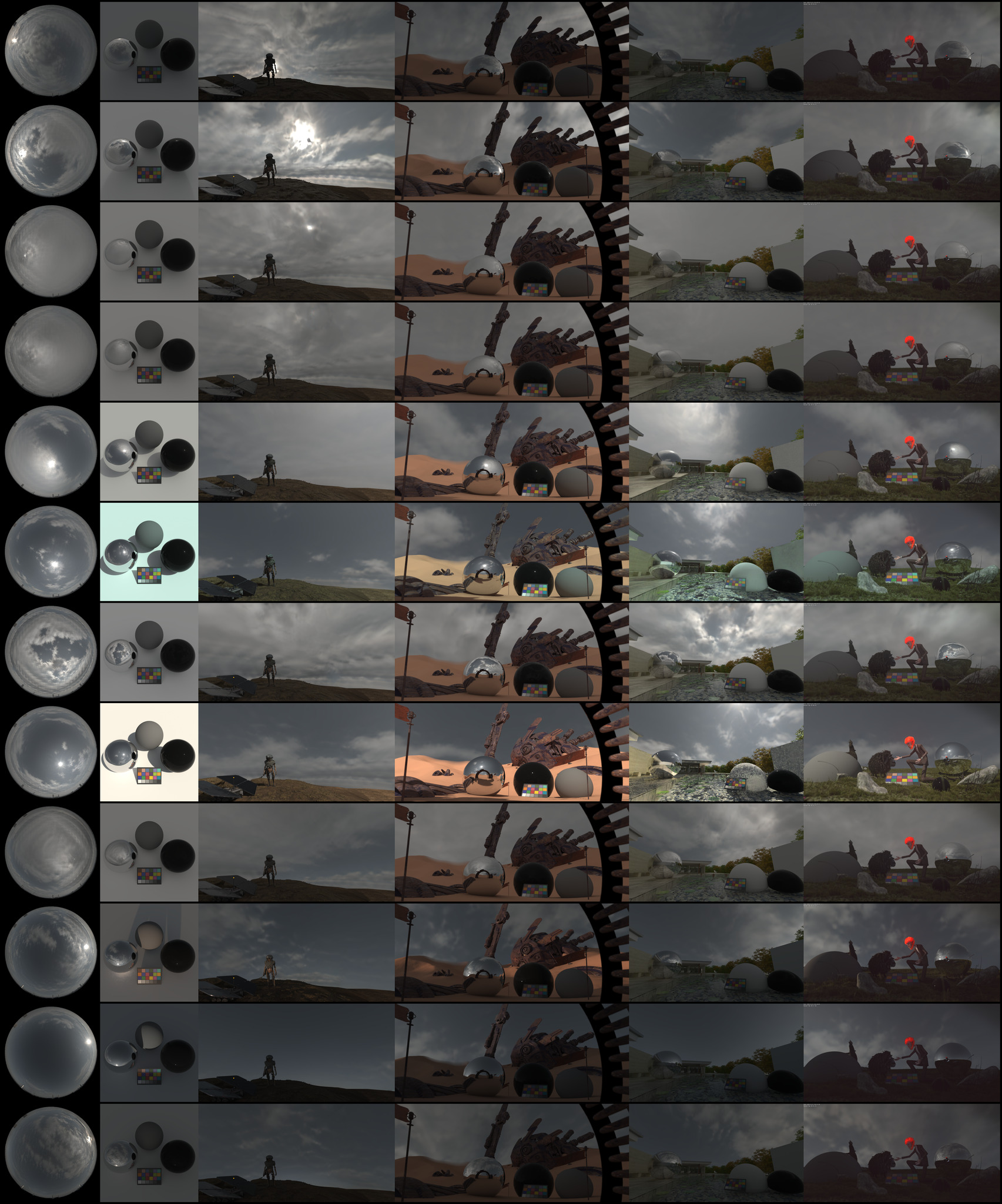}
    \caption{Collection of environment maps generated by AllSky \cite{Ian_towardsSkyModels}, demonstrating IBL rendering performance from sunrise to sunset. }
    \label{app:fig::grid_AllSky_FixUpUnet_512}
\end{figure*}

\begin{figure*}[htbp]
    \centering
    \includegraphics[height=\textheight,width=\linewidth,keepaspectratio]{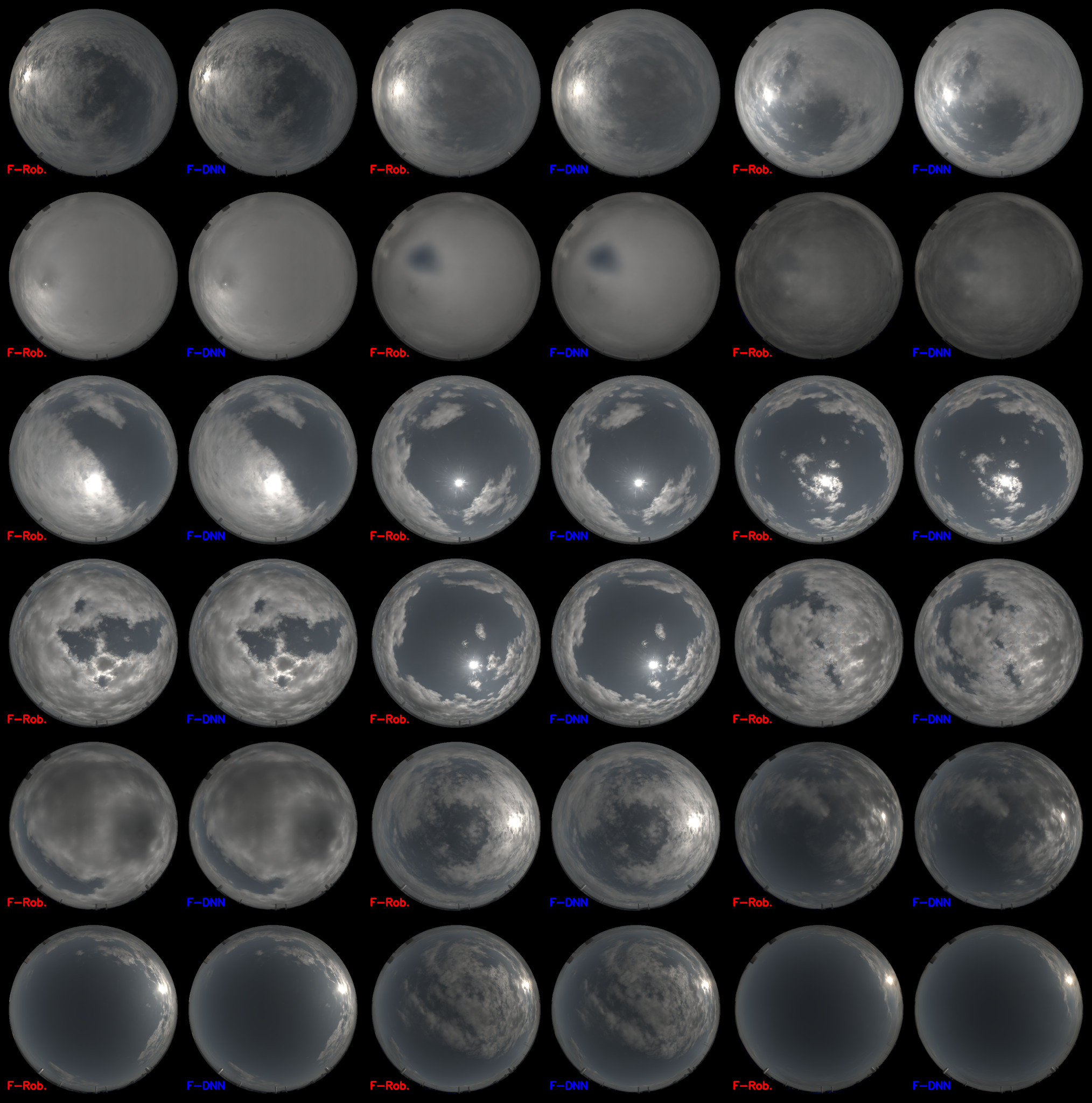}
    \caption{Side-by-side comparison of $f_{\text{\tiny Robertson}}$ and $f_{\text{\tiny DNN}}$, demonstrating their equivalence in terms of visual quality. }
    \label{app:fig::grid_AllSky_SEAN_Fdnn_vs_Frob}
\end{figure*}

The figures in the following pages offer deeper insight into our \ourModel{}.
Please note, figures which are continuations/extension of topics from the main body can be found in respective continuation sections of the Appendix.
For the extended comparison of mitigation strategies, AllSky and \ourModel{}, please see \cref{app:background::mitigation_strategies}.

\begin{landscape}
\begin{figure}[p]
    \centering    \includesvg[height=\textwidth,width=\linewidth,keepaspectratio]{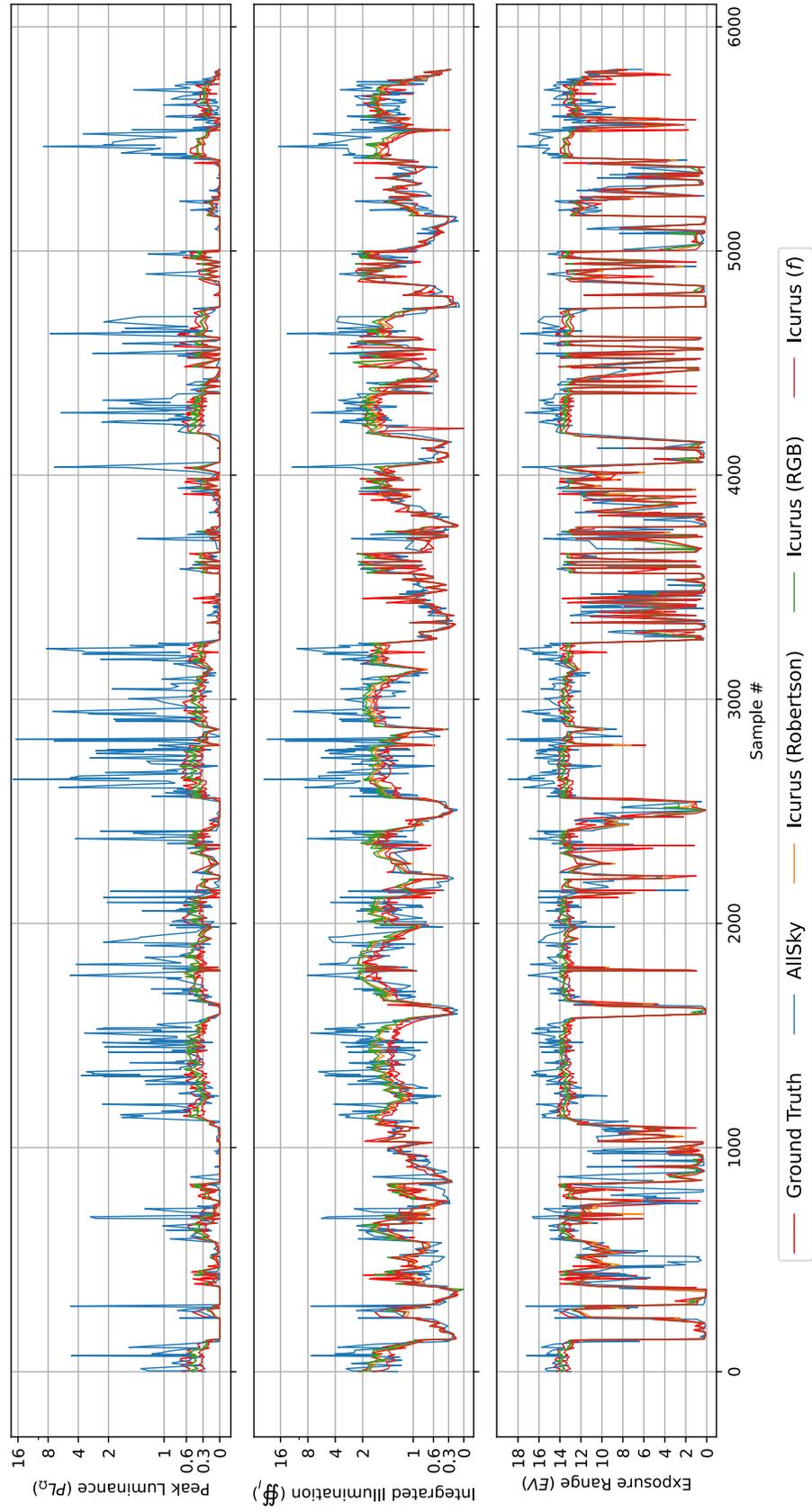}
    \caption{Comparison of model illumination ($\oiint_I$, EV) and Peak-Luminance ($PL_\Omega$) across approximately 6k sequential samples. A shown, \ourModel{} $512^2$ with \rgb-styles offers stable and accurate real-world illumination from sunrise-to-sunset.}
    \label{app:fig::results_sequential_samples}
\end{figure}
\end{landscape}

\begin{landscape}
\begin{figure}[p]
    \centering
    \includegraphics[height=\textwidth,width=\linewidth,keepaspectratio]{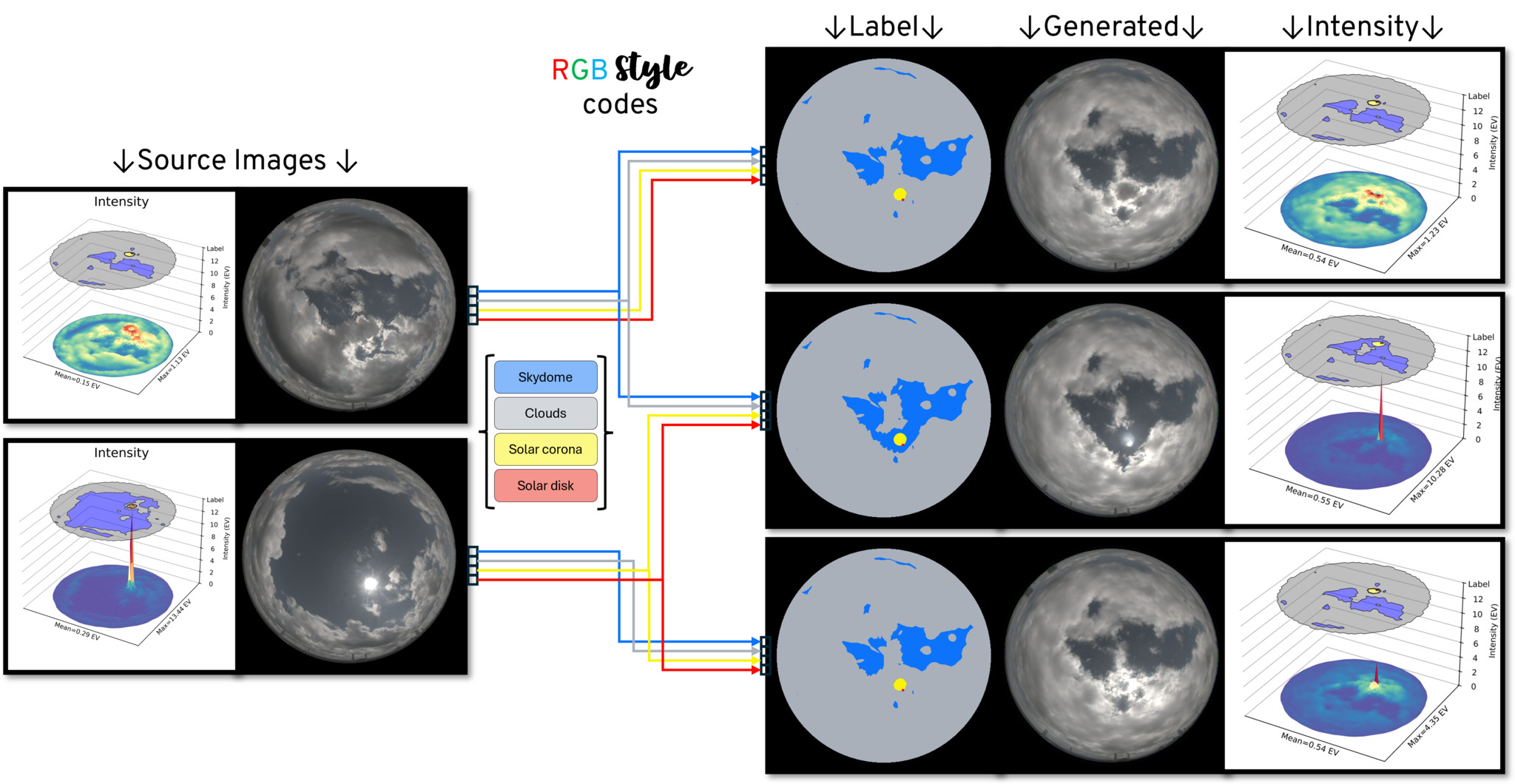}
    \caption{Illustration of \ourModel{}'s ability to transfer solar illumination with \rgb-styles. A shown, an obfuscated sun (top row) can be un-obfuscated (middle row) or brightened (bottom row) by transferring the style of solar corona and solar disk from a preferable source image.}
    \label{app:fig::demo_1_style_transfer_sun}
\end{figure}
\end{landscape}

\begin{landscape}
\begin{figure}[p]
    \centering
    \includegraphics[height=\textwidth,width=\linewidth,keepaspectratio]{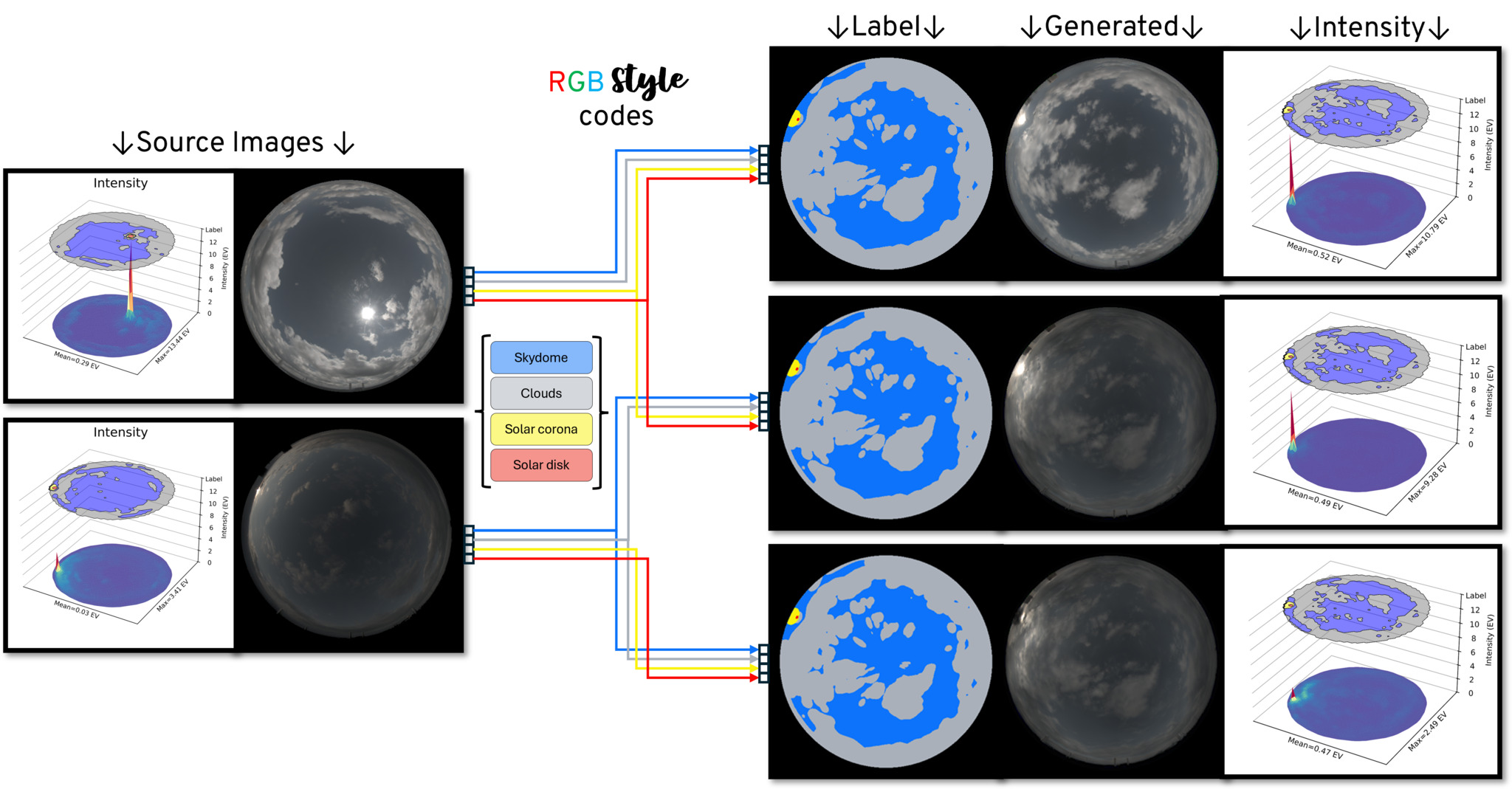}
    \caption{Illustration of \ourModel{}'s ability to transfer solar illumination. A shown, a sunset (bottom row) can be brightened (middle row) or given a daytime-like appearance (top row) by transferring the styles of a daytime source image. \ourModel{} $512^2$ with \rgb-styles and $f_{\text{\tiny Roberson}}$ fusion.}
    \label{app:fig::demo_2B_style_transfer_sun}
\end{figure}
\end{landscape}

\begin{landscape}
\begin{figure}[p]
    \centering
    \includegraphics[height=\textwidth,width=\linewidth,keepaspectratio]{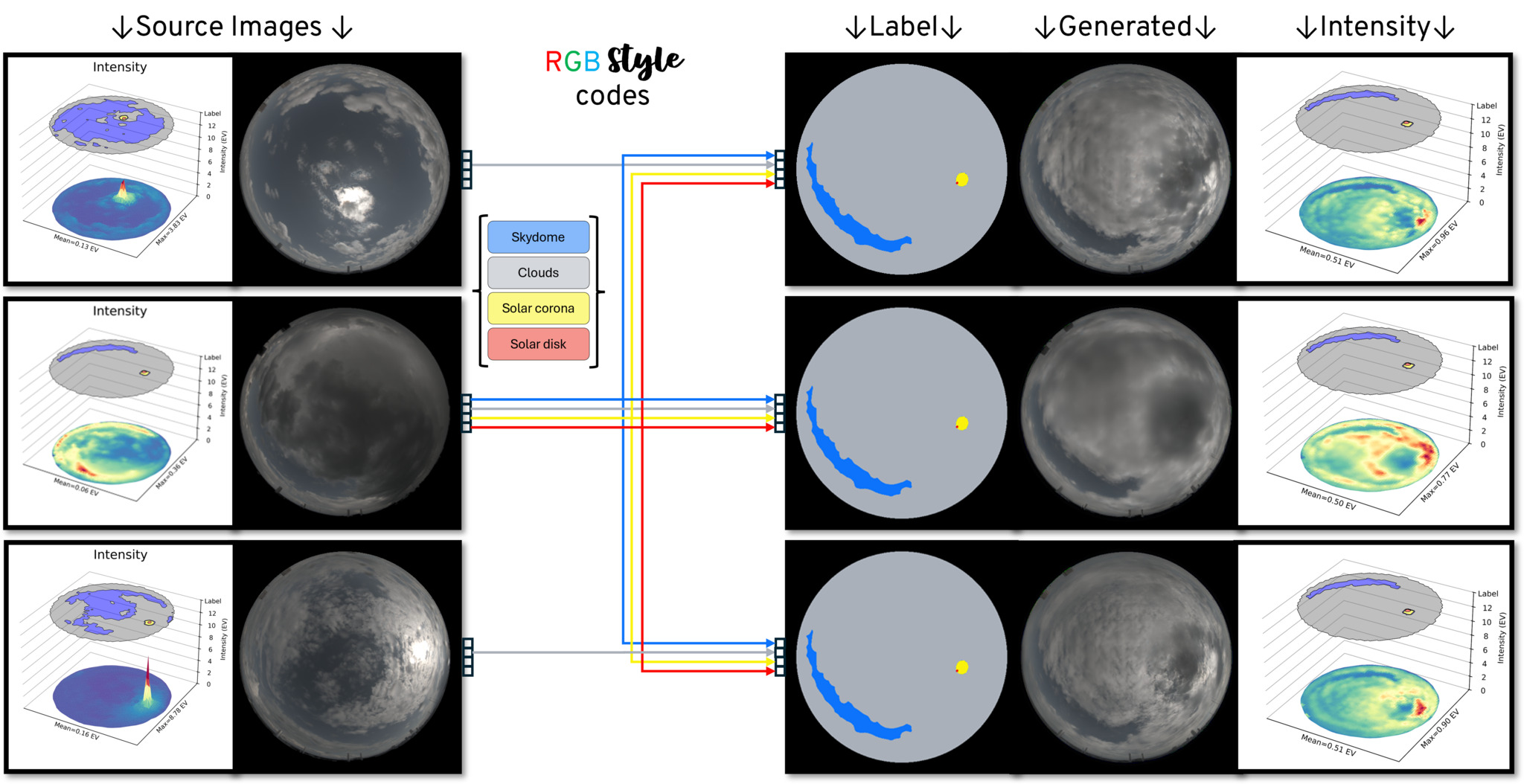}
    \caption{Illustration of \ourModel{}'s ability to transfer cloud formation. A shown, the gloom of dark clouds (middle row) can be alleviated (top and bottom rows) by transferring styles from preferable source images. \ourModel{} $512^2$ with \rgb-styles and $f_{\text{\tiny Roberson}}$ fusion.}
    \label{app:fig::demo_3_style_transfer_clouds}
\end{figure}
\end{landscape}

\begin{landscape}
\begin{figure}[p]
    \centering
    \includegraphics[height=\textwidth,width=\linewidth,keepaspectratio]{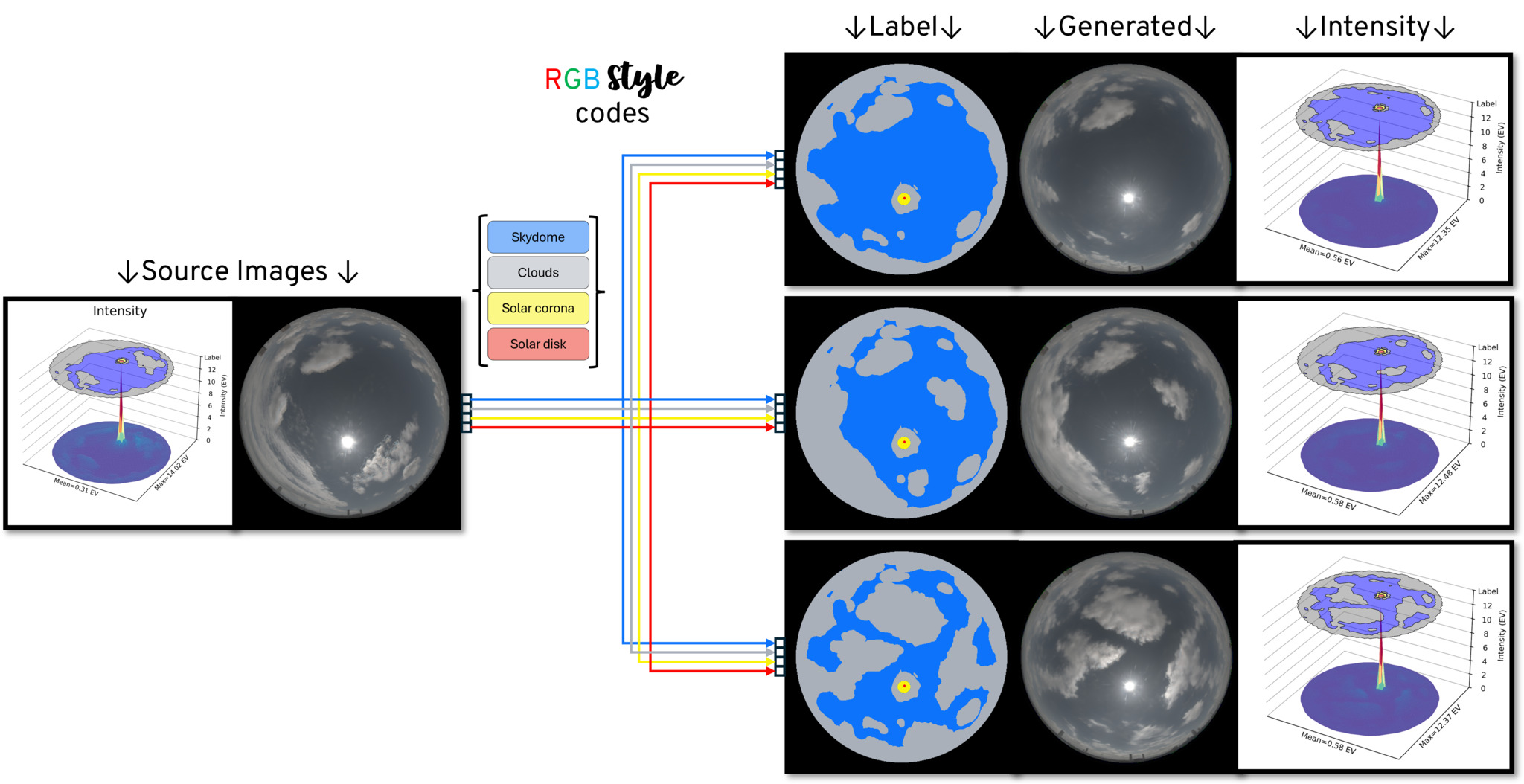}
    \caption{Illustration of \ourModel{}'s ability to edit cloud formations. As shown, the clouds of a source image can be edited to realistically remove (top row) or add (middle and top row) clouds. \ourModel{} $512^2$ with \rgb-styles and $f_{\text{\tiny Roberson}}$ fusion.}
    \label{app:fig::demo_4_cloud_editing_realistic}
\end{figure}
\end{landscape}

\begin{landscape}
\begin{figure}[p]
    \centering
    \includegraphics[height=\textwidth,width=\linewidth,keepaspectratio]{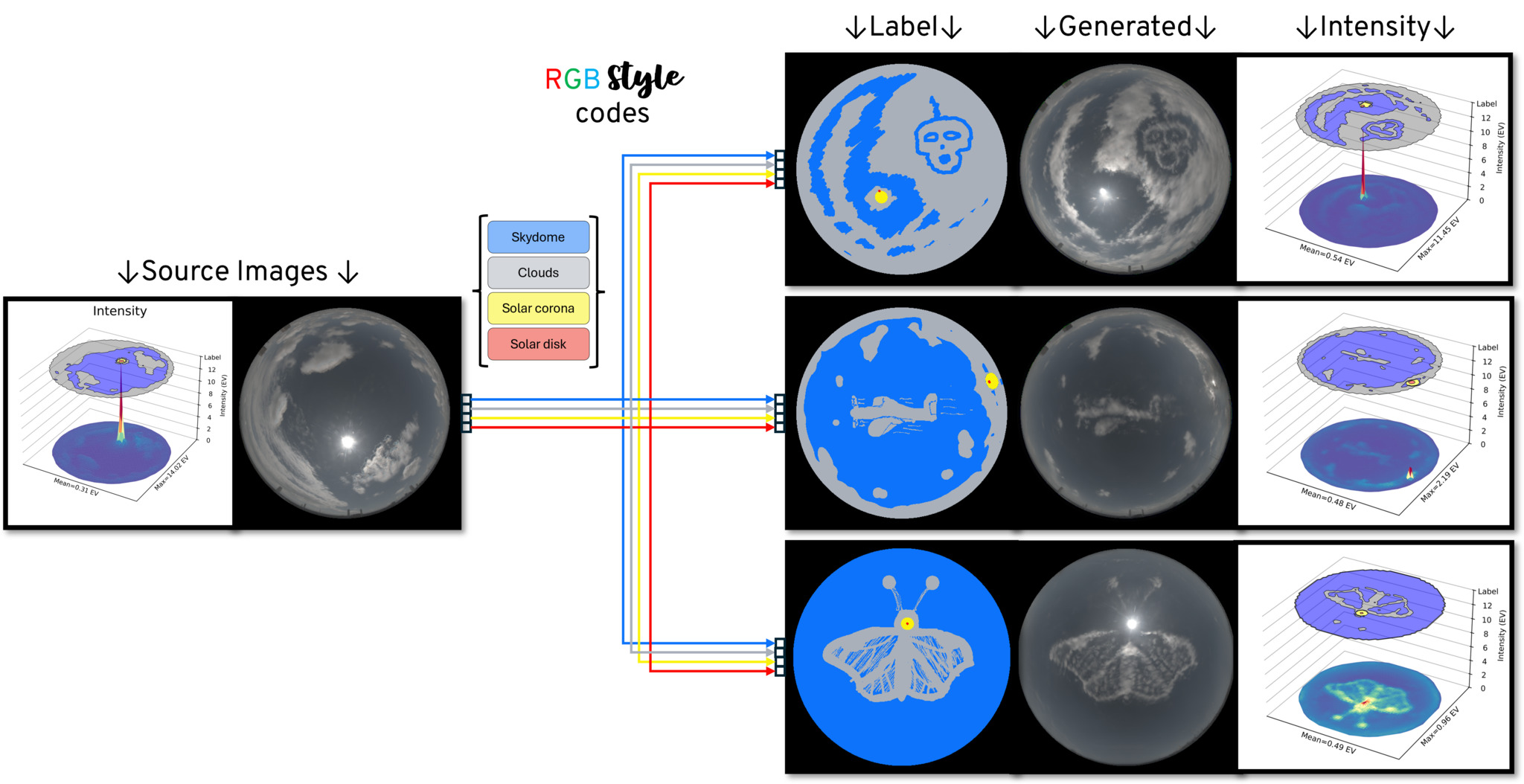}
    \caption{Illustration of \ourModel{}'s ability to support artistic liberties in creating environment maps. As shown, the styles of a source image can be used to create surreal skies. \ourModel{} $512^2$ with \rgb-styles and $f_{\text{\tiny Roberson}}$ fusion.}
    \label{app:fig::demo_5A_cloud_editing_artistic}
\end{figure}
\end{landscape}
\begin{landscape}
\begin{figure}[p]
    \centering
    \includegraphics[height=\textwidth,width=\linewidth,keepaspectratio]{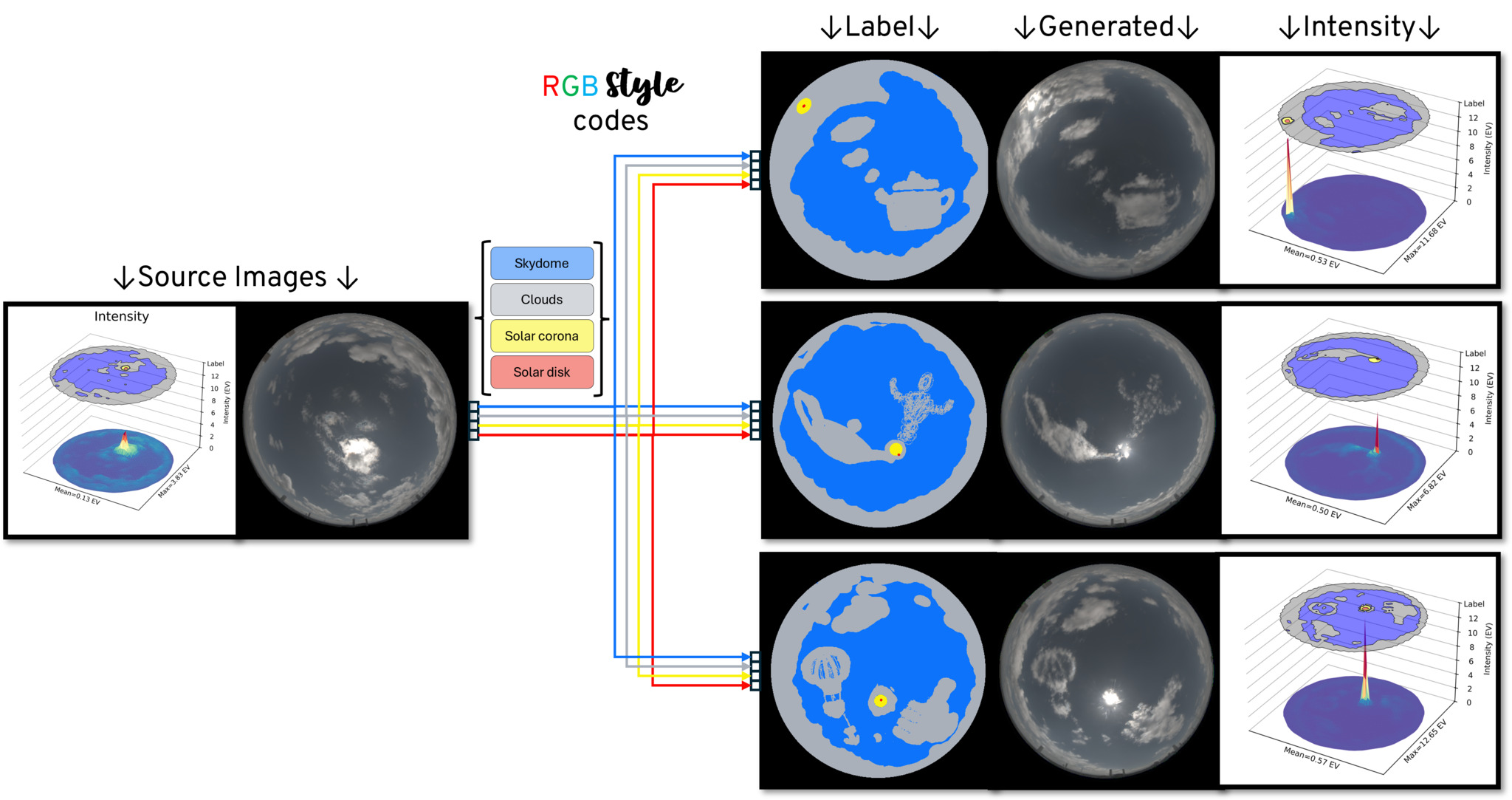}
    \caption{Illustration of \ourModel{}'s ability to support artistic liberties in creating environment maps. As shown, the styles of a source image can be used to create surreal skies. \ourModel{} $512^2$ with \rgb-styles and $f_{\text{\tiny Roberson}}$ fusion.}
    \label{app:fig::demo_5B_cloud_editing_artistic}
\end{figure}
\end{landscape}
\begin{landscape}

\begin{figure}[p]
    \centering
    \includegraphics[height=\textwidth,width=\linewidth,keepaspectratio]{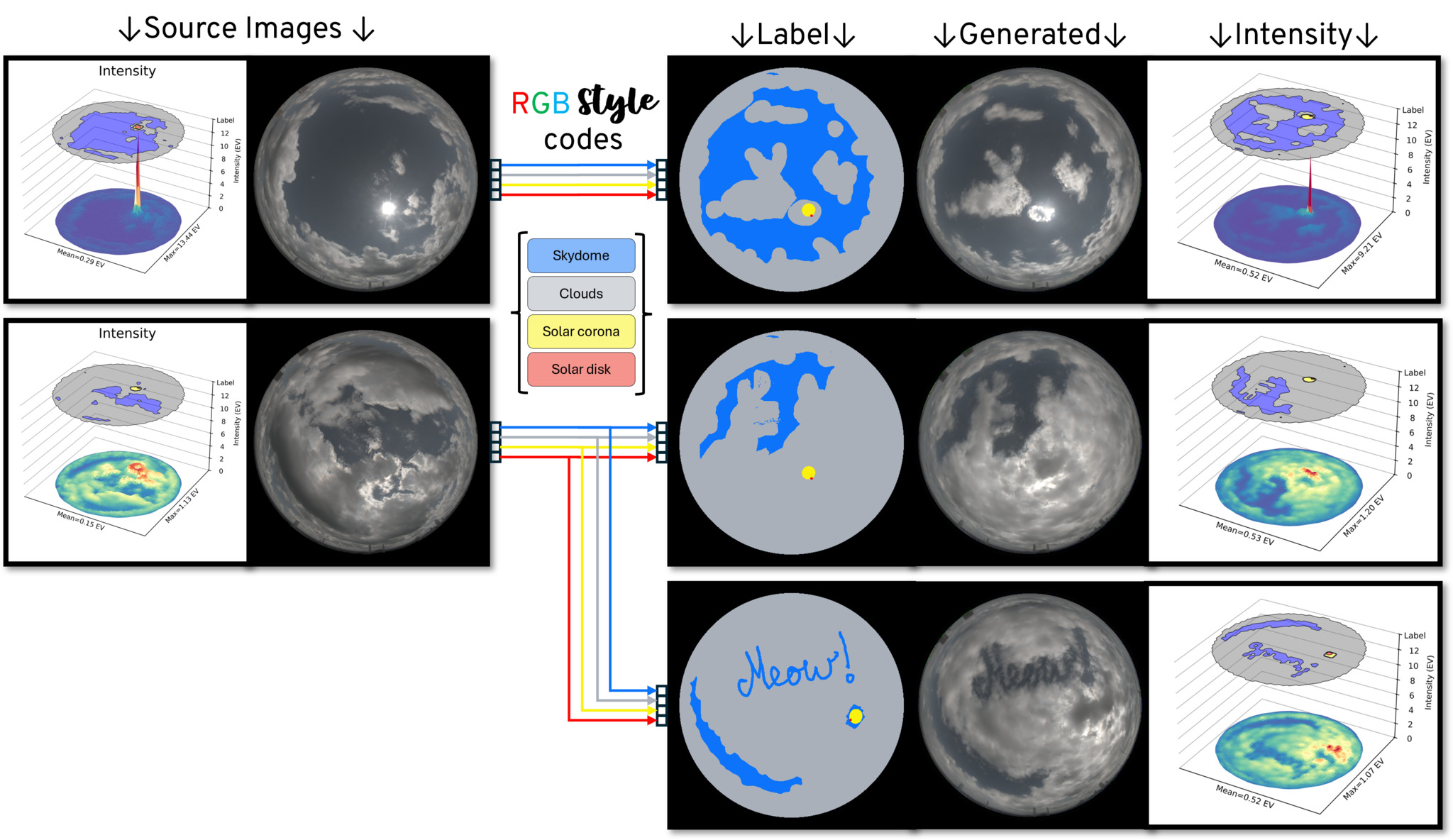}
    \caption{Illustration of \ourModel{}'s ability to support artistic liberties in creating environment maps. As shown, the styles of a source image can be used to create surreal skies. \ourModel{} $512^2$ with \rgb-styles and $f_{\text{\tiny Roberson}}$ fusion.}
    \label{app:fig::demo_5C_cloud_editing_artistic}
\end{figure}
\end{landscape}

\begin{landscape}
\begin{figure}[p]
    \centering
    \includegraphics[height=\textwidth,width=\linewidth,keepaspectratio]{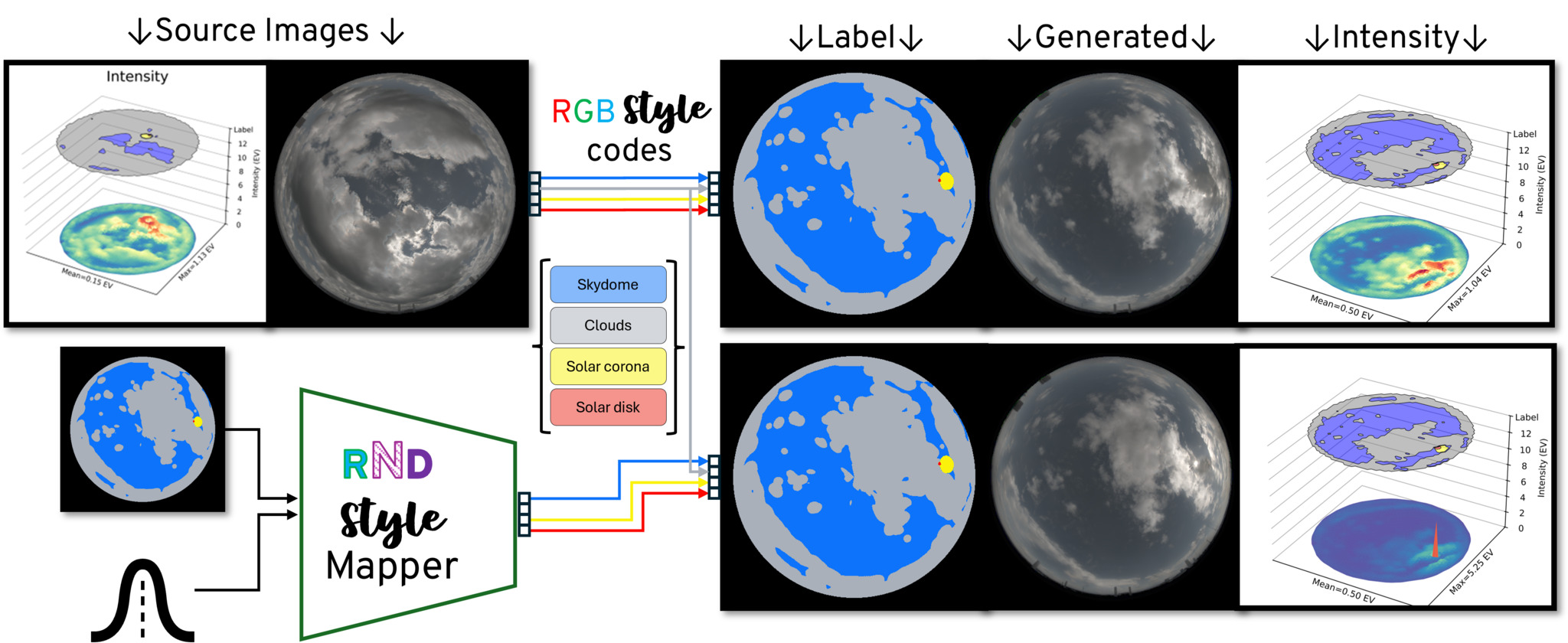}
    \caption{Illustration of \ourModel{}'s ability to support mixing of \rgb- and \rnd-styles in creating environment maps. \ourModel{} $512^2$ with \rgb-styles, \rnd-styles and $f_{\text{\tiny Roberson}}$ fusion.}
    \label{app:fig::demo_6_mixed_styles}
\end{figure}
\end{landscape}

\begin{landscape}
\begin{figure}[p]
    \centering
    \includegraphics[height=\textwidth,width=\linewidth,keepaspectratio]{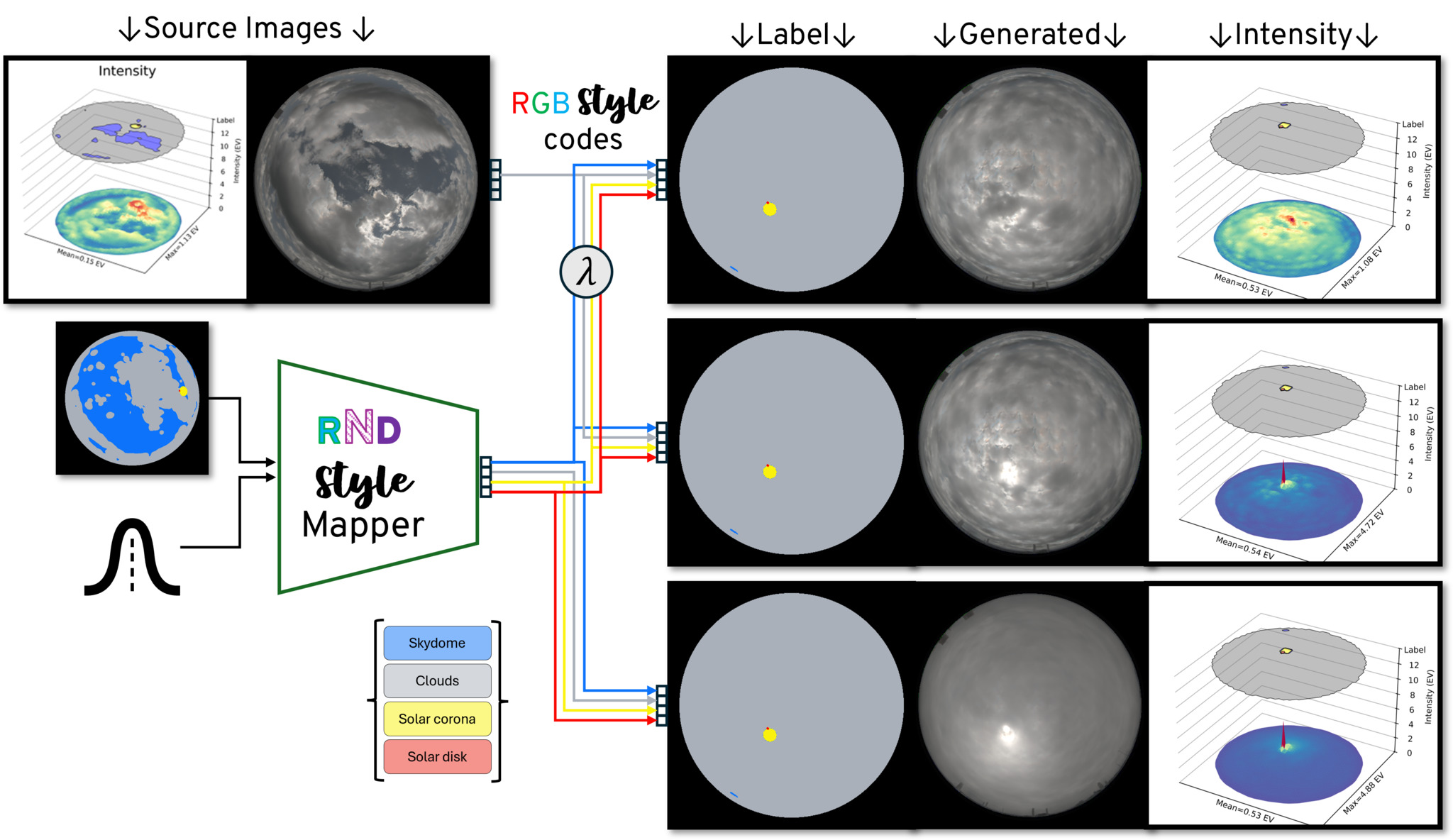}
    \caption{
    Illustration of \ourModel{}'s ability to support weighted mixing of \rgb- and \rnd-styles to creating customized environment maps.
    As depicted by $\lambda$, the weights of styles can be user-modulated to emphasize or subdue selected styles. As shown, the weights of solar features were reduced to create a more uniform overcast sky.
    \ourModel{} $512^2$ with \rgb-styles, \rnd-styles and $f_{\text{\tiny Roberson}}$ fusion.}
    \label{app:fig::demo_6b_mixed_styles}
\end{figure}
\end{landscape}
\clearpage
\section{Background (Extended)}
\label{app:background}

The following sections expand on select topics from the background in \cref{sec:background}.

\subsection{Tone Mapping Dynamic Range}
\label{app:background::tonemapping}

\begin{figure}[htb]
    \centering
    \includegraphics[width=\linewidth,keepaspectratio]{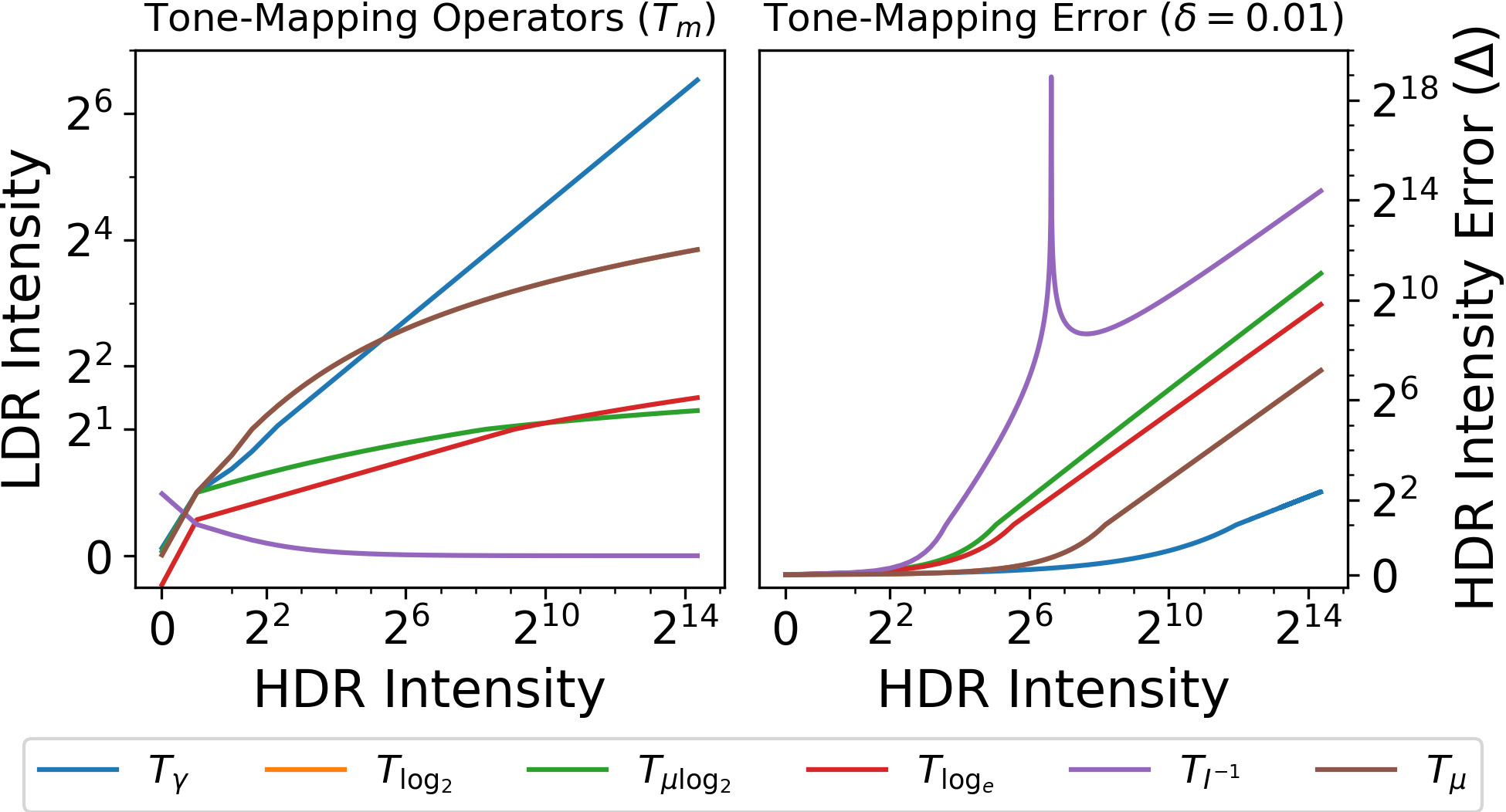}
    \caption{
    (left) Tone mapping operator ($T_m$) compression of HDR Intensity ($\hat{I}$) to LDR intensity ($\Check{I}$).  (right) Respective non-linearity between error ($\delta$=$0.01$) in compressed LDR-space and error ($\Delta$) in uncompressed HDR-space.
    As target HDR intensity increases, small errors in LDR-space result in exponentially larger errors in reconstructed HDR-space.
    }
    \label{fig:plt_tm}
\end{figure}

Tone mapping operators ($T_m$) compress HDRI to a visible, displayable, or latent colour-space otherwise favourable given $\check{I} = {T_{m}}\left(\hat{I}\right)$.
Various operators have been used in DNN illumination and sky modelling, including:
Power-Law ($T_\gamma$ \cite{YANNICK_2019_SKYNET}),
logarithmic ($T_{log_n}$ \cite{text2light}),
$\mu\text{-law}\log_2$ ($T_{\mu \log_2}$ \cite{Ian_towardsSkyModels}),
natural logarithmic ($T_{\log_e}$ \cite{SKYGAN_2022}),
and
inverted ($T_{I^{-1}}$ \cite{DEEPCLOUDS_22})
as shown in \cref{eq:tm_gamma,eq:tm_log_n,eq:tm_natural_log,eq:tm_muLawLog2,eq:tm_inverted} and \cref{fig:plt_tm} (left).

\vspace{-\baselineskip}
\begin{align}
\text{None: }& T_\varnothing \left(\hat{I}\right) = \hat{I}
    \label{eq:tm_none} \\
\text{Power-Law: }& T_\gamma \left(\hat{I}\right) = \hat{I}^{\frac{1}{\gamma}}
    \label{eq:tm_gamma} \\
\text{Logarithmic: }& T_{log_n} \left(\hat{I}\right) = \log_n(\hat{I}+1) \label{eq:tm_log_n} \\
\text{Natural Log.: }& T_{\log_e}\left(\hat{I}\right) = \left[ log(\hat{I} + \epsilon) + \beta \right] * \alpha \label{eq:tm_natural_log} \\
\mu\text{-lawLog}_2 \text{: }& T_{\mu \log_2} \left(\hat{I}\right) = \log_2\left[ \frac{\log_e(1+\mu \hat{I})}{\log_e(1.0+\mu)} +1\right] \label{eq:tm_muLawLog2} \\
\text{Inverted: }& T_{I^{-1}}\left(\hat{I}\right)
    = 1 / \left(1 + \hat{I} + 0.01\right) \label{eq:tm_inverted}
\end{align}
\vspace{-\baselineskip}

\begin{figure}[htb]
    \centering
    \includegraphics[width=\linewidth,keepaspectratio]{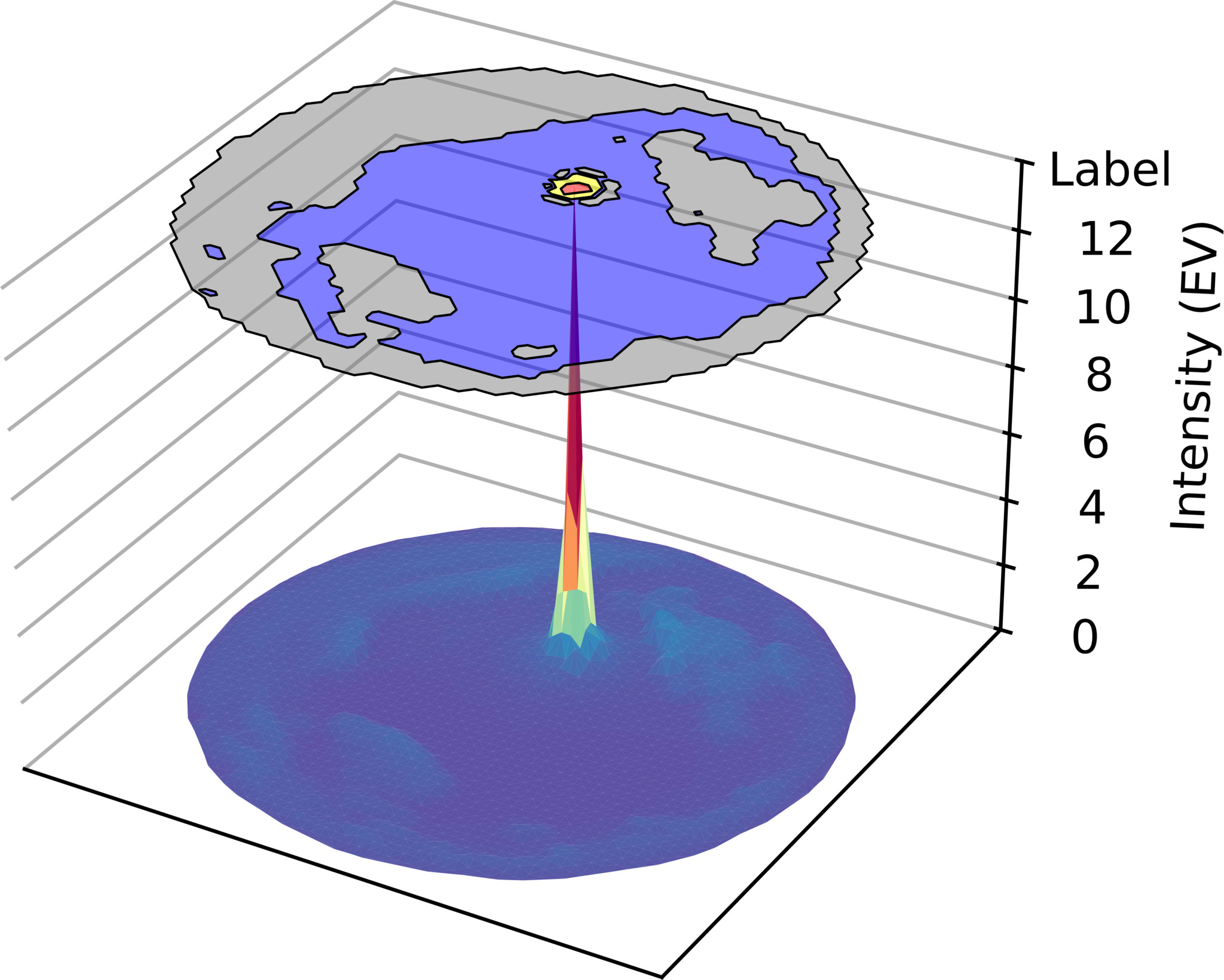}
    \caption{3D surface of skydome luminance, coloured from \textcolor{blue}{low} to \textcolor{red}{high} intensity. Disk above 3D surface illustrates skydome segmentation. The sun (\textcolor{red}{red} spike) is a small subset of pixels whose intensities belittle the remainder of the skydome. HDRDB sample from June 7th, 2016 at 1:54PM \cite{LavalHDRdb}.}
    \label{app:plt::plt_3D_surface}
\end{figure}

Each operator is a bijection allowing for the recovery of the original image via $\hat{I} = {T_{m}}^{-1}(\check{I})$.
Given $\Delta \hat{I} = | \hat{I} - {T_m}^{-1}({T_m}(\hat{I})-\delta)|$ as shown in \cref{fig:plt_tm} (right), these operators introduce a non-linearity between error ($\delta$) in LDR compressed space and error ($\Delta$) in uncompressed HDR space.
The impact of non-linear-error is most pronounced with the solar disk (intensity-`spike' illustrated in \cref{app:plt::plt_3D_surface}), where a small error ($\delta$) to an LDR space solar disk results in a large error ($\Delta$) to the HDR space solar disk.
This results in a profound impact to illumination in downstream applications such as IBL rendering.

\subsection{Mitigation Strategies (Continued)}
\label{app:background::mitigation_strategies}

\begin{landscape}
\begin{figure}[p]
    \centering
    \includegraphics[height=\textwidth,width=\linewidth,keepaspectratio]{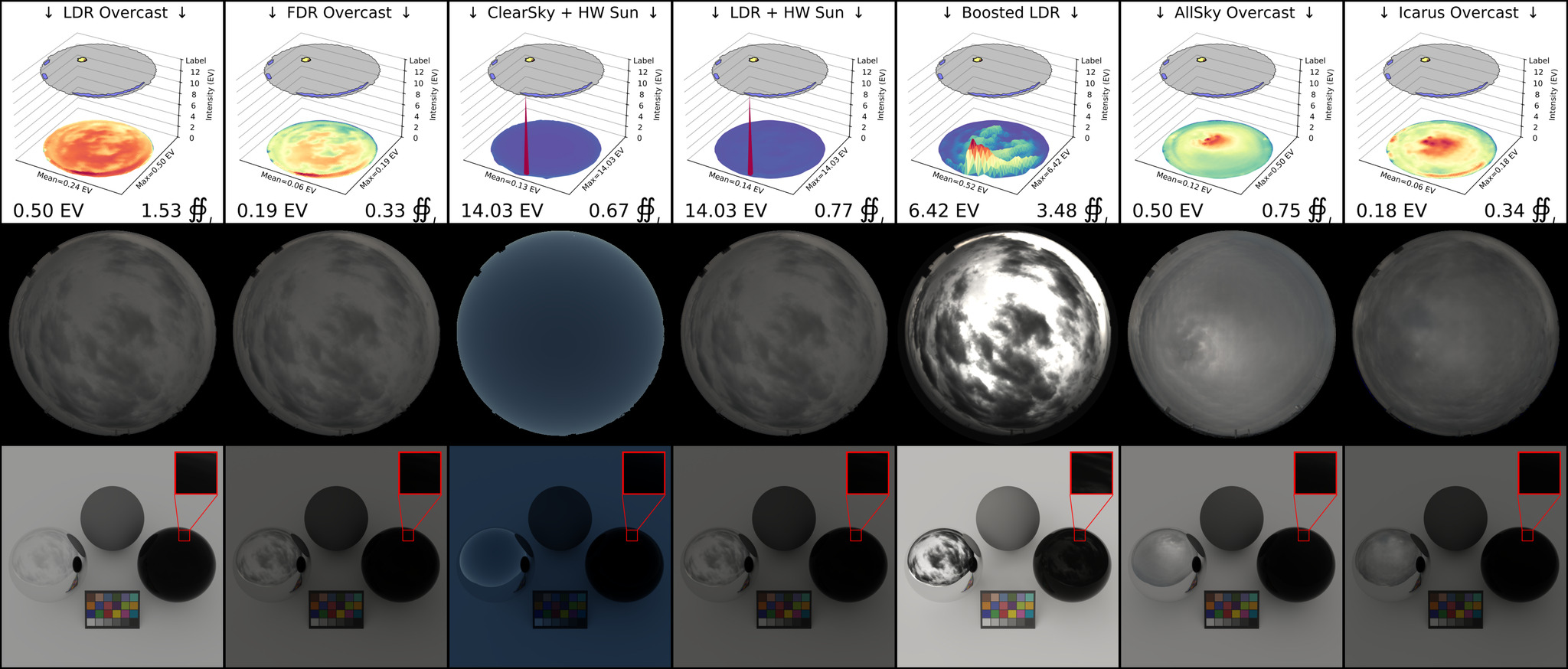}
    \caption{
        \textbf{Overcast sky}: Mitigation of solar modelling in environment maps through substitution of a parametric Hošek-Wilkie sun (HW \cite{HOSEK_13Sun,DEEPCLOUDS_22}) and manual parametric boosting of the HDR environment map (Boosted \cite{text2light}; $\gamma$=0.5, $\beta$=2, $\rho$=6).
        Adding an HW sun (\textit{LDR+HW Sun}) skips atmospheric attenuation and allows the sun to `pierce' through clouds to create strong shadows.
        If the sun is obstructed, Boosting an HDR image (\textit{Boosted LDR}) is prone to over-exposing and producing unpredictable shadows.
        Both mitigation strategies alter the perceived tones in IBL renderings (lambertian planar surface and color chart), but AllSky \cite{Ian_towardsSkyModels} and our model (\textit{\ourModel{}}, right column) accurately model real-world FDR illumination for photorealistic IBL renderings.
    }
    \label{app:fig::quickFix_overcastSky}
\end{figure}
\end{landscape}

\begin{landscape}
\begin{figure}[p]
    \centering
    \includegraphics[height=\textwidth,width=\linewidth,keepaspectratio]{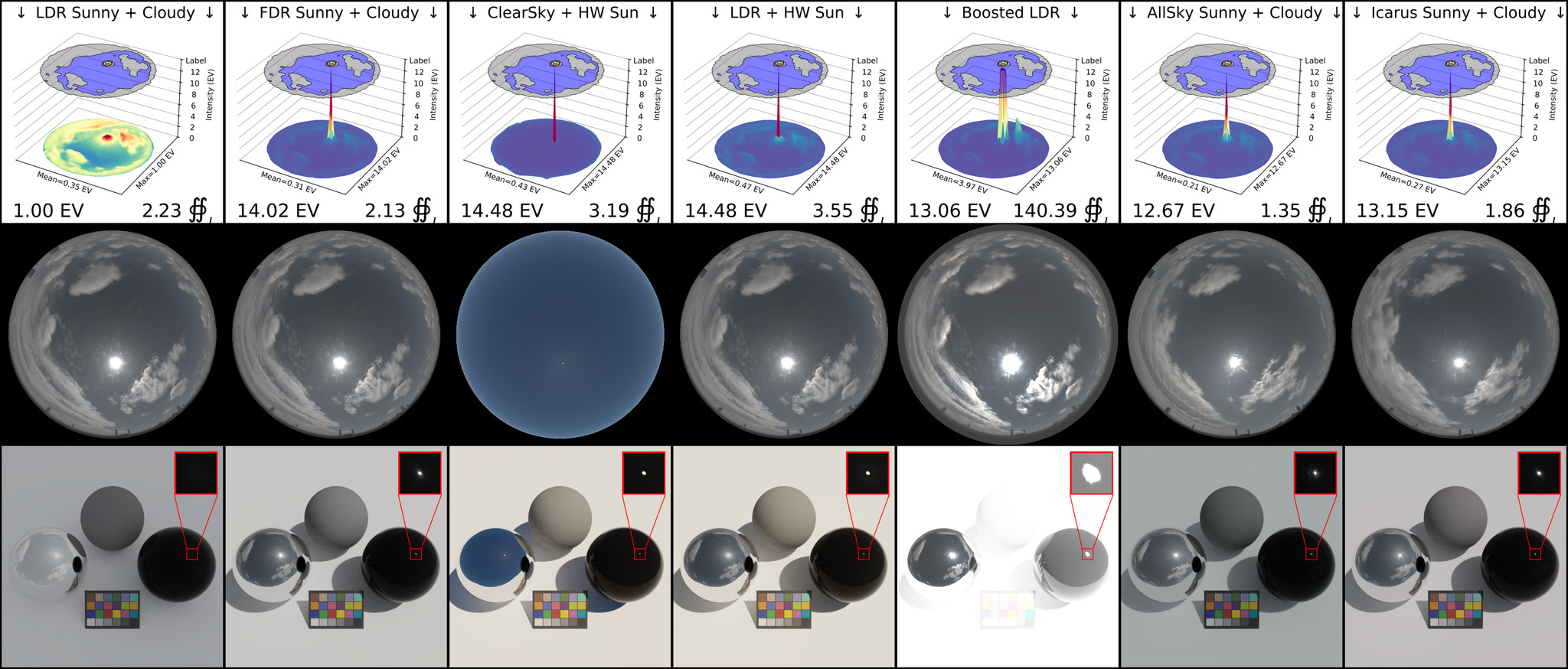}
    \caption{
        \textbf{Cloudy with unobstructed sun}: Mitigation of solar modelling in environment maps through substitution of a parametric Hošek-Wilkie sun (HW \cite{HOSEK_13Sun,DEEPCLOUDS_22}) and manual parametric boosting of the HDR environment map (Boosted \cite{text2light}; $\gamma$=0.5, $\beta$=2, $\rho$=6).
        When the sun is unobstructed, adding an HW sun creates crisp shadows, realistic light-transmission (black glass orb) and only minor alteration of perceived tones (lambertian surface, color chart).
        Boosting an HDR image is subjective to user-selection of parameters, which drives both the algorithms' selection of features and enhancement of intensity.
        Boosting is prone to over-exposing rendered scenes, producing unpredictable shadows, excessive light-transmission (black glass orb), and severe alteration of perceived tones (lambertian planar surface and color chart).
        While \textit{AllSky} \cite{Ian_towardsSkyModels} mitigates the skewed tones from the parametric sun and over-saturation of boosting, \textit{\ourModel{}} (right column) offers greater photorealism and accuracy in illumination (EV and $\oiint_I$).
    }
    \label{app:fig::quickFix_cloudy_wSun}
\end{figure}
\end{landscape}

\begin{landscape}
\begin{figure}[p]
    \centering
    \includegraphics[height=\textwidth,width=\linewidth,keepaspectratio]{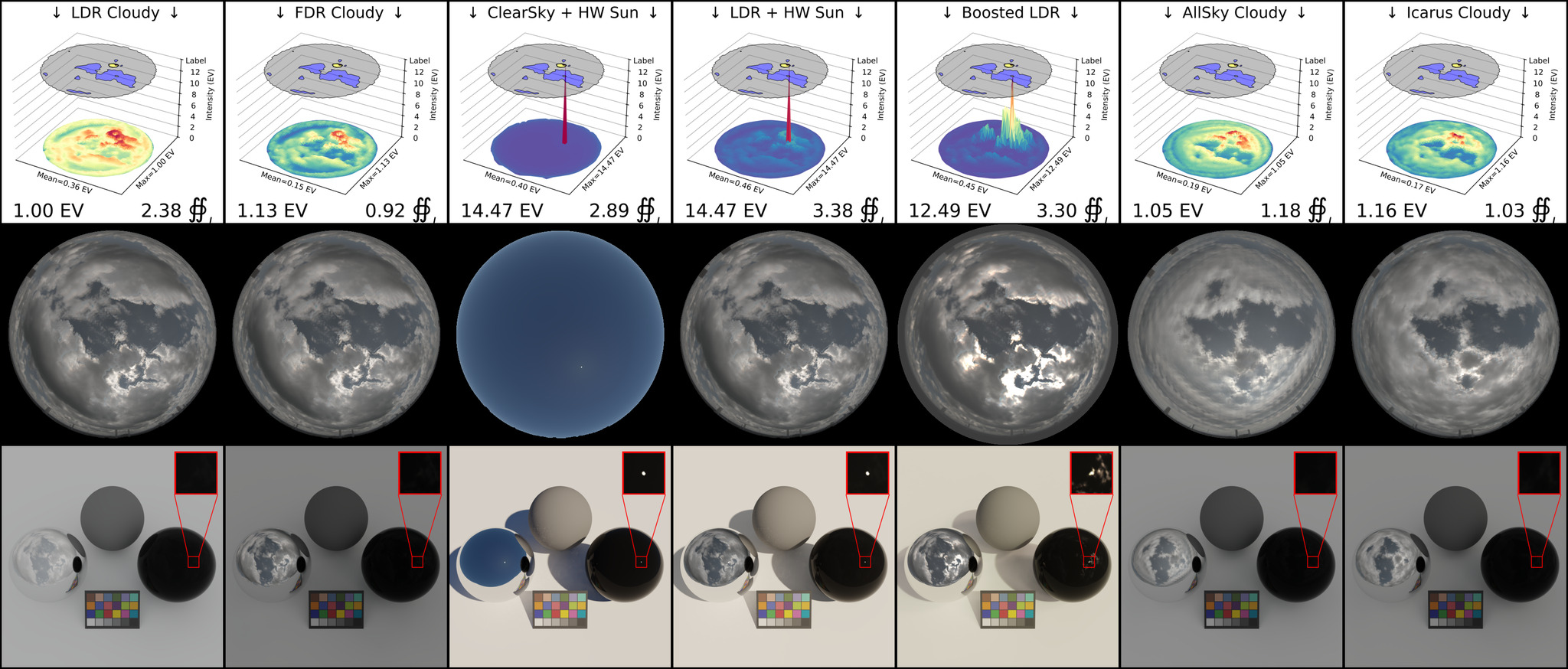}
    \caption{
        \textbf{Cloudy with obstructed sun}: Mitigation of solar modelling in environment maps through substitution of a parametric Hošek-Wilkie sun (HW \cite{HOSEK_13Sun,DEEPCLOUDS_22}) and manual parametric boosting of the HDR environment map (Boosted \cite{text2light}; $\gamma$=0.5, $\beta$=2, $\rho$=6).
        Adding an HW sun skips atmospheric attenuation and allows the sun to `pierce' through clouds to create strong shadows and unexpected light-transmission (black glass orb).
        If the sun is obstructed, boosting is prone to over-exposing (see skydome in $2^{nd}$ row and light-transmission to black-glass orb), producing unpredictable shadows and excessive light-transmission (black glass orb).
        \textit{AllSky} \cite{Ian_towardsSkyModels} and \ourModel{} both mitigate alteration of perceived tones (lambertian planar surface and color chart), with \ourModel{} (right column) offering greater photorealism and accuracy in illumination (EV and $\oiint_I$).
    }
    \label{app:fig::quickFix_cloudy_woSun}
\end{figure}
\end{landscape}

\begin{landscape}
\begin{figure}[p]
    \centering
    \includegraphics[height=\textwidth,width=\linewidth,keepaspectratio]{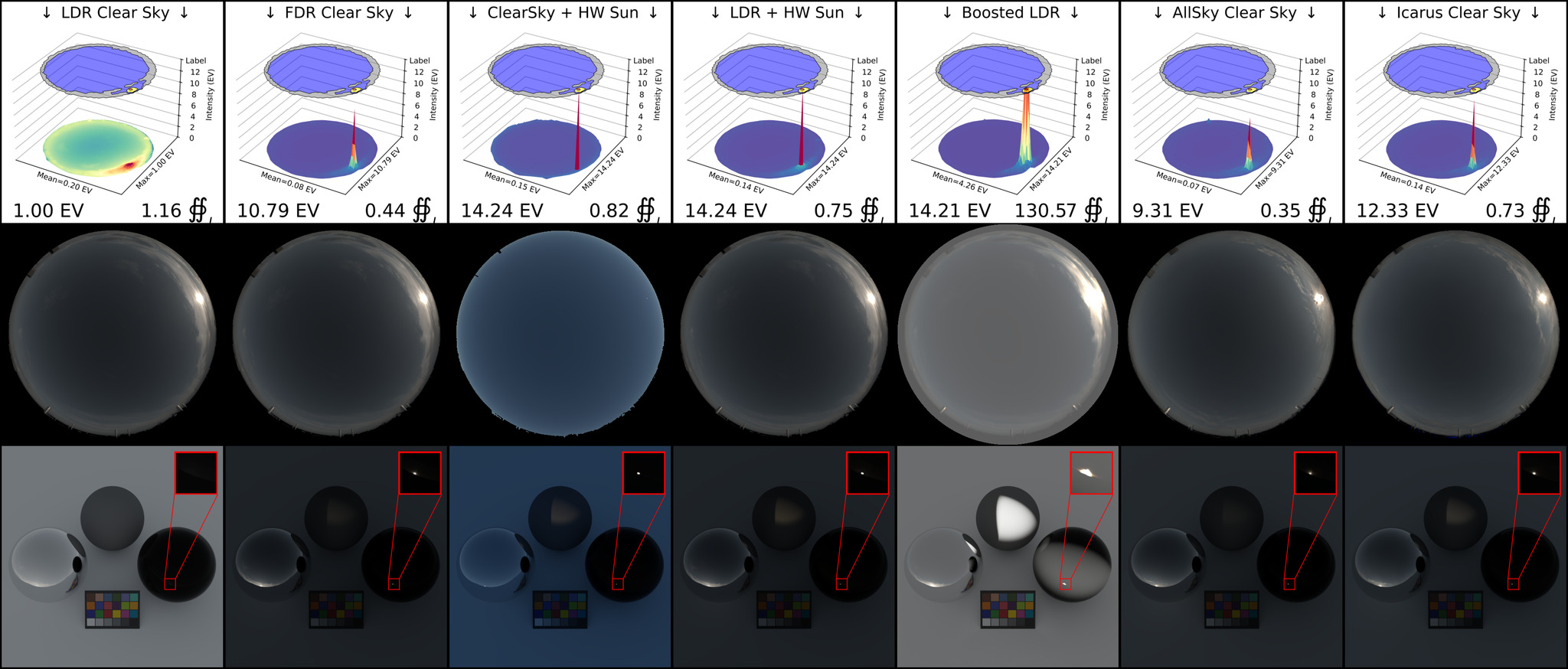}
    \caption{
        \textbf{Clear sky}: Mitigation of solar modelling in environment maps through substitution of a parametric Hošek-Wilkie sun (HW \cite{HOSEK_13Sun,DEEPCLOUDS_22}) and manual parametric boosting of the HDR environment map (Boosted \cite{text2light}; $\gamma$=0.5, $\beta$=2, $\rho$=6).
        In low-light settings such as dusk/sunset/sunrise, adding a parametric sun can be difficult given minor errors in camera calibration.
        Though an added HW sun `pierces' through clouds formations with $10.9\!\times$ linear intensity of ground truth \textit{FDR Clear Sky}, the absence of clouds acting as light-sources allows for relatively-indirect HW solar illumination to produce visually appealing renders without skewed tones.
        Boosting is subjective to user-selection of parameters driving the algorithms' selection of features, which in low-light settings is shown to be prone to over-exposing the entirety of the skydome.
        This over-exposure results in unpredictable shadows (lambertian orb), excessive light-transmission (black glass orb) and altered tones (lambertian planar surface and color chart).
        \textit{AllSky} \cite{Ian_towardsSkyModels} and \ourModel{} both mitigate alteration of perceived tones (lambertian planar surface and color chart), producing renderings with accurate tones, shadows and light transmission.
    }
    \label{app:fig::quickFix_clearSky}
\end{figure}
% \end{sidewaysfigure}
\end{landscape}

Strategies have been proposed to mitigate inaccurate modelling of solar luminous intensity including substitution of the solar disk \cite{DEEPCLOUDS_22}, manual parametric boosting \cite{text2light} and composite shading.
In \cref{app:fig::quickFix_overcastSky,app:fig::quickFix_clearSky,app:fig::quickFix_cloudy_woSun,app:fig::quickFix_cloudy_wSun}, we compare these mitigation strategies to AllSky \cite{Ian_towardsSkyModels} and \ourModel{} ($512^2$ with \rgb-styles and $f_{\text{\tiny Robertson}}$ fusion), implementing parametric Hošek-Wilkie clear skies (uniform 0.5 \rgb albedo, 1.0 turbidity) and suns \cite{HOSEK_13,HOSEK_13Sun,DEEPCLOUDS_22}, and boosting with a static set of user-selected parameters ($\gamma$=0.5, $\beta$=2, $\rho$=6) selected for similar solar intensity to FDR ground truth in \cref{app:fig::quickFix_cloudy_wSun}.
To mitigate calibration errors which might result in the misalignment of the solar disk, HW clear skies are matched FDR ground truth imagery (as done by CloudNet \cite{DEEPCLOUDS_22}, though skipping histogram equalization).

As demonstrated in \cref{app:fig::quickFix_overcastSky,app:fig::quickFix_clearSky,app:fig::quickFix_cloudy_woSun,app:fig::quickFix_cloudy_wSun},  mitigation strategy performance is varied, with no strategy generalizing to the four primary sky configurations:
clear, cloudy, cloudy with overcast sun and overcast skies.
As such, mitigation strategies require per-instance human intervention to make corrections and/or tune parameters.
That said, AllSky and \ourModel{} demonstrate reliable generalization across the four primaries, with renderings illustrating a high fidelity to ground truth illumination, shadows, tones and light transmission.
Though some stochasticity in illumination is to be anticipated per natural real-world stochasticity in illumination, \ourModel{} is shown to provide improved illumination accuracy and greater photorealism.

\begin{minipage}[htb]{0.9\textwidth}
% \begin{figure}[htb]
    \centering
    \includegraphics[width=0.9\linewidth,keepaspectratio]{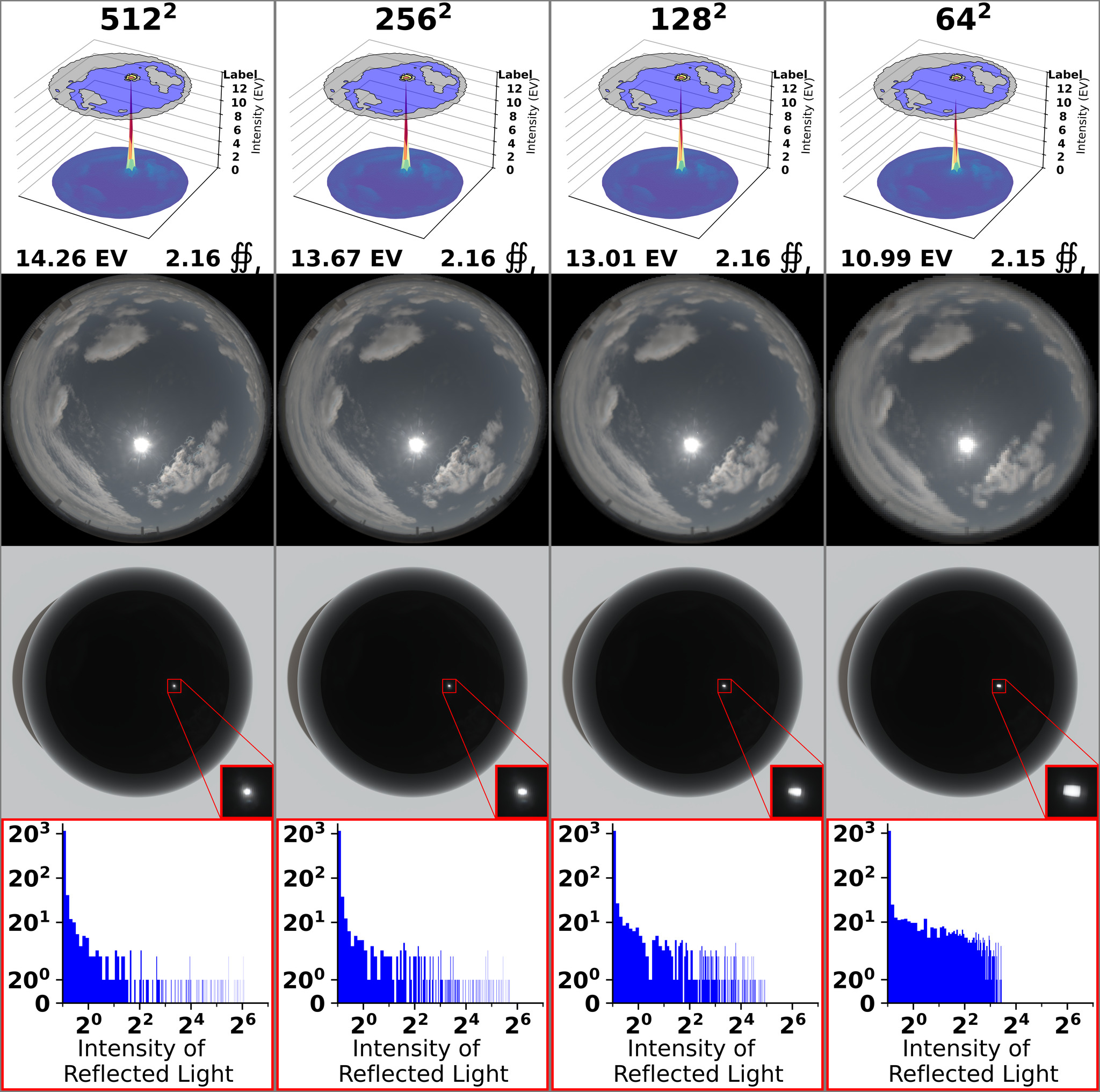}
    % \caption{
    \captionof{figure}{
    Downsampling latlong environment maps from sky-angular shape $512^2$ to $64^2$. [Top row] Downsampling with area-interpolation retains illuminance ($\oiint_I$), but results in a loss of intensity from 14$EV$ to 11$EV$ (38\% retention of linear space intensity).
    Though this has minimal impact to the visual quality of environment maps ($2^{nd}$ row), this is reflected in IBL scenes ($3^{rd}$ row, black glass orb) through a solar disk with increasing diameter both in the sky (not shown) and on incident surfaces (\textcolor{red}{zoom box}).
    As shown in the histograms (bottom row) of the \textcolor{red}{zoom boxes}, the intensity of light reflected off of the transmissible surface decrease from $2^{6.1}$ at $512^2$ to $2^{3.6}$ at $64^2$ (16\% retention of linear space intensity).
    }
    \label{app:fig::grid_demo_downsampling}
% \end{figure}
\end{minipage}

\subsection{Impact of Downsampling}
\label{app:background::downsampling}

As shown in \cref{app:fig::grid_demo_downsampling}, inter-area downsampling preserves an environment map's luminance flux ($\oiint_I$), but dilation of the solar disk results in the loss of the exposure range ($EV$).
The distinction between these measures is the quantification of global-illumination and peak-solar intensity, where in this example the four-pixels solar region with intensity $EV{\geq}11$ of the $512^2$ skydome represents 55\% of the environment maps' illumination.
Though illuminance of the scene is consistent, the luminous intensity of the solar disk drives shadows and light-transmission, key to downstream applications including photorealistic IBL rendering \cite{Ian_towardsSkyModels} and simulating the output of solar systems \cite{MOUSAZADEH20091800}.

\clearpage

\begin{figure*}[htb]
\centering
\includegraphics[height=0.95\textheight,width=0.8\textwidth,keepaspectratio]{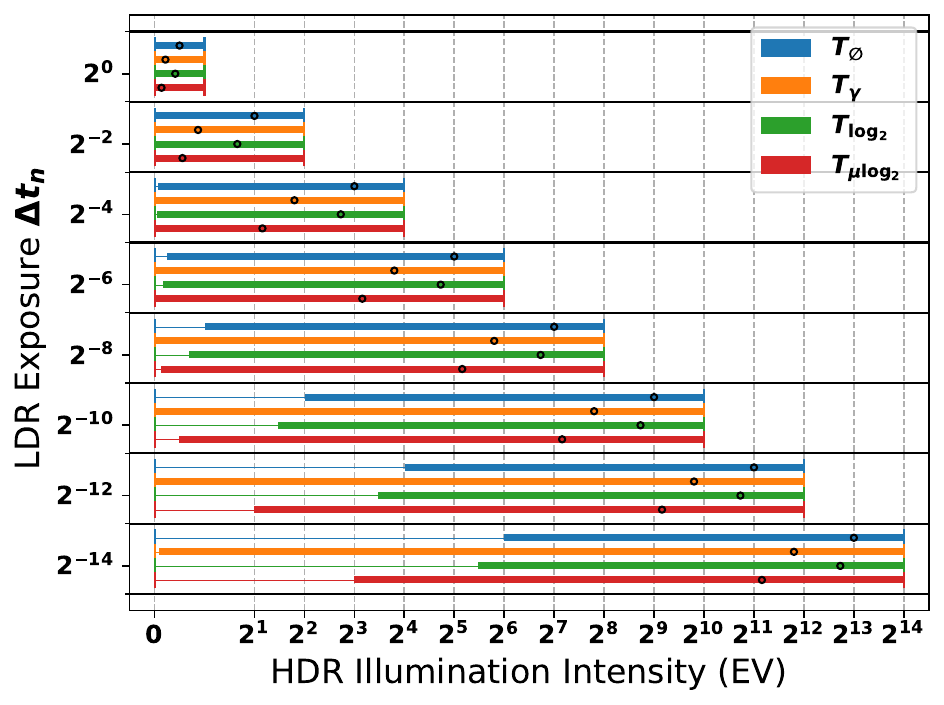}
\caption{
    Visual representation an LDR exposure bracket normalized to HDR-space by \cref{eq:normalize_ldr_exposure} for various tone mapping operators.
    Each LDR exposure is a `candle-stick' where
    upper- and lower-limits illustrate min/max HDR intensities for $\Check{I}$ clipped to $[0,1]$ and the body min/max HDR intensities for $\Check{I}$ clipped to $[\epsilon, 1-\epsilon]$ for $\epsilon=\underline{\epsilon}=\overline{\epsilon}=\frac{1}{255}$.
    Markers ($\circ$) indicate the HDR illumination intensity of an LDR exposure value of 0.5.
    Insufficient overlap and/or gaps between LDR exposures should be avoided during exposure selection.
}
\label{app:dia::dia_HDR_bracket}
\end{figure*}

\begin{figure*}[htbp]
\centering
\includegraphics[height=0.95\textheight,width=1.\textwidth,keepaspectratio]{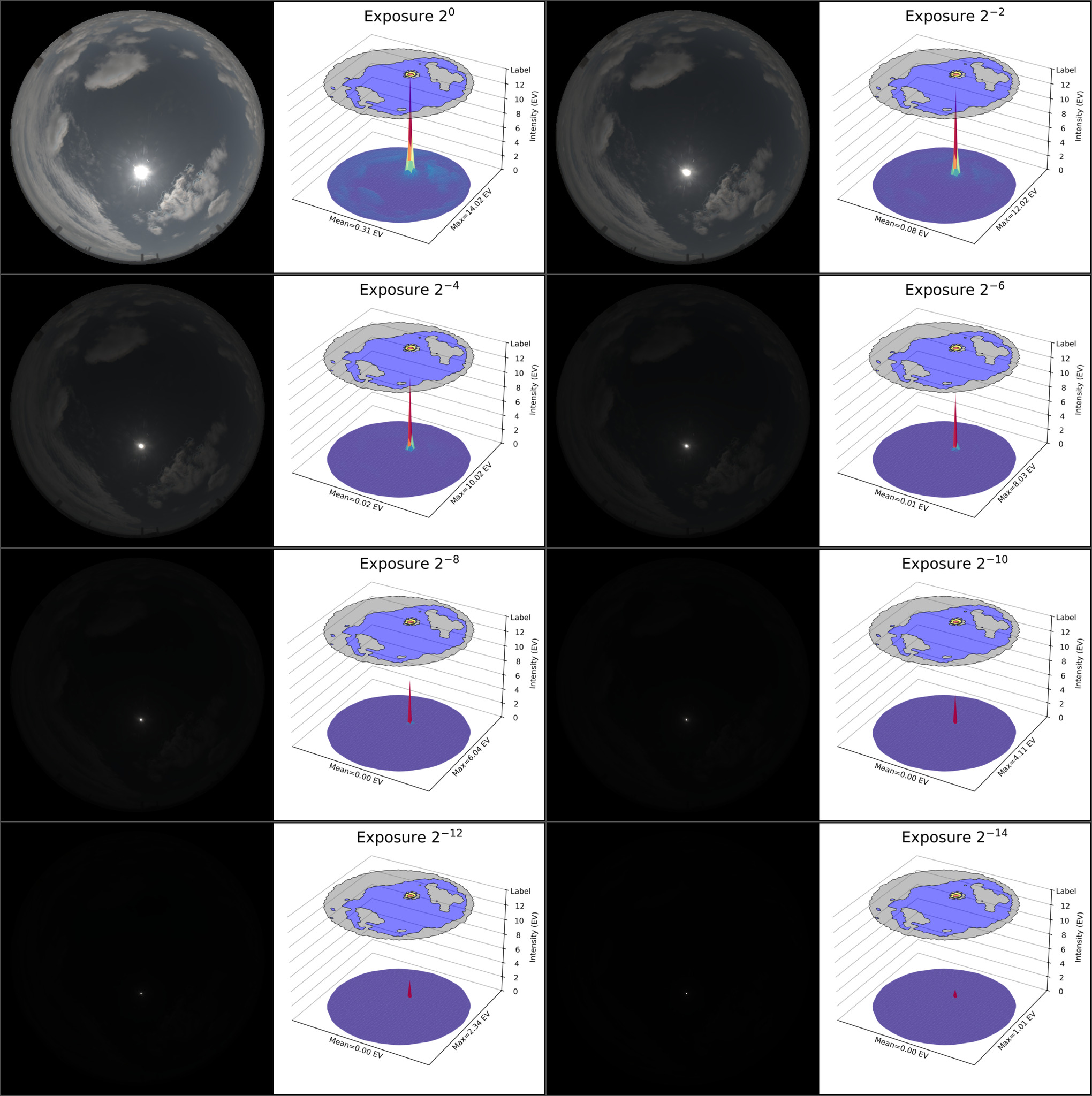}
\caption{
    Decomposition of an FDR image to LDR exposures by modulation of exposure $\Delta t=2^{-x}$, demonstrating atmospheric features are often visually indiscernible in exposures modulated by $2^{-6}$ or less.
    In exposure $2^{-14}$, the features of the solar corona is indiscernible and only the solar disk remains.
}
\label{app:fig::dia_HDR_bracket}
\end{figure*}

\section{Methodology (Extended)}
\label{app:methodology}

We mathematically express images $I$ with the following modifiers and characteristics:
\begin{enumerate}
\item{$I$: for an image}
\item{$\mathcal{I}$: for a generated image}
\item{$\Check{I}$: for LDR colour images}
\item{$\hat{I}$: for HDR colour images}
\item{$\xi$: for latent images }
\item{$I_c$: for an image's $c^{th}$ colour channel}
\item{$I_{ij}$: for row $i$ and col $j$ indexing of 2D matrices.}
\item{$I_{n,c,ij}$: for the $n^{th}$ exposure at color-channel $c$, row $i$ and col $j$ for 4D matrices of size $N\!\times\!C\!\times\!I\!\times\!J$.}
\item{$I_n$: for the $n^{th}$ image of an $N$-exposure LDR bracket}.
\item $\{I_n\}^N$: for an $N$-exposures bracket (set) of $I_n$ images.
\item{$k$: class $k$ of the set of classes $\mathcal{K}$}
\item{$\rchi_k$: for a style code for class $k$}.
\item{${\rchi_k}^\mathcal{K}$: for the set of style codes corresponding to the $k$ classes in $\mathcal{K}$}.
 \end{enumerate}

This allows for HDR images to be expressed as $\hat{I}_{n,c,ij}$ and LDR images  part of an LDR bracket as $\{\Check{I}_{n,c,ij}\}^N$.
Where possible, we omit indiscriminative indexes to keep equations succinct.

\subsection{LDR Bracketing}
\label{app:methodology::ldr_brackets}

In \cref{app:dia::dia_HDR_bracket} we expand the visual of LDR exposure brackets in \cref{sec:methodology::ldr_bracketing} to include various tone mapping operators.
By comparing candle-stick lower-limits and bodies, it can be shown that tone mapping operators can reduce the number of exposures required for coverage of a desired exposure range.
This is well exemplified through the range of illumination intensity saturated in $T_\varnothing$'s lower limit but encompassed by $T_\gamma$'s elongated body.
In this regard, \cref{fig:plt_tm} demonstrates the non-linearity of tone mapping operators where, for example, $T_\gamma$ decompresses values $<1$ and compresses values $>1$.

Careful consideration should be taken in determining requirements for overlapping exposures given tone mapping operator selection, distribution of image values, and desired granularity in modeling of HDR intensity.
As illustrated in \cref{app:fig::dia_HDR_bracket}, linear coverage of FDR outdoor illumination intensity can result in many exposures with limited features and a limited ability to contribute to the desired exposure range.

\subsection{LDR Exposure Fusion: \hsv}
\label{app:methodology::ldr_brackets:::HSV}

We propose a simple fusion algorithm which takes advantage of \hsv color space, converting $I_{\rgb}$ to $I_{\hsv}$ per \cite{opencv}:

\begin{align}
\textcolor{darkgray}{V}  = & \max(\textcolor{red}{R},\textcolor{teal}{G},\textcolor{blue}{B}) \label{eq:rgb2hsv::V} \\
\textcolor{red}{S}  =  &
\begin{dcases}
    \frac{(\textcolor{darkgray}{V}-\min(\textcolor{red}{R},\textcolor{teal}{G},\textcolor{blue}{B})}{\textcolor{darkgray}{V}} & \text{if } \textcolor{darkgray}{V}\!\neq\!0 \\
    0 & \text{otherwise}
\end{dcases} \label{eq:rgb2hsv::S} \\
\textcolor{cyan}{H}  = &
\begin{dcases}
   \frac{60(\textcolor{teal}{G} - \textcolor{blue}{B})}{\textcolor{darkgray}{V}-\min(\textcolor{red}{R},\textcolor{teal}{G},\textcolor{blue}{B}) }& \text{if } \textcolor{darkgray}{V}\!=\!\textcolor{red}{R}  \\
   \frac{120+60(\textcolor{blue}{B} - \textcolor{red}{R})}{(\textcolor{darkgray}{V}-\min(\textcolor{red}{R},\textcolor{teal}{G},\textcolor{blue}{B}))} & \text{if } \textcolor{darkgray}{V}\!=\!\textcolor{teal}{G} \\
   \frac{240+60(\textcolor{red}{R} - \textcolor{teal}{G})}{(\textcolor{darkgray}{V}-\min(\textcolor{red}{R},\textcolor{teal}{G},\textcolor{blue}{B}))} & \text{if } \textcolor{darkgray}{V}\!=\!\textcolor{blue}{B} \\
   0 & \text{if }  \textcolor{red}{R}\!=\!\textcolor{teal}{G}\!=\!\textcolor{blue}{B}
\end{dcases}
\label{eq:rgb2hsv::H}
\end{align}

Converting images to \hsv disentangles brightness from chroma and hue, enabling the fusion of brightness channels (e.g. per \cref{eq:LDR_exposure_fusion} with $\check{I}_{c=V}$), and the creation of a fused HDR image by recombining as $\hat{I}_{\hsv} = \left\{\Check{I}_{0,c={\textcolor{cyan}{H}}},\Check{I}_{0,{c=\textcolor{red}{S}}},\hat{I}_{c=\textcolor{darkgray}{V}}\right\}$.

\subsection{LDR Exposure Fusion: Robertson}
\label{app:methodology::ldr_brackets:::Robertson}

\begin{figure}[htb]
    \centering
    \begin{subfigure}{.45\linewidth}
        \centering
        \includegraphics[width=\linewidth]{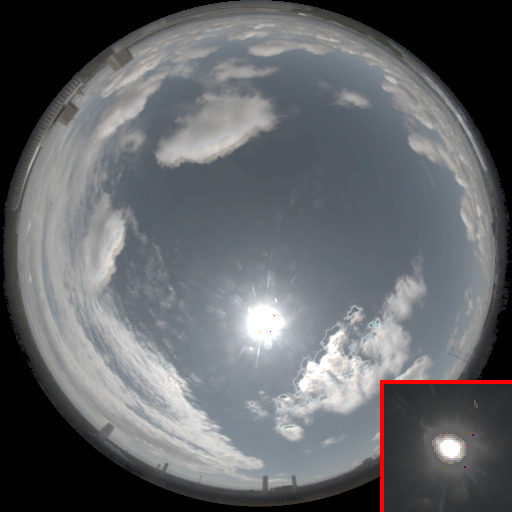}
        \caption{Robertson (OpenCV \cite{opencv})}
        \label{app:fig::HDR_robertson_cv2}
    \end{subfigure}
    \begin{subfigure}{.45\linewidth}
        \centering
        \includegraphics[width=\linewidth]{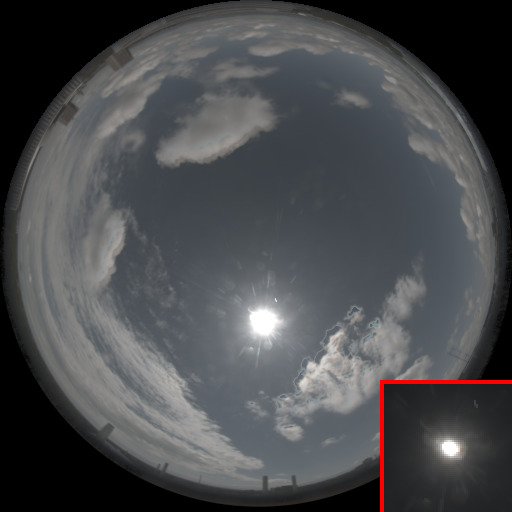}
        \caption{Robertson (Ours)}
        \label{app:fig::HDR_robertson_ours}
    \end{subfigure}
        \begin{subfigure}{\linewidth}
        \centering
        \includegraphics[width=\linewidth]{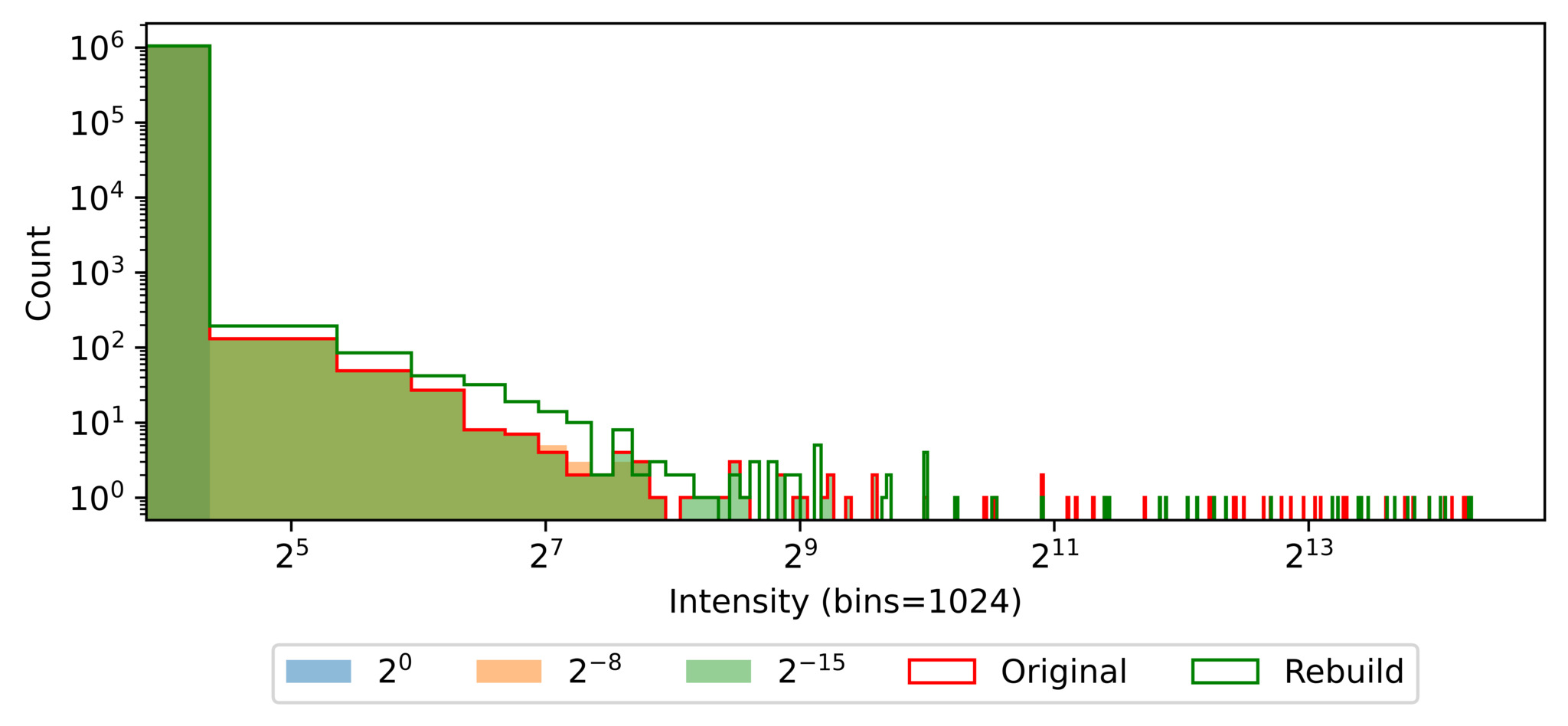}
        \caption{Robertson (OpenCV \cite{opencv})}
        \label{app:hist::HDR_robertson_cv2}
    \end{subfigure}
    \begin{subfigure}{\linewidth}
        \centering
        \includegraphics[width=\linewidth]{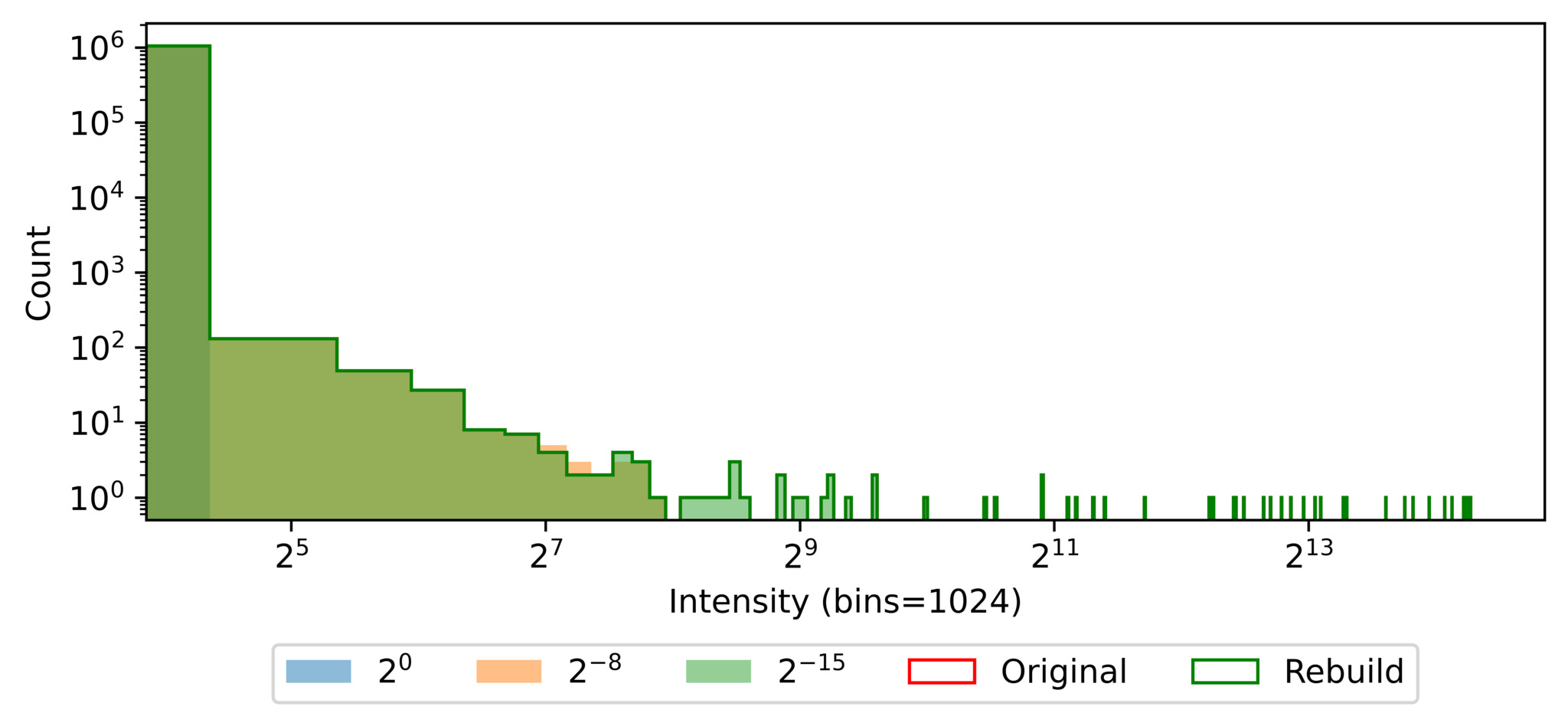}
        \caption{Robertson (Ours)}
        \label{app:hist::HDR_robertson_ours}
    \end{subfigure}
    \caption{
    LDR exposure bracket $[2^0, 2^{8},2^{15}]$ merged with Robertson exposure fusion, as computed by OpenCV LUT \cite{opencv} and by our inline implementation \cref{app:eq::LDR_exposure_fusion_robertson}.
    As shown in \cref{app:fig::HDR_robertson_ours}, our implementation mitigates seams for a smoother FDR sun.
    Error between the ground truth (original, \textcolor{red}{red outline}) and reconstructed (rebuild, \textcolor{teal}{green outline}) exposure range is shown in \cref{app:hist::HDR_robertson_cv2} and \cref{app:hist::HDR_robertson_ours} to be significantly reduced.
    }
    \label{app:fig::merge_robertson}
\end{figure}

Initial results showed that Robertson exposure fusion \cite{merge_Robertson} lost visual performance due to conventional implementations (e.g. OpenCV \cite{opencv}) requiring integer representation to support Look-Up-Table (LUT) weighting.
As shown in \cref{app:fig::HDR_robertson_cv2} erroneous pixels appear within the solar corona and throughout the histogram in \cref{app:hist::HDR_robertson_cv2}.

\begin{figure}[htb]
    \centering
    \includegraphics[width=\linewidth,keepaspectratio]{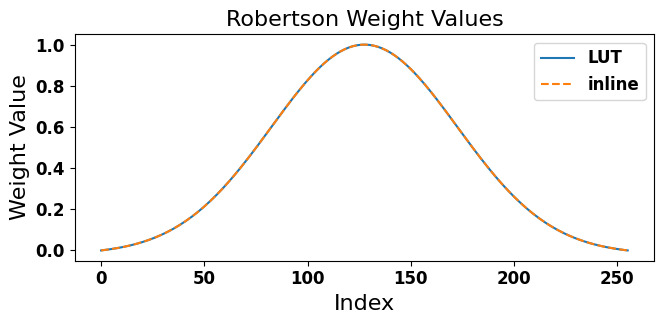}
    \caption{
    Robertson weights by LDR index values ranged [0,255], as computed by OpenCV LUT \cite{opencv} and by our inline implementation through \cref{app:eq::LDR_exposure_fusion_robertson}.
    As shown, the weighting matches exactly.
    }
    \label{app:fig::merge_robertson_inline}
\end{figure}

We found discretization could be mitigated by re-implement the algorithm as an inline function:

\begin{align}
    W_{n} &= \alpha e^{\frac{\psi\Check{I}_{n}}{\lambda}-2} + \beta \\
    \begin{split}
    \hat{I} &= f_{\text{\tiny Robertson}}\left(\left\{\Check{I}_{n},\Delta{t}_n \right\}^N\right) \\
    &= \sum_{n=1}^N \psi\Delta{t}_n W_{n}{\Check{I}_{n}} \Bigg/ \sum_{n=1}^N {\Delta{t}_n^2} W_{n}
    \label{app:eq::LDR_exposure_fusion_robertson}
    \end{split}
\end{align}
Where exposure $\Check{I}_{n}$ is range $[0,1]$, LDR max value is $\psi\!=\!255$, scale is $\alpha\!=\!\frac{e^4}{e^4 -1}$, shift is $\beta=\frac{1}{1-e^4}$, and $\lambda\!=\!(\psi_{LDR})/4 \!=\! 255/4$.
As shown in \cref{app:fig::merge_robertson_inline}, our inline function reproduces the Gaussian-like weighting and in \cref{app:fig::HDR_robertson_ours} is shown to produce a smoother solar corona.
The histogram in \cref{app:hist::HDR_robertson_cv2} also shows a significant reduction of error in image reconstruction.

We note Robertson include a priori assuming a Gaussian-like weighting of pixel values.
This assumes the characteristics of conventional cameras but may be ill-suited to the noise characteristics of DNN-generated imagery.

\clearpage
\section{HDRDB Dataset}
\label{app:dataset}

\begin{figure}[htb]
    \centering
    \includegraphics[width=\linewidth]{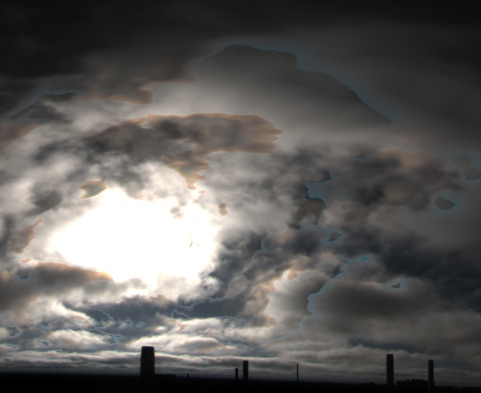}
    \caption{Example of ghosting in HDRDB \cite{LavalHDRdb}, resulting in visual artifacts including discoloration and blurred cloud formations.}
    \label{app:fig::ghosting}
\end{figure}

\begin{figure*}[ht]
    \centering
    \begin{subfigure}{.2\linewidth}
        \centering
        \includegraphics[width=\linewidth]{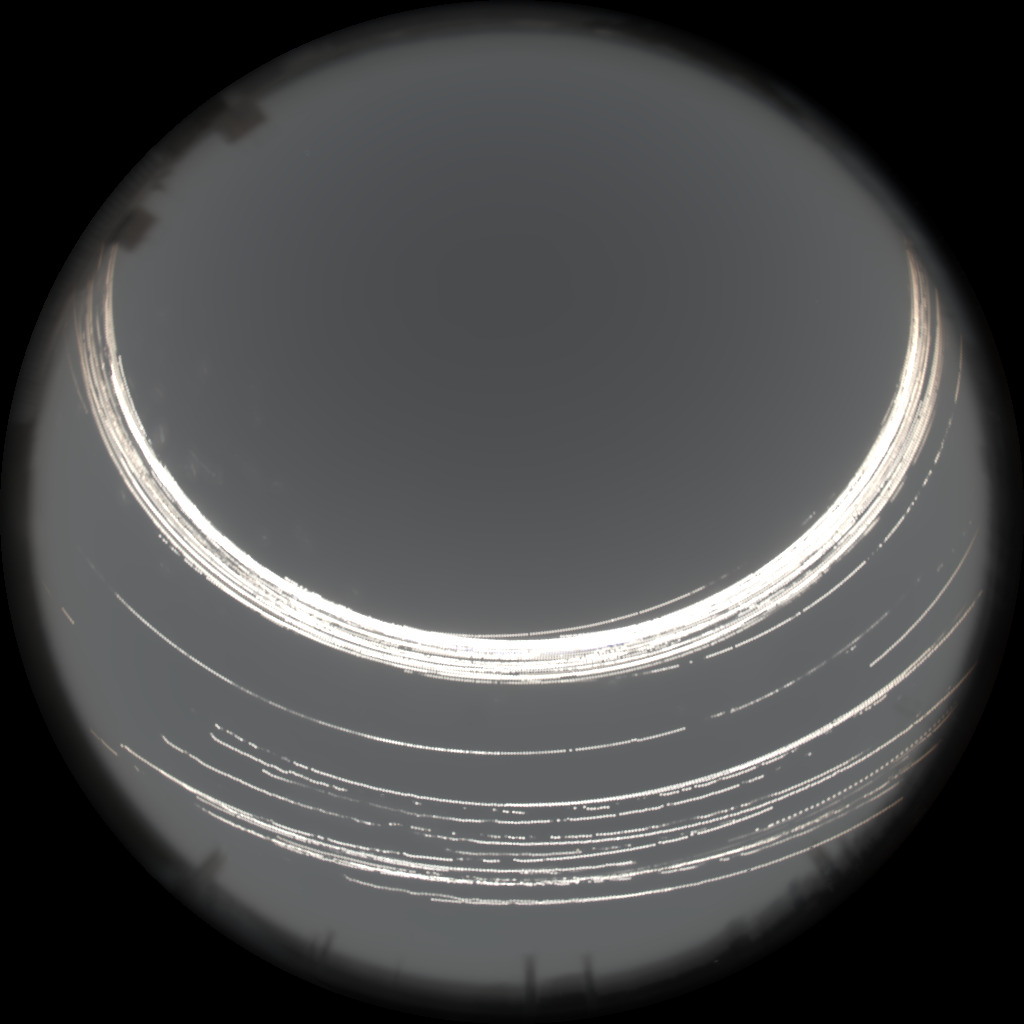}
        \caption{Mean Skydome}
        \label{app:fig::img_mean_skydome}
    \end{subfigure}
    \begin{subfigure}{.2\linewidth}
        \centering
        \includegraphics[width=\linewidth]{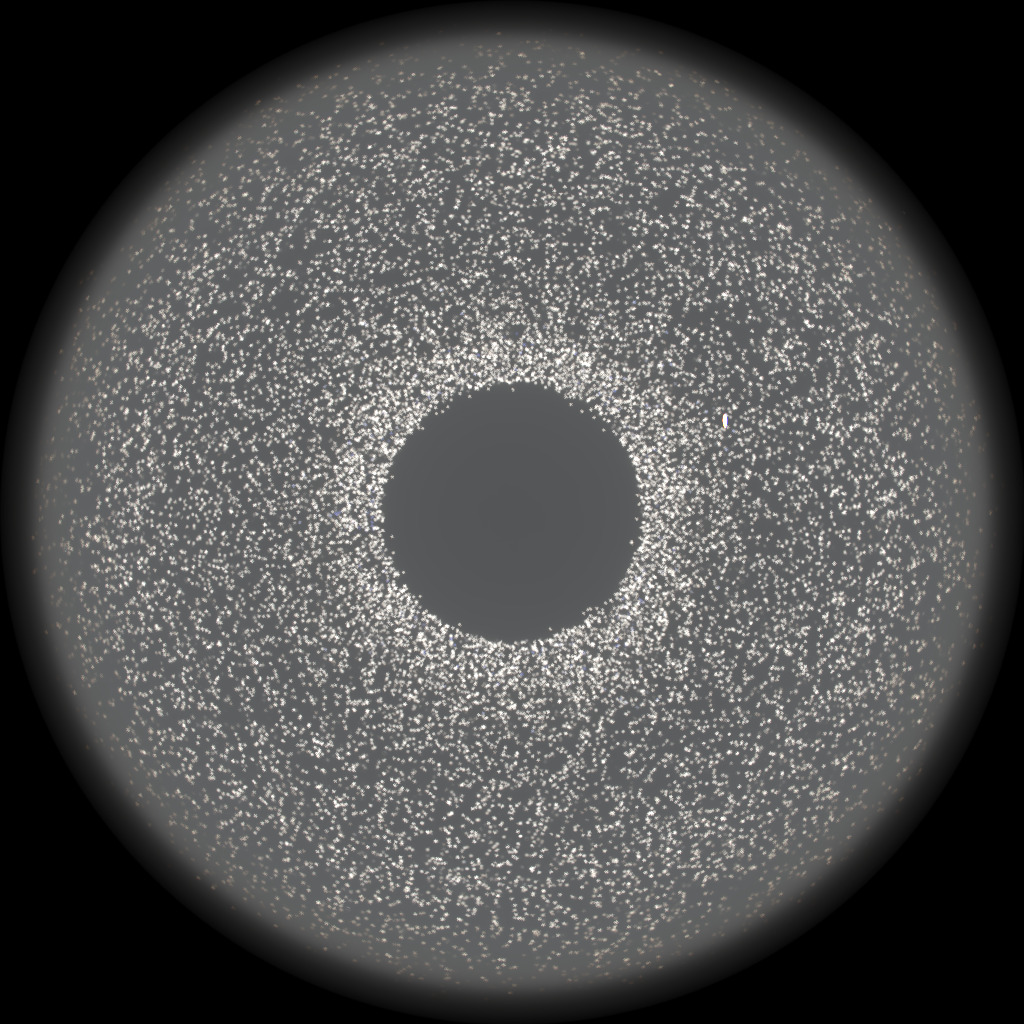}
        \caption{Mean Augmentation}
        \label{app:fig::img_mean_augmented_skydome}
    \end{subfigure}
    \begin{subfigure}{.2\linewidth}
        \centering
        \includegraphics[width=\linewidth]{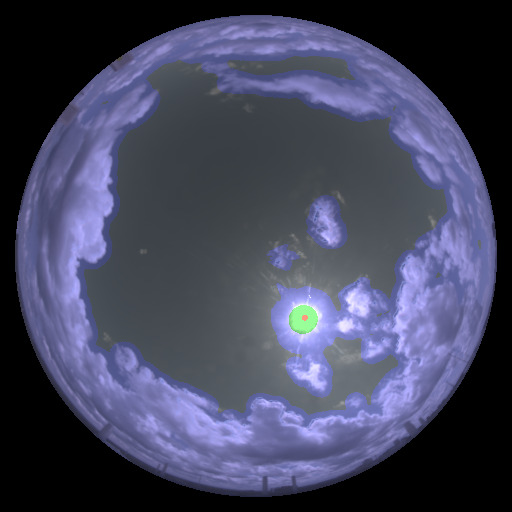}
        \caption{Overlayed segmentation}
        \label{app:fig::img_skydome_overlayed_label}
    \end{subfigure}
    \begin{subfigure}{.2\linewidth}
        \centering
        \includegraphics[width=\linewidth]{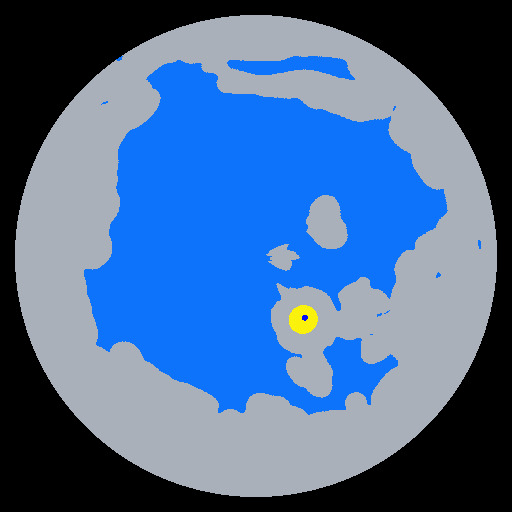}
        \caption{Label}
        \label{app:fig::img_label_disc}
    \end{subfigure}
    \caption{HDRDB mean and mean-augmented skydomes \cite{LavalHDRdb}.
    Labels (\cref{app:fig::img_label_disc}) are 1-channel composites of solar, cloud, skydome (clear-sky), and border masks as proposed by \cite{Ian_towardsSkyModels}.
    Cloud segmentation is produced via colour ratio \cite{clouds_segmentation} and eroded to `hand-drawn' representations
    \Cref{app:fig::img_label_disc} is colourized for visualization purposes.}
\end{figure*}

The Laval HDR Sky database (HDRDB, \cite{LavalHDRdb}) consists of 34K+ HDR images captured in Quebec City, Canada at varied intervals between 2014 and 2016 using a capture method synonymous to that proposed by Stumpfel et al.~\cite{STUMPFEL_HDR_Sky_Capture}.

\subsection{Known Flaws}
\label{app:dataset::flaws}
For the purpose of this work, we ignore the presence of `ghosting' in the dataset.
As illustrated by the example in \cref{app:fig::ghosting}, ghosting is the result of the movement during the capture process which translates to visual artifacts in the recovered FDR imagery.
Without knowledge of wind speed, altitude and other specifics, ghosting is an untractable characteristic of the dataset.

As we do not remove such flawed environment maps, we note that untractable artifacts including `ghosting' impact measures of DNN model performance, prohibiting perfect reconstruction and modeling of all `features' of the ground truth dataset.
For the purpose of adversarial training, we assume discriminators learn to ignore the presence/absence of sporadic visual artifacts including `ghosting'.
We justify this given sporadic artifacts are un-indicative of real or fake classification per a dataset with a fractional subset of HDRI with artifacts.

\begin{figure*}[htb]
    \centering
    \begin{subfigure}{.2\linewidth}
        \centering
        \scalebox{-1}[1]{\includegraphics[width=\linewidth]{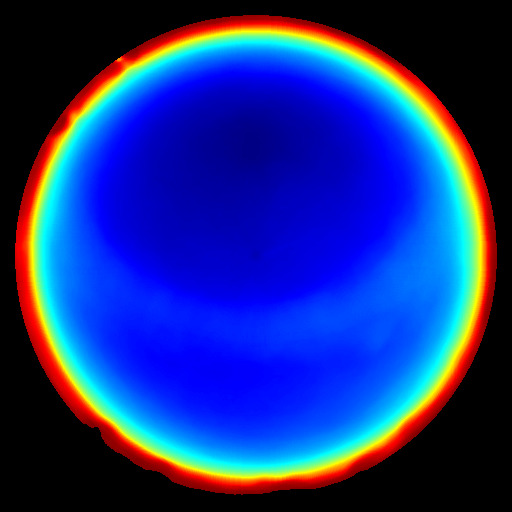}}
        \caption{Skydome (blue sky)}
        \label{app:fig::label_heatmap_skydome}
    \end{subfigure}
    \begin{subfigure}{.2\linewidth}
        \centering
        \scalebox{-1}[1]{\includegraphics[width=\linewidth]{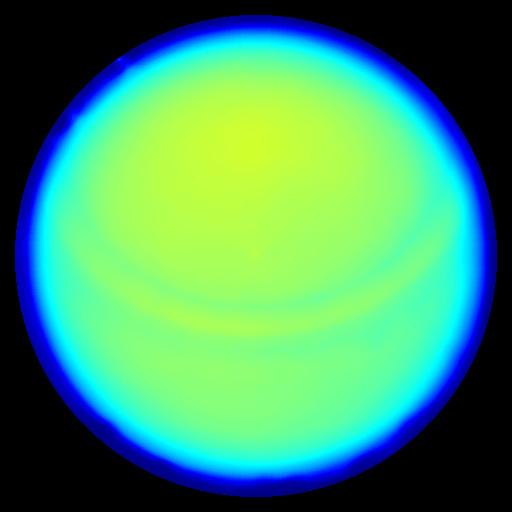}}
        \caption{Clouds}
        \label{app:fig::label_heatmap_clouds}
    \end{subfigure}
    \begin{subfigure}{.2\linewidth}
        \centering
        \scalebox{-1}[1]{\includegraphics[width=\linewidth]{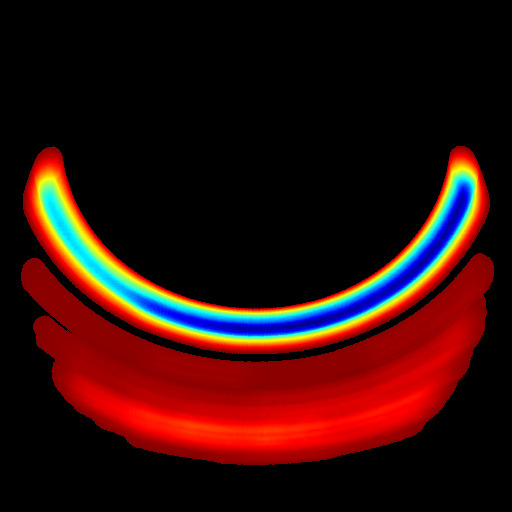}}
        \caption{Solar Corona}
        \label{app:fig::label_heatmap_sun_corona}
    \end{subfigure}
    \begin{subfigure}{.2\linewidth}
        \centering
        \scalebox{-1}[1]{\includegraphics[width=\linewidth]{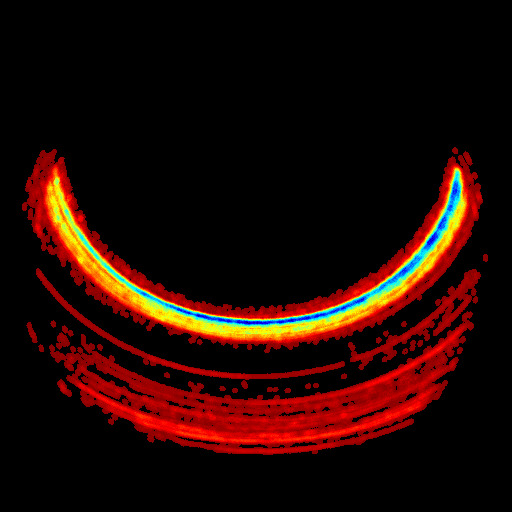}}
        \caption{Solar disk}
        \label{app:fig::label_heatmap_sun_disk}
    \end{subfigure}
    \caption{Class-locality heat-maps for decomposed `hand-drawn' HDRDB Discrete ($\mathbb{Z}^{+}$, \cref{app:fig::img_label_disc}) labels.
    Heat-maps are colored from low- to high-frequency (\textcolor{red}{red} to \textcolor{blue}{blue}), illustrating a disproportional volume of samples from summer months (high-solar elevation) and a near-constant `ring' of clouds around the horizon.
    }
    \label{app:fig::label_heatmaps}
\end{figure*}

Additionally, we note the distribution of samples in HDRDB is flawed temporally and geologically.
As illustrated in \cref{app:fig::label_heatmap_sun_corona,app:fig::label_heatmap_sun_disk}, the bulk of HDRDB's samples were captured during the summer months and favour sunset (west).
As illustrated in \cref{app:fig::label_heatmap_skydome,app:fig::label_heatmap_clouds}, HDRDB's samples are shown to exhibit a `crown of clouds'.
When compared to wind and pressure maps, this appears to be the result of topology and pressure systems guiding cloud formations around Quebec City.

\subsection{Environment Maps}
\label{app:dataset::envmap_format}

For the purpose of this work, we assume that HDRDB latlong physically captured FDR HDRI were correctly captured, calibrated to linear RGB colour space with BT.709 primaries and converted to latlong format without loss.
We assume no artifacts were introduced and no alterations were made to the exposure range or illumination.

\subsection{Pre- and Post-Processing}
\label{app:sec::pre_post_processing}
To prevent discontinuities in generated skydomes, we convert the HDRI from latlong to sky-angular format as defined in \cref{app:dia::dia_skyangular}.
We reproduce the segmentation proposed by AllSky \cite{Ian_towardsSkyModels}, expanding on previously identified challenges with environment map format conversions.

\begin{figure*}[htb]
    \centering
    \begin{subfigure}{.48\linewidth}
        \includegraphics[width=\linewidth,keepaspectratio]{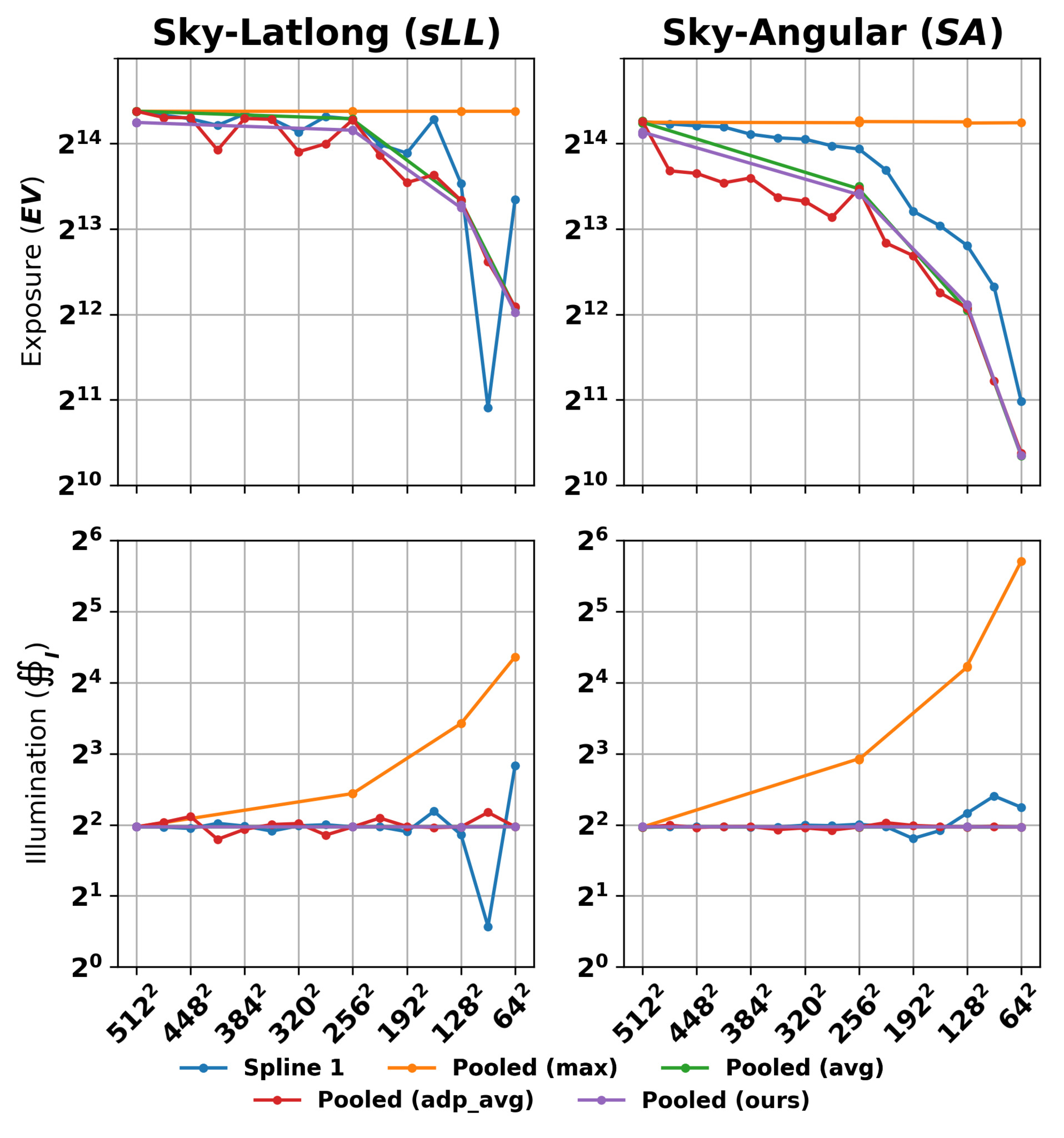}
        \label{fig::scaling_plot:::plot}
    \end{subfigure}
    \begin{subfigure}{.48\linewidth}
        \includegraphics[width=\linewidth,keepaspectratio]{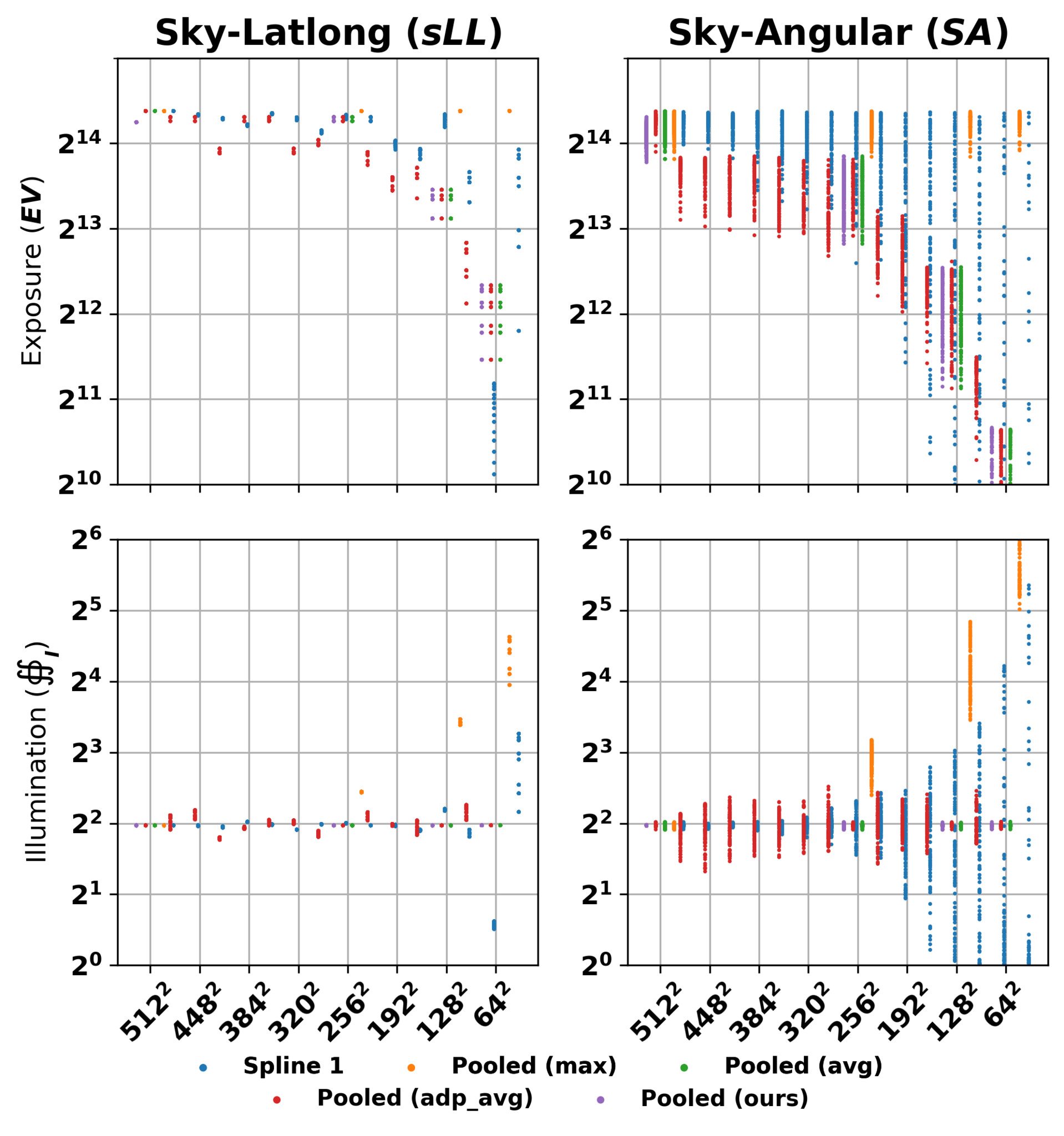}
        \label{fig::scaling_plot:::scatter}
    \end{subfigure}
    \caption{Loss of exposure range (EV) and illumination ($\oiint_I$) in environment map format conversion and downsampling across 100 augmentations of an HDRI from June 7th, 2016 at 12:52PM \cite{LavalHDRdb}.
    \cref{fig::scaling_plot:::plot,fig::scaling_plot:::scatter} Left columns: Downsampling sky-latlong (sLL) environment maps.
    \cref{fig::scaling_plot:::plot,fig::scaling_plot:::scatter} Right columns: Converting and downsampling sky-latlong (sLL) environment maps to sky-angular (SA) format.
    The results show format conversions negatively impact exposure range and illumination, with \cref{fig::scaling_plot:::scatter} demonstrating the instability of output characteristics.
    Though spline (\textit{spline}, order 1) interpolation preserves exposure ranges well in \cref{fig::scaling_plot:::plot}, \cref{fig::scaling_plot:::scatter} demonstrates significant variance in performance.
    Pooling (\text{Pooled}) can provide greater stability in downsampling (lower variance), where \textit{ours} refers \ourModel{}'s implementation of \textit{avg} pooling while assuring format conversions are performed at $2\!\times$ target downsampling resolution.
    Adaptive average pooling (\textit{adp-avg}) extends to offers support for downsampling arbitrarily (non powers of 2), which \ourModel{} mitigates.
    Max pooling (\text{max}) results in steadily increasing environment luminance ($\oiint_I$).
    }
    \label{fig::scaling_plot}
\end{figure*}

\begin{figure}[htb]
    \centering
    \includegraphics[width=\linewidth]{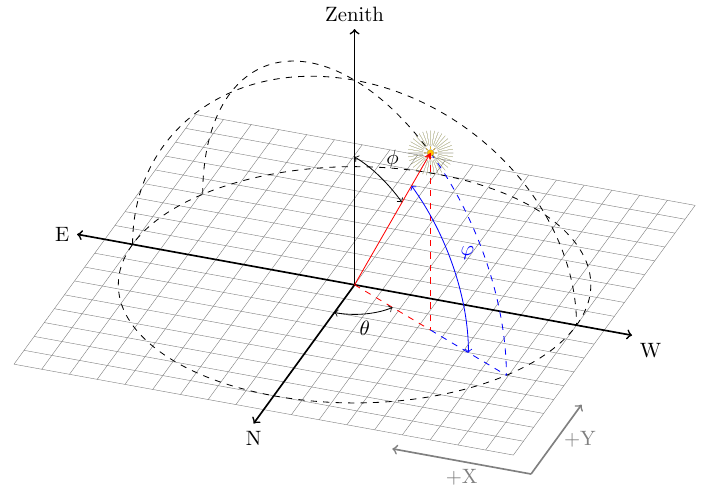}
    \caption{Sky-Angular Environment Map}
    \label{app:dia::dia_skyangular}
\end{figure}

As illustrated in \cref{fig::scaling_plot}, loss of exposure range (EV) and illumination ($\oiint_I$) in downsampling HDRDB Sky-Latlong imagery is exacerbated by conversion to Sky-Angular format.
In experimentation, we observed the findings to be exacerbated further if downsampling and format conversion are completed as a singular uniform transformation.
To mitigate this destabilization of HDRI illumination characteristics, we convert formats at $\geq 2\times$ the target format resolution (inter-area upsampling if required), before inter-area downsampling (average pooling) to the target resolution.
This approach (\textit{ours}) is illustrated in \cref{fig::scaling_plot} and shown to offer reliable downsampling.

\subsection{Minimum Viable Resolution}
\label{app:minimum_viable_resolution}
From an extraterrestrial Point-Of-View (POV), the sun is a $0.5^\circ$ angular-diameter disk with near-constant illumination \cite{seeds_astronomy}.
To accurately model the illumination and shadows cast by the solar disc, environment maps must offer pixel-granularity equal or greater to the angular size of the solar disk.

\begin{equation}
\Omega = 2\pi \left(1-\cos\left({\frac{\theta}{2}}\right)\right)
\label{eq:angularDiameter_to_steradians}
\end{equation}

The Field-Of-View (FOV) of an environment map's pixel is expressed as solid angles in steradians ($\Omega$).
Using \cref{eq:angularDiameter_to_steradians} with the solar disk angular diameter in radians ($\theta$), the solar disk represents $5.98\times10^{-5}$ steradians.
Comparing to the maximum solid angle of common environment map formats, this translates to a minimum resolution of $512^2$ in sky-angular format.
This translates to resolution of $1024\!\times\!2048$ in latlong (equirectangular) format.

\subsection{Subsets and Augmentation}
\label{app:dataset::subset_augmentation}

From the database—the mean of which is shown in \cref{app:fig::img_mean_skydome}—we augment the dataset with random rotations around the zenith to increase solar placement coverage (\cref{app:fig::img_mean_augmented_skydome}) and enable generation of skies outside of HDRDB.
We split the dataset into training (23,249), validation (2,576), and testing (6,634) subsets by arbitrarily splitting by date of capture such that each subset has a random assortment of images from each year and season of capture.
We then prune these subsets to remove images where solar elevation is less than $10^{\circ}$ as the sun is obfuscated (apparent sunset) and the resulting low-light is reflected by poor image quality.
This reduces the aforementioned training (20,064), validation (2,238), and testing (5,811) subset sizes.
For ablation studies, the training subset is further reduced to 4,096 samples while retaining the full validation and testing subsets.

Note, among other, we do not account for the distribution of solar elevations, solar intensities, cloud coverage, wind speeds, or families of cloud formations across subsets.
As shown in \cref{app:fig::label_heatmap_sun_corona,app:fig::label_heatmap_sun_disk}, the bulk of HDRDB's samples were captured during the summer months and favour sunset (west).
As a results, sunrise, sunset, winter and other settings are underrepresented.

\subsection{Segmentation}
\label{app:dataset::segmentation}

We recreate the segmentation of HDRDB proposed by AllSky \cite{Ian_towardsSkyModels} to create `hand-drawn' labels of the environment maps.

Solar positioning is refined from ephemeris calculations \cite{pySolar}, labelling the extraterrestrial 0.5$^{\circ}$ sun as the solar disk.
To reflect terrestrial imagery, we extend the masked solar-region to a diameter of $5^{\circ}$ to include the solar corona and atmospheric attenuation of the extraterrestrial solar disc.

Cloud formations are segmented by thresholding the ratio  $Y = \frac{B-R}{B+R}$ proposed by Dev et al.~\cite{clouds_segmentation} to $\mu$LawLog$_2$ tone mapped HDRI.
Clouds masks are further processed morphologically to reduce complexity and produce `hand-drawn' masks emulating having been drawn with circular brush (kernel size 15).
The final label is a 1-channel composite of solar, cloud, skydome (clear-sky), and border masks as shown in \cref{app:fig::img_label_disc}.

The accuracy of this approach is illustrated by the overlayed label in \cref{app:fig::img_skydome_overlayed_label}.
Note, segmentation exhibits variability with illumination intensity (e.g.\ sunrise/sunset) and seasonality \cite{koehler1991status}.

\subsection{Tone Mapping Operators}
\label{app:dataset::tonemapping}

\begin{figure*}[ht]
        \centering
        \includegraphics[width=0.95\textwidth,keepaspectratio]{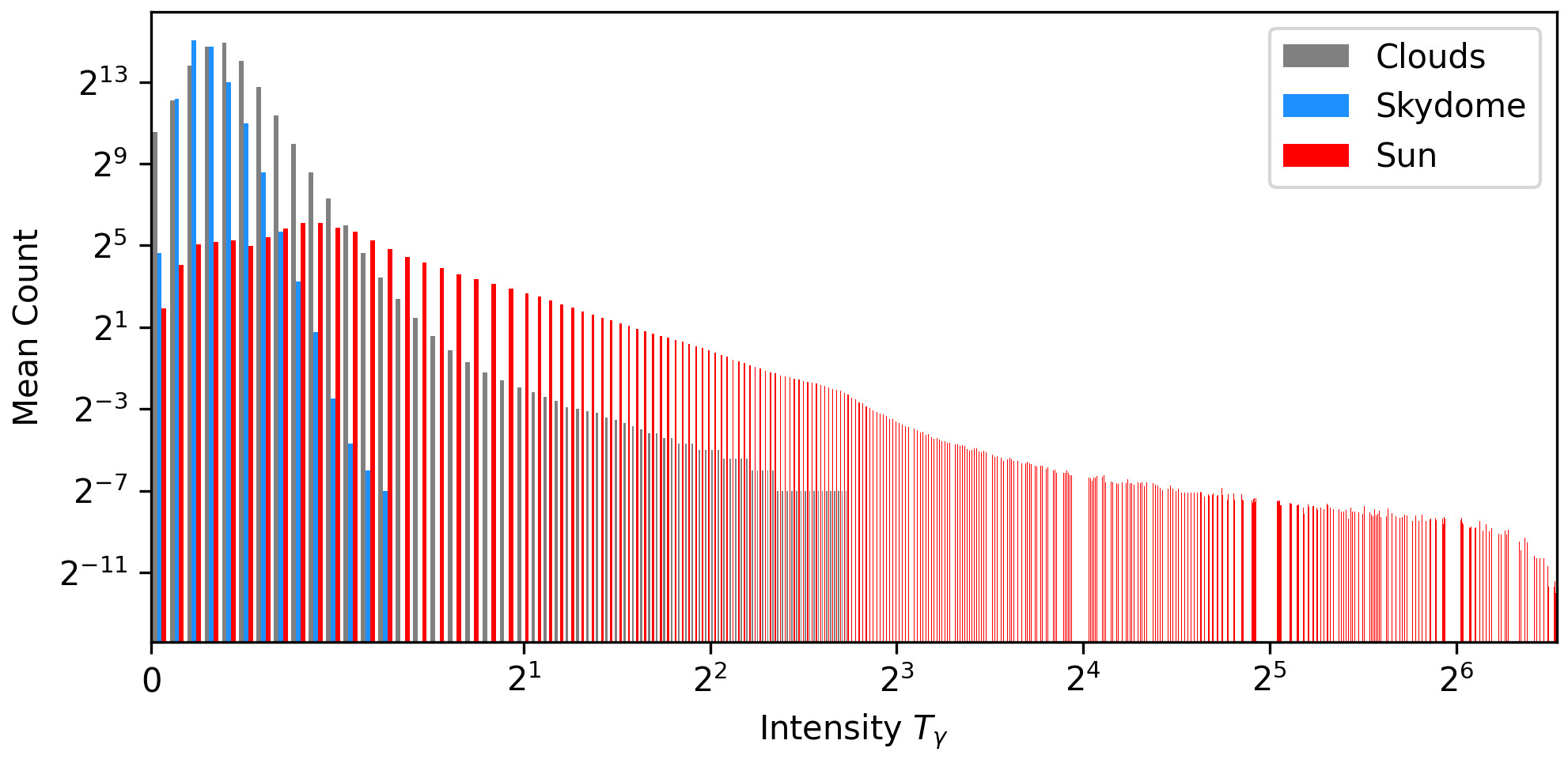}
        \caption{Histogram of Power-Law ($T_\gamma$) tone-mapped HDRDB \cite{LavalHDRdb}}
        \label{app:plt::hdrdb_hist_tm_gamma}
\end{figure*}

\begin{figure*}[ht]
        \centering
        \includegraphics[width=0.95\textwidth,keepaspectratio]{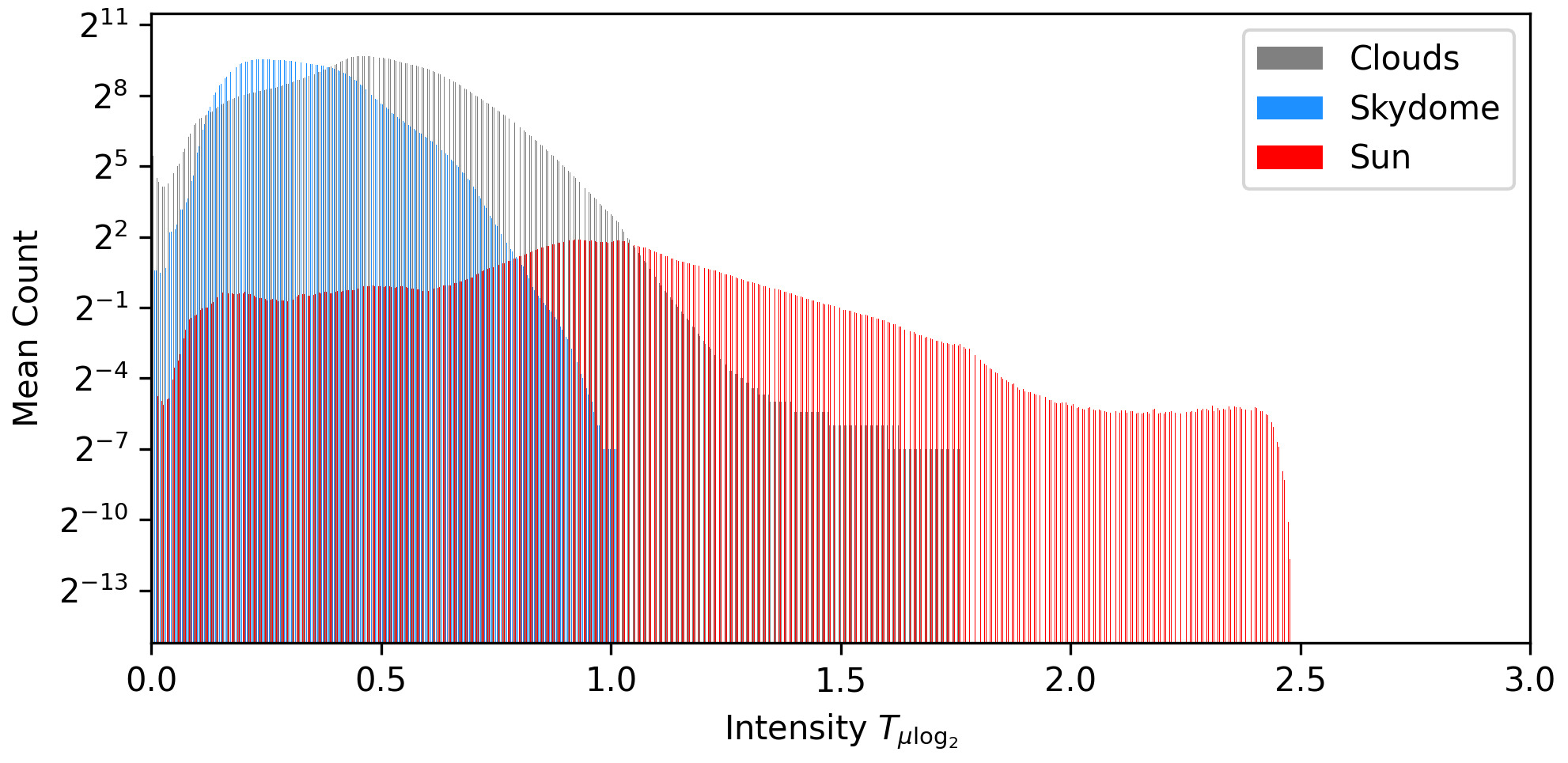}
        \caption{Histogram of $\mu$-lawLog$_2$ ($T_{\mu \log_2}$) tone-mapped HDRDB \cite{LavalHDRdb}}
        \label{app:plt::hdrdb_hist_tm_mixed}
\end{figure*}

For visualization of the impact on LDR-space intensity by tone mapping operators, we include the mean histograms from tone mapping HDRDB (all subsets, 34K+ HDR images) with $T_\gamma$ and $T_{\mu\log_2}$ tone mapping operators in \cref{app:plt::hdrdb_hist_tm_gamma,app:plt::hdrdb_hist_tm_mixed}.

\subsection{Metrics}

Recent works have shown HDR imagery should be evaluated as perceptually uniform (PU) values such that quantization is aligned to human perception of visible differences \cite{PU21}.
We do not implement a perceptually uniform encoding function as HDRDB:
\begin{enumerate*}
\item Is not metrically calibrated for luminance
\item The original physically captured LDR brackets are not available
\item Records of the platform's CRF, intrinsic, colorimetric, photometric and most other calibration are unavailable/lost.
\end{enumerate*}

\clearpage

\begin{figure*}[htb]
    \centering
    \includesvg[width=\linewidth,keepaspectratio]{appendix/diagrams/diagram_SEAN.svg}
    \caption{
        SEAN \cite{SEAN_2020}
    }
    \label{app:fig::dia_SEAN}
\end{figure*}

\section{Models}
\label{app:X_models}

In the following subsections, we expand on our baselines and model variants.

\subsection{SEAN}
\label{app:X_models::SEAN}

We selected SEAN \cite{SEAN_2020} as a proven model for image-translation and configured as shown in \cref{app:fig::dia_SEAN} to train a baseline. 
Although initial experimentation was successful, we found a significant number of errors in the authors' code base, including:

\begin{enumerate}
    \item {
        The dimension of the latent \textit{z}-vector  
        \href{https://github.com/ZPdesu/SEAN/blob/04c7536ff3fecd2d1a09c9ae046a1144636033a5/options/base_options.py#L61}{is set by default to} 
        $z\_dim=256$ in configuration 
        \href{https://github.com/ZPdesu/SEAN/blob/04c7536ff3fecd2d1a09c9ae046a1144636033a5/models/networks/generator.py#L32}{but is fused to} 
        $z\_dim=512$ at  (style encoder) initialization.
    }
    \item {
        Adversarial training
        \href{https://github.com/ZPdesu/SEAN/blob/04c7536ff3fecd2d1a09c9ae046a1144636033a5/models/networks/loss.py#L57}{is non-functional due to a cut gradient} in the hinge-loss.\cref{fn:sean::cut_gradient}
    }
    \item {
        The adversarial Feature Matching Loss does not positively contribute to results 
        \footnote{See \cref{app:X_experiments}}. 
    }
\end{enumerate}

We found that fixing the adversarial hinge loss destabilized training, resulting in frequent collapses \footnote{\label{fn:sean::cut_gradient}
    \href{https://github.com/ZPdesu/SEAN/issues/51}{See open issue on GitHub:}
    \url{https://github.com/ZPdesu/SEAN/issues/51}
}.
We therefore discarded SEAN's discriminator and replaced it with one of our own design, integrating the Adversarial Feature Matching Loss \cite{SEAN_FeatureMatchingLoss}. 

\subsection{SEAN (Ours)}
\label{app:X_models::SEAN_ours}

We create our own implementation of SEAN by refactoring all modules for improved throughput and footprint.
This reduced SEAN's parameters by 25\% (from 266 to 200 million parameters at a resolution of $512^2$) and significantly improves computational performance.  

We redesigned the static generator architecture to be dynamically configurable, allowing for a desired input seed-label shape (e.g.\ $32^2$) or a desired number of upsampling layers.
This change is necessary to accommodate HDRDB semantic labels where key classes (e.g.\ the solar disk) can be lost in seed-label shapes below $32^2$.

\subsection{\ourModel{} (Ours)}
\label{app:X_models::AllSky_SEAN}

We adopt SEAN (Ours) as a backbone architecture and develop a decoder supporting the generation of FDR exposure ranges as LDR exposure brackets.
We characterize the final layer of the architecture as the decoder (`\textcolor{magenta}{latent} $\rightarrow$ \rgb' in \cref{app:fig::dia_SEAN}) and explore different decoder architectures (\ourModel{} decoder in \cref{fig:dia_AllSky_SEAN}). 

\begin{figure}[htb]
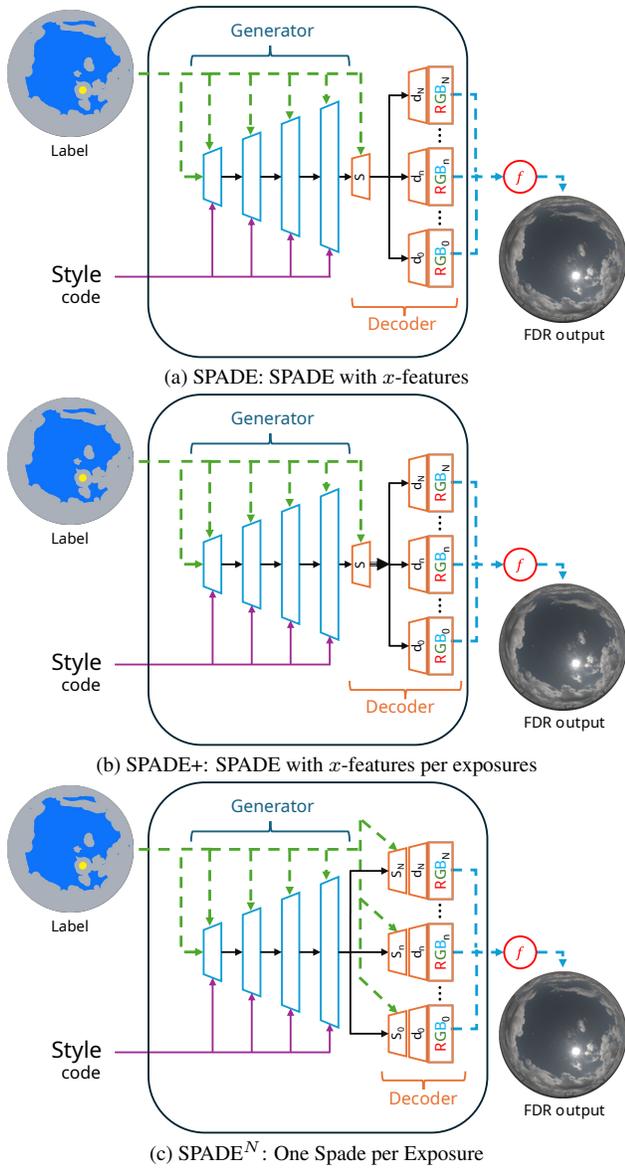

    \centering
        \begin{subfigure}{\linewidth}
        \centering
        \includesvg[width=\linewidth]{appendix/diagrams/diagram_AllSky_v2_decoder_SE.svg}
        \caption{SPADE: SPADE with $x$-features}
        \label{app:dia::decoder_SE}
    \end{subfigure}
        \begin{subfigure}{\linewidth}
        \centering
        \includesvg[width=\linewidth]{appendix/diagrams/diagram_AllSky_v2_decoder_SnE.svg}
        \caption{SPADE+: SPADE with $x$-features per exposures}
        \label{app:dia::decoder_SnE}
    \end{subfigure}
    \begin{subfigure}{\linewidth}
        \centering
        \includesvg[width=\linewidth]{appendix/diagrams/diagram_AllSky_v2_decoder_nSnE.svg}
        \caption{SPADE$^N$: One Spade per Exposure}
        \label{app:dia::decoder_nSnE}
    \end{subfigure}
    \caption{Decoder architectures for LDR bracket generation, exploring configurations of SPADE (S) blocks and their output features. In \cref{app:dia::decoder_SnE}, SPADE+ is configured such that SPADE outputs n-features per exposure as indicated by ${\Rrightarrow}$ instead of $\rightarrow$. }
\end{figure}

\subsubsection{Decoder: SPADE}
We implement a simple variant which replaces the output convolution layer (d) with n-parallel convolutional layers ($d_0,d_n,\dots,d_N$). 
As illustrated in \cref{app:dia::decoder_SE,app:dia::decoder_SnE}, we explore SPADE (S,\cite{SEAN_2020}) output features, passing them uniformly ($\rightarrow$; denoted as SPADE) and equally-split ($\Rrightarrow$, grouping by n-exposures; denoted as SPADE+) to n-parallel convolution layers.  

\subsubsection{Decoder: n-SPADE}
As illustrated in \cref{app:dia::decoder_nSnE}, we explore a larger architecture which implemented n-SPADE blocks ($S_0,S_n,\dots,S_N$;  denoted as SPADE$^3$). This allowed for independent modulation of features for each exposure of a LDR exposure bracket.

\subsubsection{Decoder: Additional Configuration Details}
We found enabling affine synchronized batch normalization in the SPADE blocks of the decoder improved performance.
We note that synchronous training of all decoders with decayed exposure facilitates training and provides better visual results. 
Due to the limited features shared between low and high exposures, exposure decay (see \cref{sec:methodology::model::decoder}) aligns features and mitigates collapse.
Asynchronous incremental training of the model is precarious as effort must be taken to ensure visual quality of $\Check{I}_0$ is not lost/degraded in training higher exposures $\Check{I}_{k>0}$. 
We note that a model trained to generate $\Check{I}_0$ does not generalize well to $\Check{I}_{k>0}$ and therefore recommend the generator (G) not be frozen during training of higher exposures.

\subsubsection{\rgb-Style Encoder Augmentation}
We note that supervised losses are possible with SEAN as the model learns to reconstruct the input per segmentation and style-codes distilled from paired ground truth imagery.
This entanglement between input and generated imagery is undesirable, but cannot be fully mitigated given the contents of environment maps are not necessarily translatable (e.g.\ an overcast sun presumably cannot be substituted for a clear-sky's sun).  
To mitigate and discourage entanglement with ground truth imagery, we augment the input to the style-encoder with flips and rotations around the zenith. 
We further $T_\gamma$ tone-map and clip style-encoder input to LDR space. 

\subsubsection{\rgb-Style Code Limitations}

In training, style encoder inputs are independently augmented, but we do not address the reconstruction enabled from paired segmentation, style-codes, and ground truth imagery.
This is particularly difficult, given the architecture is ignorant of lens calibration and thus does not distill the same style-code for the same texture sampled at the horizon and at the zenith. 
As such, not all styles are translatable from one segmentation to another (e.g.\ clouds at the horizon to clouds at the zenith).

To address this issue, a methodology for sampling textures as style codes independent of positioning would need to be developed.
Addressing this issue was determined to be outside the scope of the contributions of this work, but is believed to be feasible through a study of style-codes.
This would involve undistorting textures while ensuring a uniform granularity of sampled textures. 

\subsubsection{Independently Training The \rnd-Style Mapper}
Our default training routine for \ourModel{} is oriented around uniformly training of the generator per a singular source of style-codes.
If \ourModel{}'s generator is trained with the \rgb-style encoder, it is possible for the \rnd-style mapper to learn \rgb-style encoder's distribution of codes. 
This can be achieved by iteratively decaying the weight of per-exposures losses per \cref{eq:exp_decay} for a simulated incremental training of exposures.

Additionally, we explored the mixing of \rgb- and \rnd- styles during training. 
Though we do not report this work, we found that passing subsets of the \rgb-style codes through the mapper can aid in training the mapper. 
This suggests a complementary training routine could be developed which uniformly trains \ourModel{} with both the \rgb-style encoder and \rnd- style mapper.

\subsubsection{Independently Training The \rgb-Style Encoder}
We do not report training the \rgb-style encoder per a frozen generator trained with the \rnd-style mapper, but the encoder and mapper are fully interchangeable.

\subsection{Fusion}
We decompose fusion with the objective of mitigating the priori of a Gaussian-like weighting of pixel values.
We achieve this by proposing $f_{\text{\tiny DNN}}$ in \cref{eq:LDR_exposure_fusion} to predict content-aware weights for each exposure of an LDR bracket $\left\{\check{I}_n\right\}^N$.

As shown in \cref{app:dia::fusion_DNN}, we experiment with two architectures which enable class-aware modulation of $\left\{\check{I}_n\right\}^N$ via a SPADE \cite{SEAN_2020} per $\left\{\check{I}_n\right\}^N$ input or generator latent $\xi$ input.
In \cref{app:dia::fusion_DNN_SnE_W}, a weight $\left\{W_{n}\right\}^N$ is generated to be uniformly applied to LDR exposures $\left\{\Check{I}_n\right\}^N$ and exposures times $\left\{\Delta t_n\right\}^N$ as shown in \cref{app:eq::LDR_exposure_fusion_DNN_W}.
In \cref{app:dia::fusion_DNN_SnE_WtWe}, separate weights $\left\{W^{\Delta t}_{n}\right\}^N$ and $\left\{W^e_{n}\right\}^N$ are generated for LDR exposures $\left\{\Check{I}_n\right\}^N$ and exposures times $\left\{\Delta t_n\right\}^N$ as shown in \cref{app:eq::LDR_exposure_fusion_DNN_WtWe}.
Additionally, we explore passing SPADE ($S$)  features uniformly ($\rightarrow$) and equally-split ($\Rrightarrow$, grouped by exposure) to convolutional layers.  

\begin{align}
    \hat{I} &= f_{\text{\tiny DNN}}\left(\left\{\Check{I}_{n},\Delta{t}_{n}\right\}^N\right) \\
    &= \sum_{n=1}^N \Delta{t}_{n} W_{n}{\Check{I}_{n}} \Bigg/ \sum_{n=1}^N {\Delta{t}_{n}^2} W_{n}
    \label{app:eq::LDR_exposure_fusion_DNN_W} \\
    &= \sum_{n=1}^N \Delta{t}_{n} W^e_{n}{\Check{I}_{n}} \Bigg/ \sum_{n=1}^N {\Delta{t}_{n}^2} W^{\Delta t}_{n}
    \label{app:eq::LDR_exposure_fusion_DNN_WtWe}
\end{align}

% Chromatic aberration with II and pII
\begin{figure}[htbp]
    \centering
    \begin{subfigure}{\linewidth}
        \centering
        \includesvg[width=\linewidth]{appendix/diagrams/diagram_AllSky_v2_fusion_SnE_Wncij.svg}
        \caption{Uniform Weights $W^e_{n}=W^{\Delta t}_n=W_{n,c,ij}$.}
        \label{app:dia::fusion_DNN_SnE_W}
    \end{subfigure}
    \begin{subfigure}{\linewidth}
        \centering
        \includesvg[width=\linewidth]{appendix/diagrams/diagram_AllSky_v2_fusion_SnE_WnWt.svg}
        \caption{Separate weights $W^e_{n,c,ij}$ and $W^{\Delta t}_{n,c,ij}$.}
        \label{app:dia::fusion_DNN_SnE_WtWe}
    \end{subfigure}
    \caption{Architectures for LDR bracket Fusion. 
    In \cref{app:dia::fusion_DNN_SnE_W}, a weight $W_{n,c,ij}$ is generated to be uniformly applied to LDR exposures $\left\{\Check{I}_n\right\}^N$ and exposures times $\left\{\Delta t_n\right\}^N$.
    In \cref{app:dia::fusion_DNN_SnE_WtWe}, a separate weight  $W^e_{n,c,ij}$ is generated to module exposures $\left\{\Check{I}_n\right\}^N$ and a $W^{\Delta t}_{n,c,ij}$ to modulate exposures times $\left\{\Delta t_n\right\}^N$.
    Features output by SPADE (${\Rrightarrow}$) are configurable to output n-features or n-features per exposure.}
    \label{app:dia::fusion_DNN}
\end{figure}
\begin{figure*}[htbp]
    \centering
    \includegraphics[height=.9\textheight,keepaspectratio]{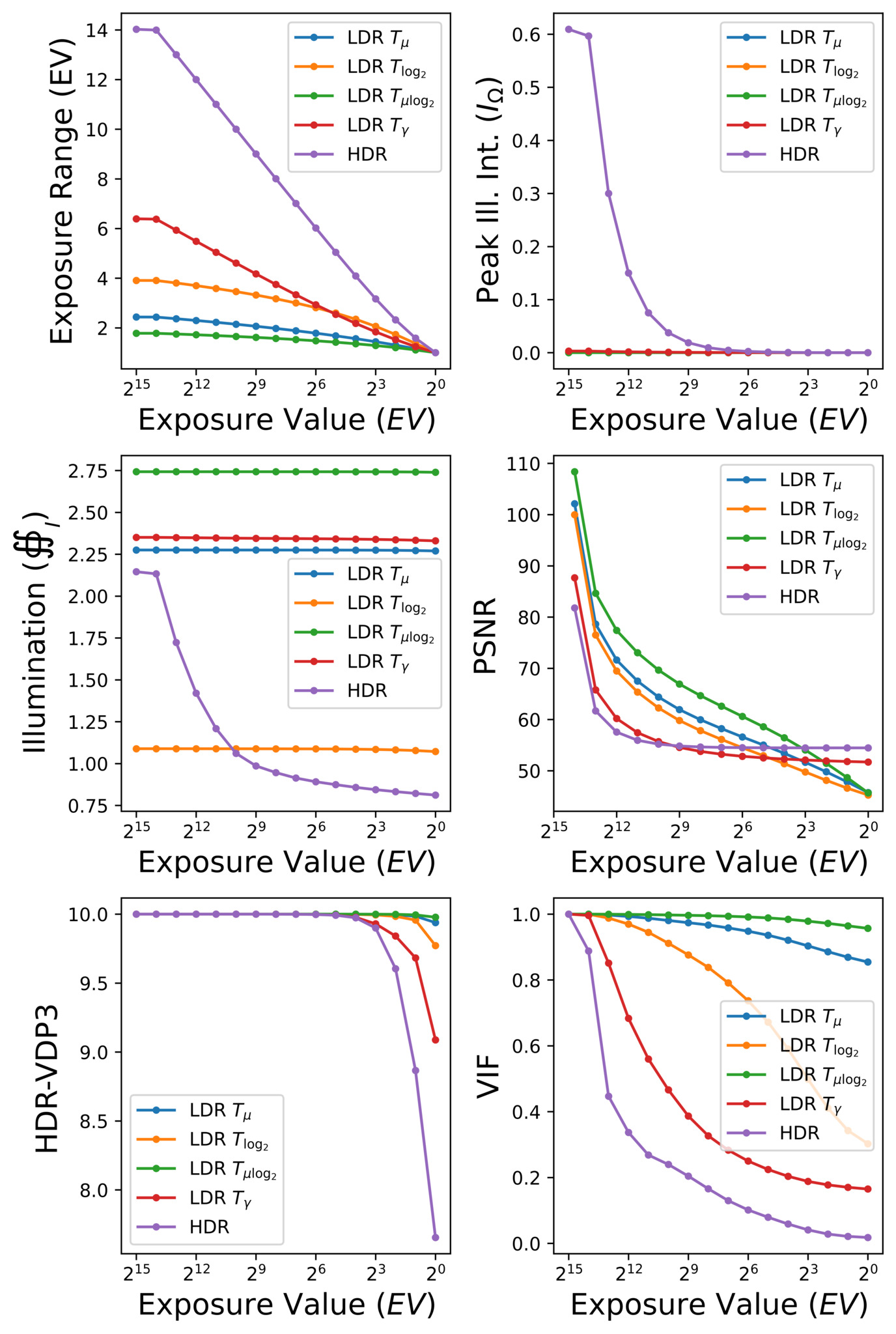}
    \caption{Sensitivity of various commonly reported HDR metrics to truncation of the exposure range.
    As shown, Integrated Illumination ($\oiint_I$, \cref{eq:oiint_I}), Peak-Luminance ($PL_\Omega$ \cref{eq:PeakRandiantIntensity}, Peak Ill. Int.$I_\Omega$), and VIF \cite{VIF} offer sensitivity to truncations of HDR-space exposure.
    Though PSNR offers sensitivity in LDR-space, the metric offers little sensitivity in HDR-space.
    HDR-VDP3 offers little sensitivity to truncations of exposure in both LDR- and HDR-space.
    }
    \label{fig:placeholder}
\end{figure*}

\clearpage 

\begin{table*}[htb]
\centering
\caption{
    The results show that training with the Discriminator Feature Matching loss (indicated by \textcolor{red}{\textbf{*}}) negatively impacts cLDR metrics with higher FID and MiFID scores. 
    Though $\mu\text{-lawLog}_2$ ($T_{\mu \log_2}$) tone mapping generally offers better cLDR metric performance, FDR performance is variable and prone to over-exposure ($\oiint_I$). 
    Gamma ($T_{\gamma}$) tone mapping offers similar cLDR performance with stable exposure range (EV) and illumination ($\oiint_I$).
    Average training time is reduced by over 50\%, from 9hr 15min for SEAN to 4hr 25min for SEAN (Ours) with no loss in model performance.
    \textbf{Bold} values indicate category best and \underline{underlined} values indicate overall best. 
}
\label{app:tab::SEAN}
\begin{tabular}{r|c|c|ccc|cccc|}
    \multicolumn{2}{c|}{} & \textbf{System} & \multicolumn{3}{|c|}{\textbf{$T_{\gamma}$-cLDR}} & \multicolumn{4}{c|}{\textbf{HDR}} \\
    \cline{3-10} 
     & $T_{m}$ & Time & LPIPS $\downarrow$  & FID $\downarrow$  & MiFID $\downarrow$
    & HDR-VDP3 $\uparrow$ & PSNR$_{\log_2}$ $\uparrow$ & EV $\leftarrow$ & $\oiint_I \leftarrow$ \\ 

\hline
Ground Truth $128^2$         & - & - & - & - & - & - & - & 7.34 & 1.22 \\
\hdashline 
SEAN\textcolor{red}{\textbf{*}} &$T_\gamma$              & 8:31 & 0.16 & \textbf{19.6} & \textbf{303} & \textbf{8.29} & \textbf{140} & 6.94 & \textbf{1.04}  \\ % 6
SEAN\textcolor{red}{\textbf{*}} &$T_{\mu \log_2}$        & 8:22 & 0.16 & 21.6 & 322 & 7.96 & 115 & \underline{\textbf{7.27}} & 3.38  \\ % 2
\hdashline 
SEAN &$T_\gamma$             & 9:28 & 0.16 & 20.0 & 306 & \textbf{8.25} & \textbf{138} & \textbf{7.16} & \textbf{1.08} \\ % 5
SEAN &$T_{\mu \log_2}$       & 10:25 & 0.16 & \textbf{18.5} & \textbf{292} & 8.04 & 123 & 6.49 & 1.6$e^6$ \\ % 1
\midrule
SEAN\textcolor{red}{\textbf{*}} (Ours) &$T_\gamma$       & 5:06 & 0.15 & 27.9 & 385 & \textbf{\underline{8.48}} & \textbf{140} & \textbf{7.15} & \textbf{\underline{1.16}} \\ % 8
SEAN\textcolor{red}{\textbf{*}} (Ours) &$T_{\mu \log_2}$ & 4:24 & \textbf{\underline{0.14}} & \textbf{ 20.9} & \textbf{320} & 8.25 & 125 & 6.28 & 0.98 \\ % 4
\hdashline 
SEAN (Ours) &$T_\gamma$      & 4:17 & 0.15 & 17.6 & 275 & \textbf{8.46} & \textbf{138} & \textbf{7.03} & \textbf{\underline{1.16}} \\ % 7
SEAN (Ours) &$T_{\mu \log_2}$& 3:56 & 0.15 & \textbf{16.8} & \textbf{269} & 8.08 & 124 & 5.93 & 0.93 \\ % 3
\hline
Ground Truth $256^2$                       & - & - & - & - & - & - & - & 8.58 & 1.22 \\
\hdashline 
SEAN\textcolor{red}{\textbf{*}} (Ours) &$T_\gamma$        & - & 0.18 & 22.3 & 348 & \textbf{8.28} & \underline{\textbf{159}} & 8.81 & \textbf{1.12} \\ % 9
SEAN\textcolor{red}{\textbf{*}} (Ours) &$T_{\mu \log_2}$ & - & 0.18 & \textbf{20.9} & \textbf{334} & 8.20 & 138 & \textbf{8.25} & 1.05 \\ % 11
\hdashline 
SEAN (Ours) &$T_\gamma$                 & - & 0.18 & 19.9 & 322 & \textbf{8.28} & \underline{\textbf{159}} & \textbf{8.44} & \textbf{1.02} \\ % 13
SEAN (Ours) &$T_{\mu \log_2}$           & - & 0.18 & \textbf{19.0} & \textbf{308} & 8.00 & 127 & 8.97 & 3.34 \\ % 15
\hline
Ground Truth $512^2$                       & - & - & - & - & - & - & - & 9.62 & 1.22 \\
\hdashline 
SEAN\textcolor{red}{\textbf{*}} (Ours) &$T_\gamma$       & - & 0.21 & 53.8 & 662 & 7.32 & \textbf{152} & \textbf{7.84} & 0.72 \\ % 12
SEAN\textcolor{red}{\textbf{*}} (Ours) &$T_{\mu \log_2}$       & - & 0.21 & \textbf{45.2} & \textbf{587} & \textbf{7.59} & 147 & 4.83 & \textbf{0.87} \\ % 10
\hdashline 
SEAN (Ours) &$T_\gamma$                 & - & 0.36 & 186.4 & - & \textbf{4.43} & - & \textbf{24.86} & \textbf{$3e^5$} \\ % 14
SEAN (Ours) &$T_{\mu \log_2}$           & - & \textbf{0.28} & \textbf{144.2} & - & 2.62 & - & 118.72 & $1e^{32}$ \\ % 16
\bottomrule
\end{tabular}
\end{table*}

\section{Experiments}
\label{app:X_experiments}

In the following sections, we expand on our reported results to include more detailed discussion of our baselines and experimentation.

\subsection{SEAN}
\label{app:X_experiments::SEAN}

We train SEAN \cite{SEAN_2020} and SEAN (Ours) at a resolution of $128^2$ to demonstrate the equivalence of the models. 
As shown in \cref{app:tab::SEAN}, SEAN (Ours) achieves similar or better cLDR and FDR performance while reducing the computational expense by over 50\%.

At low resolution, the results in \cref{app:tab::SEAN} demonstrate that SEAN's Discriminator-Feature Matching loss negatively impacts cLDR performance with higher FID and MiFID scores.
As shown through lost performance as resolution is doubled from $128^2$ to $512^2$, SEAN (Ours) is negatively impacted by increasing exposure range.
This coincides with findings by AllSKy \cite{Ian_towardsSkyModels} which identified that increasing resolution results in the solar region becoming the predominant means of differentiation at higher resolutions.
Though quantitatively the Discriminator Feature Matching loss mitigates numerical overflow at a resolution of $512^2$, the quality of the generated images is undesirable. 
Given the negative impact to cLDR performance at low exposure ranges (implicit per low resolution) and poor FDR performance, we elected to omit SEAN's Discriminator Feature Matching loss in later experimentation.

Handling of exposure range is crucial to the performance of DNN sky-models.
Tone mapping provides a means of mitigating, but AllSky \cite{Ian_towardsSkyModels} found tone mapping introduces non-linearity between error in LDR and HDR space, resulting in unstable illumination.
This finding is reflected in \cref{app:tab::SEAN}, with aggressive $\mu\text{-lawLog}_2$ ($T_{\mu \log_2}$) tone mapping offering better cLDR metric performance but variable FDR performance prone to over- and under-exposure ($\oiint_I$). 
As a result, we conclude weaker Gamma ($T_{\gamma}$) tone mapping fosters greater stability in model performance.
% \clearpage

\begin{table*}[htb]
\centering
\caption{
    We ablate \ourModel{} to ascertain the impact of multiple exposures on visual quality by training for a singular exposure $[2^0]$ ($1\!\times\!f_{\text{\tiny \rgb}}$) and three exposures $[2^0,2^0,2^0]$ ($3\!\times\!f_{\text{\tiny \rgb}}$).
    As in \cref{app:tab::SEAN}, the results show that training with the Discriminator Feature Matching loss (indicated by \textcolor{red}{\textbf{*}}) negatively impacts cLDR metrics with higher FID and MiFID scores.
    Aggressive $\mu\text{-lawLog}_2$ ($T_{\mu \log_2}$) tone mapping offers better cLDR metric performance and is shown to retain a larger exposure range in FDR metrics.
    As denoted by $\textcolor{red}{^\dagger}$, exposures are clipped in LDR-space, crippling FDR performance.
    The results show that overlapping exposures negatively impact cLDR performance by introducing noise. 
    This is reflected in lost performance in repeating model evaluation with \rgb- ($3\!\times\!f_{\text{\tiny \rgb}}$) and \hsv-fusion ($3\!\times\!f_{\text{\tiny \hsv}}$), where \hsv-fusion sustains performance by retaining chroma only from the first exposure.
    \textbf{Bold} values indicate category best and \underline{underlined} values indicate overall best. 
}
\label{app:tab::AllSky_baseline}
\begin{tabular}{r|c|ccc|cccc|}
    \multicolumn{2}{c}{} & \multicolumn{3}{c}{\textbf{$T_{\gamma}$-cLDR}} & \multicolumn{4}{c}{\textbf{HDR}} \\
    \cline{3-9} 
    \multicolumn{1}{c}{} & $T_{m}$ & LPIPS $\downarrow$  & FID $\downarrow$  & MiFID $\downarrow$
    & HDR-VDP3 $\uparrow$ & $\textcolor{red}{^\dagger}$PSNR$_{\log_2}$ $\uparrow$ & $\textcolor{red}{^\dagger}$EV $\leftarrow$ & $\textcolor{red}{^\dagger}$$\oiint_I \leftarrow$ \\ 
    
\hline
Ground Truth $128^2$                        & - & - & - & - & - & - & 7.34 & 1.22 \\
\midrule
\ourModel{}\textcolor{red}{\textbf{*}} (SPADE$^n$; $1\!\times\!f_{\text{\tiny \rgb}}$)  &$T_\gamma$                &  \underline{0.14} & \textbf{16.0} & \underline{\textbf{255}} & \textbf{7.7} & 124 & 1.57 & 0.60 \\ % 1
\ourModel{}\textcolor{red}{\textbf{*}} (SPADE$^n$; $1\!\times\!f_{\text{\tiny \rgb}}$) &$T_{\mu \log_2}$          &\underline{0.14} & 19.1 & 307 & \textbf{7.8} & 124 & \textbf{2.91} & \textbf{0.68} \\ % 2
\hdashline 
\ourModel{} (SPADE$^n$; $1\!\times\!f_{\text{\tiny \rgb}}$) &$T_\gamma$                 & \underline{0.14} & 19.0 & 306 & 7.8 & 122 & 1.72 & \underline{\textbf{0.77}} \\ % 5
\ourModel{} (SPADE$^n$; $1\!\times\!f_{\text{\tiny \rgb}}$) &$T_{\mu \log_2}$           & \underline{0.14} & \underline{\textbf{15.7}} & \textbf{260} & \underline{\textbf{7.9}} & \textbf{124} & \textbf{3.30} & 0.69 \\ % 6
\hdashline 
\ourModel{}\textcolor{red}{\textbf{*}} (SPADE$^n$; $3\!\times\! f_{\text{\tiny \rgb}}$) &$T_\gamma$        & 0.15 & 24.1 & 377 & 7.7 & 123 & $1.62$ & $0.65$ \\ % 3
\ourModel{}\textcolor{red}{\textbf{*}} (SPADE$^n$; $3\!\times\!f_{\text{\tiny \rgb}}$) &$T_{\mu \log_2}$  & \underline{0.14} & \textbf{21.2} & \textbf{335} & \underline{\textbf{7.9}} & \textbf{124} & \textbf{3.17} & \textbf{0.67} \\ % 4
\hdashline 
\ourModel{} (SPADE$^n$; $3\!\times\!f_{\text{\tiny \rgb}}$) &$T_\gamma$         & \underline{0.14} & 25.2 & 383 & 7.8 & 123 & $1.60$ & \textbf{0.63} \\ % 7
\ourModel{} (SPADE$^n$; $3\!\times\!f_{\text{\tiny \rgb}}$) &$T_{\mu \log_2}$   & \underline{0.14} & \textbf{21.8} & \textbf{350} & 7.8 & \textbf{124} & $\mathbf{2.81}$ & $0.62$ \\ % 8
\hdashline 
\ourModel{}\textcolor{red}{\textbf{*}} (SPADE$^n$; $3\!\times\!f_{\text{\tiny \hsv}}$) &$T_\gamma$         & 0.15 & 18.4 & 299 & 7.6 & 142 & $1.64$ & $0.65$ \\ % 3
\ourModel{}\textcolor{red}{\textbf{*}} (SPADE$^n$; $3\!\times\!f_{\text{\tiny \hsv}}$) &$T_{\mu \log_2}$   & \underline{0.14} & \underline{\textbf{15.7}} & \textbf{262} & \textbf{7.8} & 142 & \underline{$\mathbf{3.42}$} & $\mathbf{0.67}$ \\ % 4
\hdashline 
\ourModel{} (SPADE$^n$; $3\!\times\!f_{\text{\tiny \hsv}}$) &$T_\gamma$          & \underline{0.14} & 18.9 & \textbf{304} & 7.7 & \underline{\textbf{143}} & $1.6$ & $0.61$ \\ % 7
\ourModel{} (SPADE$^n$; $3\!\times\!f_{\text{\tiny \hsv}}$) &$T_{\mu \log_2}$    & \underline{0.14} & \textbf{18.6} & 307 & 7.7 & \underline{\textbf{143}} & $\mathbf{2.99}$ & $\mathbf{0.63}$ \\ % 8
\bottomrule
\end{tabular}
\end{table*}

% \FloatBarrier
\subsection{\ourModel{} with \rgb-Style Encoder}
\label{app:X_experiments::AllSky_V2}

In \cref{app:tab::AllSky_baseline}, we ablate \ourModel{} to ascertain the impact of generating multiple exposures. 
The results show that overlapping exposures introduce noise in fused HDR images which is reflected by a loss in cLDR metric performance. 
The impact of this noise is demonstrated through the gap between {\rgb}- and {\hsv}-fusion performance, where \hsv-fusion sustains performance by retaining chroma only from the first exposure.
As shown by comparison of SPADE$^n$ variants in \cref{app:tab::AllSky_baseline_SPADE_cohesion}, this performance gap is mitigatable by limiting exposure overlap.

We expand on this ablation in \cref{app:tab::AllSky_baseline_SPADE_cohesion,app:tab::AllSky_baseline_conv_cohesion} to determine the impact of decoder architecture selection on cohesion between exposures and further study tone mapping. 
As show in \cref{app:tab::AllSky_baseline_SPADE_cohesion}, tone mapping improves cLDR and FDR performance, though some bias may exist given LPIPS' conditioning on $T_\gamma$ LDR imagery.
The results show SPADE+ decoder architecture with $T_\gamma$ tone mapping comparatively offers better performance and no cohesion between exposures is necessary at the convolution layer. 
This reflected through comparatively improved performance with SPADE+ block conditioning on all exposures in \cref{app:tab::AllSky_baseline_SPADE_cohesion}, and comparatively lower performance in \cref{app:tab::AllSky_baseline_conv_cohesion} given cohesion between exposures at the convolution layer.

In table \cref{app:tab::AllSky_baseline_discriminators_fusion} we ablate our discriminators to ascertain their conditioning of \ourModel{} with SPADE+, finding that training with both the LDR and HDR discriminators provides a performance advantage with larger exposure ranges. 
The evaluation of these models was repeated with various fusion operators, demonstrating Robertson fusion ($f_{\text{\tiny Robertson}}$) significantly improves visual quality, while \rgb{} ($f_{\text{\tiny \rgb}}$) and \hsv{} ($f_{\text{\tiny \hsv}}$) fusion offer improved illumination.
Debevec fusion ($f_{\text{\tiny Debevec}}$) is shown to underperform across both visual quality and illumination metrics. 

\begin{table*}[htbp]
\centering
\caption{
    We ablate \ourModel{} with SPADE, SPADE+ and SPADE$^n$ decoder architectures to ascertain the requirement for cohesion between SPADE features. 
    We train \ourModel{} with $T_\gamma$ and $T_{\mu \log_2}$ tone mapping for exposure set $[2^0,2^{-8},2^{-12}]$ and train \ourModel{} without tone mapping ($T_\varnothing$) for exposures $[2^0,2^{-8},2^{-15}]$.
    Model evaluation is repeated for \rgb- and \hsv-fusion, demonstrating the absence of a gap between \rgb and \hsv-fusion performance which suggests limiting exposure overlap reduces noise in fused imagery.
    The results demonstrate that tone mapping ($T_\gamma$ or $T_{\mu \log_2}$) improves cLDR and FDR performance, with $T_\varnothing$ under-performing for all configurations.
    \textbf{Bold} values indicate category best and \underline{underlined} values indicate overall best. 
    In this regard, the SPADE+ decoder architecture with $T_\gamma$ tone mapping is shown to comparatively offer better cLDR and FDR performance.
}
\label{app:tab::AllSky_baseline_SPADE_cohesion}
\begin{tabular}{r|c|ccc|cccc|}
    \multicolumn{2}{c}{} & \multicolumn{3}{c}{\textbf{$T_{\gamma}$-cLDR}} & \multicolumn{4}{c}{\textbf{HDR}} \\
    \cline{3-9} 
    \multicolumn{1}{c}{} & $T_{m}$ & LPIPS $\downarrow$  & FID $\downarrow$  & MiFID $\downarrow$
    & HDR-VDP3 $\uparrow$ & PSNR$_{\log_2}$ $\uparrow$ & EV $\leftarrow$ & $\oiint_I \leftarrow$  \\ 
    
\hline
Ground Truth $128^2$                        & - & - & - & - & - & - & 7.96 & 1.22 \\
\hdashline 
\ourModel{} (SPADE$^n$; $f_{\text{\tiny \rgb}}$) & $T_{\gamma}$ & \textbf{0.14} & 19.8 & 316 & 8.4 & \textbf{147} & 7.37 & 0.92 \\ %17
\ourModel{} (SPADE+; $f_{\text{\tiny \rgb}}$) & $T_{\gamma}$ & \textbf{0.14} & \underline{\textbf{18.8}} & \underline{\textbf{306}} & \underline{\textbf{8.5}} & \textbf{147} & \textbf{8.03} & 1.05 \\ %18
\ourModel{} (SPADE; $f_{\text{\tiny \rgb}}$) & $T_{\gamma}$ & 0.15 & 23.8 & 351 & 8.4 & 139 & 8.60 & \underline{\textbf{1.25}} \\ %19

\hdashline 
\ourModel{} (SPADE$^n$; $f_{\text{\tiny \rgb}}$) & $T_{\mu \log_2}$  & 0.15 & \textbf{22.0} & \textbf{328} & 8.4 & 145 & 7.75 & 1.05 \\ %20
\ourModel{} (SPADE+; $f_{\text{\tiny \rgb}}$) & $T_{\mu \log_2}$  & \textbf{0.14} & 23.9 & 354 & \underline{\textbf{8.5}} & \textbf{147} & 7.82 & 1.01 \\ %21
\ourModel{} (SPADE; $f_{\text{\tiny \rgb}}$) & $T_{\mu \log_2}$  & \textbf{0.14} & 22.3 & 340 & 8.4 & \textbf{147} & \textbf{7.92} & \textbf{1.07} \\ %22

\hdashline 
\ourModel{} (SPADE$^n$; $f_{\text{\tiny \rgb}}$) & $T_{\varnothing}$ & 0.20 & 71.2 & 790 & 7.8 & \textbf{138} & \textbf{7.73} & \textbf{0.96} \\ %23
\ourModel{} (SPADE+; $f_{\text{\tiny \rgb}}$) & $T_{\varnothing}$ & 0.17 & \textbf{35.1} & \textbf{462} & 7.0 & 116 & 9.54 & 3.74 \\ %24
\ourModel{} (SPADE; $f_{\text{\tiny \rgb}}$) & $T_{\varnothing}$ & \textbf{0.16} & 37.4 & 496 & \textbf{8.1} & 127 & 8.91 & 1.54 \\ %25

\midrule 
% HSV

\ourModel{} (SPADE$^n$; $f_{\text{\tiny \hsv}}$) & $T_{\gamma}$ & \textbf{0.14} & 21.3 & 345 & 8.3 & \textbf{147} & 7.38 & 0.95 \\ %17
\ourModel{} (SPADE+; $f_{\text{\tiny \hsv}}$) & $T_{\gamma}$ & \textbf{0.14} & \textbf{18.9} & \textbf{312} & \underline{\textbf{8.5}} & \textbf{147} & \underline{\textbf{7.95}} & 1.06 \\ %18
\ourModel{} (SPADE; $f_{\text{\tiny \hsv}}$) & $T_{\gamma}$ & 0.15 & 22.3 & 355 & 8.4 & 140 & 8.52 & \textbf{1.27} \\ %19

\hdashline 
\ourModel{} (SPADE$^n$; $f_{\text{\tiny \hsv}}$) & $T_{\mu \log_2}$  & 0.15 & \textbf{19.1} & \underline{\textbf{306}} & 8.3 & 146 & 7.61 & \textbf{1.05} \\ %20
\ourModel{} (SPADE+; $f_{\text{\tiny \hsv}}$) & $T_{\mu \log_2}$  & \textbf{0.14} & 21.9 & 345 & \textbf{8.4} & 147 & \textbf{7.76} & 1.02 \\ %21
\ourModel{} (SPADE; $f_{\text{\tiny \hsv}}$) & $T_{\mu \log_2}$  & \textbf{0.14} & 24.4 & 381 & 8.3 & \textbf{148} & 7.65 & 1.04 \\ %22

\hdashline 
\ourModel{} (SPADE$^n$; $f_{\text{\tiny \hsv}}$) & $T_{\varnothing}$ & 0.20 & 70.2 & 872 & \textbf{7.8} & \textbf{140} & \textbf{8.01} & \textbf{1.08} \\ %23
\ourModel{} (SPADE+; $f_{\text{\tiny \hsv}}$) & $T_{\varnothing}$ & 0.17 & \textbf{25.0} & \textbf{380} & 6.7 & 110 & 9.44 & 4.19 \\ %24
\ourModel{} (SPADE; $f_{\text{\tiny \hsv}}$) & $T_{\varnothing}$ & \textbf{0.15} & 33.4 & 491 & 7.6 & 127 & 9.09 & 2.01 \\ %25

\bottomrule
\end{tabular}
\end{table*}

\begin{table*}[htbp]
\centering
\caption{
    We ablate \ourModel{} decoder architectures with a singular output convolution layer (denoted by \textcolor{red}{\textbf{$\dagger$}}) instead of parallelized per-exposure output convolution layers as done in \cref{app:tab::AllSky_baseline_SPADE_cohesion}. 
    \ourModel{} is trained for SPADE, SPADE+ and SPADE$^n$ architecture and exposures $[2^0,2^{-8},2^{-12}]$ with $T_\gamma$ and $T_{\mu \log_2}$ tone mapping. 
    Model evaluation is repeated for \rgb- and \hsv-fusion, with
    \textbf{Bold} values indicating category best and \underline{underlined} values indicating overall best. 
    The results demonstrate that $T_\gamma$ tone mapping improves SPADE performance, while $T_{\mu \log_2}$ tone mapping improves SPADE+ performance. 
    The SPADE and SPADE+ decoder architecture differ only in the number of generated latent features (n-features vs. n-features-per-exposure) and offer similar performance improvement in comparison to SPADE$^n$. 
    Given overall cLDR and FDR metrics, SPADE offers better performance over SPADE+, demonstrating increasing latent features does not correlate to improved visual quality.
    In comparison to \cref{app:tab::AllSky_baseline_SPADE_cohesion}, the results demonstrate that parallelized per-exposure output convolution layers offer improved performance.
}
\label{app:tab::AllSky_baseline_conv_cohesion}
\begin{tabular}{r|c|ccc|cccc|}
    \multicolumn{2}{c}{} & \multicolumn{3}{c}{\textbf{$T_{\gamma}$-cLDR}} & \multicolumn{4}{c}{\textbf{HDR}} \\
    \cline{3-9} 
    \multicolumn{1}{c}{} & $T_{m}$ & LPIPS $\downarrow$  & FID $\downarrow$  & MiFID $\downarrow$
    & HDR-VDP3 $\uparrow$ & PSNR$_{\log_2}$ $\uparrow$ & EV $\leftarrow$ & $\oiint_I \leftarrow$  \\ 
    
\hline
Ground Truth $128^2$                        & - & - & - & - & - & - & 7.96 & 1.22 \\

\hdashline 
\ourModel{} (SPADE$^n$; $f_{\text{\tiny \rgb}}$)\textcolor{red}{\textbf{$^\dagger$}} & $T_{\gamma}$ & \textbf{0.14} & 25.5 & 368 & \underline{\textbf{8.5}} & \textbf{149} & 7.88 & 1.07 \\ %17
\ourModel{} (SPADE+; $f_{\text{\tiny \rgb}}$)\textcolor{red}{\textbf{$^\dagger$}} & $T_{\gamma}$ & 0.15 & 22.6 & 343 & \underline{\textbf{8.5}} & 148 & \underline{\textbf{7.93}} & \underline{\textbf{1.16}} \\ %18
\ourModel{} (SPADE; $f_{\text{\tiny \rgb}}$)\textcolor{red}{\textbf{$^\dagger$}} & $T_{\gamma}$ & \textbf{0.14} & \underline{\textbf{19.5}} & \underline{\textbf{311}} & \underline{\textbf{8.5}} & \textbf{149} & \underline{\textbf{7.93}} & 1.09 \\ %19

\hdashline 
\ourModel{} (SPADE$^n$; $f_{\text{\tiny \rgb}}$)\textcolor{red}{\textbf{$^\dagger$}} & $T_{\mu \log_2}$  & 0.15 & 25.3 & 384 & \underline{\textbf{8.5}} & \underline{\textbf{150}} & \textbf{7.58} & 1.00 \\ %20
\ourModel{} (SPADE+; $f_{\text{\tiny \rgb}}$)\textcolor{red}{\textbf{$^\dagger$}} & $T_{\mu \log_2}$  & \textbf{0.14} & \textbf{20.1} & \textbf{317} & 8.4 & 148 & 7.52 & \textbf{1.03} \\ %21
\ourModel{} (SPADE; $f_{\text{\tiny \rgb}}$)\textcolor{red}{\textbf{$^\dagger$}} & $T_{\mu \log_2}$  & 0.15 & 22.2 & 330 & 8.3 & 149 & 7.23 & 0.89 \\ %22

\midrule 
% HSV

% \hdashline 
\ourModel{} (SPADE$^n$; $f_{\text{\tiny \hsv}}$)\textcolor{red}{\textbf{$^\dagger$}} & $T_{\gamma}$  & \textbf{0.14} & 21.7 & 345 & \underline{\textbf{8.5}} & \textbf{148} & 7.87 & 1.07 \\ %17
\ourModel{} (SPADE+; $f_{\text{\tiny \hsv}}$)\textcolor{red}{\textbf{$^\dagger$}} & $T_{\gamma}$  & 0.15 & 21.7 & 345 & 8.4 & 147 & \textbf{7.92} & \textbf{1.17} \\ %18
\ourModel{} (SPADE; $f_{\text{\tiny \hsv}}$)\textcolor{red}{\textbf{$^\dagger$}} & $T_{\gamma}$  & \textbf{0.14} & \textbf{20.9} & \textbf{342} & 8.4 & \textbf{148} & 8.03 & 1.14 \\ %19

\hdashline 
\ourModel{} (SPADE$^n$; $f_{\text{\tiny \hsv}}$)\textcolor{red}{\textbf{$^\dagger$}} & $T_{\mu \log_2}$   & 0.15 & 25.4 & 400 & \textbf{8.4} & \textbf{149} & \textbf{7.64} & 1.02 \\ %20
\ourModel{} (SPADE+; $f_{\text{\tiny \hsv}}$)\textcolor{red}{\textbf{$^\dagger$}} & $T_{\mu \log_2}$   & \textbf{0.14} & \textbf{19.7} & \textbf{326} & 8.3 & 147 & 7.55 & \textbf{1.04} \\ %21
\ourModel{} (SPADE; $f_{\text{\tiny \hsv}}$)\textcolor{red}{\textbf{$^\dagger$}} & $T_{\mu \log_2}$   & 0.15 & 20.7 & 331 & 8.2 & \textbf{149} & 7.26 & 0.90 \\ %22

\bottomrule
\end{tabular}
\end{table*}

\begin{table*}[htb]
\centering
\caption{
    We ablate our adversarial LDR and HDR discriminators to ascertain their conditioning of \ourModel{} with SPADE+ for exposures $[2^0,2^{-8},2^{-14}]$ and $T_\gamma$ tone mapping.
    Evaluation is repeated for \rgb-, \hsv-, Robertson- (ours) and Debevec-fusion \cite{merge_Debevec}. 
    \textbf{Bold} values indicate category best and \underline{underlined} values indicate overall best. 
    Fusion operators offer contrasting performance, with Roberson-fusion significantly improving visual quality (cLDR) metrics and \rgb- and \hsv-fusion offer improved illumination (FDR metrics). As shown, Debevec-fusion underperforms in terms of both visual quality and illumination metrics.
    At a resolution of $128^2$, our HDR discriminator outperforms our LDR discriminator, and no advantage is gained from training with both (LDR+HDR) discriminators. 
    When resolution is doubled to $256^2$, training with both (LDR+HDR) discriminators provides a clear advantage, offering better handling of the (linear) doubling of the exposure range. 
}
\label{app:tab::AllSky_baseline_discriminators_fusion}
\begin{tabular}{r|c|ccc|cccc|}
    \multicolumn{2}{c}{} & \multicolumn{3}{c}{\textbf{$T_{\gamma}$-cLDR}} & \multicolumn{4}{c}{\textbf{HDR}} \\
    \cline{3-9} 
    \multicolumn{1}{c}{} & $T_{m}$ & LPIPS $\downarrow$  & FID $\downarrow$  & MiFID $\downarrow$
    & HDR-VDP3 $\uparrow$ & PSNR$_{\log_2}$ $\uparrow$ & EV $\leftarrow$ & $\oiint_I \leftarrow$ \\ 
    
\hline
Ground Truth $128^2$                        & - & - & - & - & - & - & 7.96 & 1.22 \\
\hdashline 
\ourModel{} (LDR; $f_{\text{\tiny \rgb}}$) & $T_{\gamma}$ & 0.14 & 26.0 & 388 & 8.54 & 149 & \textbf{7.91} & 1.06 \\
\ourModel{} (HDR; $f_{\text{\tiny \rgb}}$) & $T_{\gamma}$ & 0.14 & \textbf{23.9} & \textbf{357} & \underline{\textbf{8.57}} & \underline{\textbf{150}} & 7.79 & \textbf{1.08} \\
% **31 \ourModel{} (HDR; $f_{\text{\tiny \rgb}}$) & $T_{\mu \log_2}|T_{\gamma}$ & 0.14 & 22.4 & 337 & 8.48 & 148 & 7.93 & 1.10 \\
\ourModel{} (LDR+HDR; $f_{\text{\tiny \rgb}}$) & $T_{\gamma}$ & 0.14 & \textbf{23.9} & 362 & 8.46 & 147 & 7.85 & 0.98 \\

\hdashline
\ourModel{} (LDR; $f_{\text{\tiny \hsv}}$) & $T_{\gamma}$ & 0.14 & 24.6 & 390 & \textbf{8.54} & 149 & \underline{\textbf{7.98}} & \underline{\textbf{1.10}} \\
\ourModel{} (HDR; $f_{\text{\tiny \hsv}}$) & $T_{\gamma}$ & 0.14 & \textbf{20.5} & \textbf{333} & \textbf{8.54} & \underline{\textbf{150}} & 7.76 & \underline{\textbf{1.10}} \\
\ourModel{} (LDR+HDR; $f_{\text{\tiny \hsv}}$) & $T_{\gamma}$ & 0.14 & 25.6 & 405 & 8.38 & 148 & 7.82 & 0.95 \\

\hdashline
\ourModel{} (LDR; $f_{\text{\tiny Robertson}}$) & $T_{\gamma}$ & \underline{\textbf{0.13}} & 16.3 & 269 & 7.41 & 146 & \textbf{7.82} & \textbf{1.05} \\
\ourModel{} (HDR; $f_{\text{\tiny Robertson}}$) & $T_{\gamma}$ & \underline{\textbf{0.13}} & \underline{\textbf{15.0}} & \underline{\textbf{250}} & 6.42 & 144 & 7.63 & 1.04 \\
\ourModel{} (LDR+HDR; $f_{\text{\tiny Robertson}}$) & $T_{\gamma}$ & \underline{\textbf{0.13}} & 16.9 & 284 & \textbf{7.77} & \textbf{148} & 7.64 & 0.97 \\

\hdashline
\ourModel{} (LDR; $f_{\text{\tiny Debevec}}$) & $T_{\gamma}$     & 0.26 & \textbf{51.0} & \textbf{554} & \textbf{3.44} & \textbf{116} & \textbf{8.37} & 1.66 \\
\ourModel{} (HDR; $f_{\text{\tiny Debevec}}$) & $T_{\gamma}$     & 0.26 & 53.4 & 584 & \textbf{3.44} & 115 & 8.54 & 1.70 \\
\ourModel{} (LDR+HDR; $f_{\text{\tiny Debevec}}$) & $T_{\gamma}$ & 0.26 & 55.7 & 625 & 3.43 & \textbf{116} & 8.39 & \textbf{1.60} \\

\midrule
Ground Truth $256^2$                        & - & - & - & - & - & - & 9.36 & 1.22 \\
\hdashline 
\ourModel{} (LDR; $f_{\text{\tiny \rgb}}$) & $T_{\gamma}$ & \underline{\textbf{0.17}} & 34.9 & 512 & 8.30 & 168 & 9.54 & 1.05 \\
\ourModel{} (HDR; $f_{\text{\tiny \rgb}}$) & $T_{\gamma}$ & \underline{\textbf{0.17}} & 33.2 & 490 & 8.27 & 168 & \textbf{9.46} & \textbf{1.11} \\
\ourModel{} (LDR+HDR; $f_{\text{\tiny \rgb}}$) & $T_{\gamma}$ & 0.18 & \textbf{31.5} & \textbf{465} & \underline{\textbf{8.33}} & \textbf{170} & 9.59 & 1.08 \\
\hdashline 
\ourModel{} (LDR; 3xHSV) & $T_{\gamma}$  & \underline{\textbf{0.17}} & 38.6 & 606 & 8.24 & 168 & 9.7 & 1.1 \\
\ourModel{} (HDR; 3xHSV) & $T_{\gamma}$  & \underline{\textbf{0.17}} & 34.9 & 537 & 8.23 & 168 & \textbf{9.51} & \underline{\textbf{1.17}} \\
\ourModel{} (LDR+HDR; 3xHSV) & $T_{\gamma}$  & 0.18 & \textbf{32.9} & \textbf{516} & \textbf{8.31} & \textbf{170} & 9.75 & 1.13 \\
\hdashline 
\ourModel{} (LDR; $f_{\text{\tiny Robertson}}$) & $T_{\gamma}$  & \underline{\textbf{0.17}} & 22.5 & 372 & 8.20 & 169 & 9.51 & 1.06 \\
\ourModel{} (HDR; $f_{\text{\tiny Robertson}}$) & $T_{\gamma}$ & \underline{\textbf{0.17}} & 23.0 & 375 & 8.20 & 168 & \underline{\textbf{9.44}} & \textbf{1.10} \\
\ourModel{} (LDR+HDR; $f_{\text{\tiny Robertson}}$) & $T_{\gamma}$  & \underline{\textbf{0.17}} & \underline{\textbf{20.5}} & \underline{\textbf{338}} & \textbf{8.21} & \underline{\textbf{171}} & 9.49 & 1.08 \\
\hdashline 
\ourModel{} (LDR; $f_{\text{\tiny Debevec}}$) & $T_{\gamma}$ & - & - & - & - & - & - & - \\
\ourModel{} (HDR; $f_{\text{\tiny Debevec}}$) & $T_{\gamma}$ & 0.26 & 62.0 & 690 & 3.27 & 137 & 9.67 & 1.49 \\
\ourModel{} (LDR+HDR; $f_{\text{\tiny Debevec}}$) & $T_{\gamma}$ & - & - & - & - & - & - & - \\
\bottomrule
\end{tabular}
\end{table*}

\subsection{\ourModel{}  with \rnd-Style Mapper}
\label{app:X_experiments::AllSky_V2_MGD}

In \cref{app:tab::AllSky_MGD} we evaluate \ourModel{} trained exclusively with our \rnd-style mapper. 
This faster variant is trained is trained with the same subsets of HDRDB defined in \cref{app:dataset::subset_augmentation} and the same parameters, but requires prolonged training to first establish the \rnd-style mapper. 
As such, this variant is trained for 400 epochs with all exposure decoder heads set to $\Delta t_0=1$, then uniformly decayed to target exposures over 400 epochs.

\textcolor{red}{Please note, we could not report results for \ourModel{} with \rnd-style mapper at a resolution of $512^2$ due to cluster failures which prevented the allocation of compute time. These unfortunate failures were outside our control.}

\begin{table*}[htb]
\centering
\caption{
    \ourModel{} \rnd-style mapper evaluation repeated for \rgb, \hsv, Robertson and Debevec fusion given training with $T_\gamma$ tone mapping for exposures $[2^0,2^{-8},2^{-15}]$ using our LDR- and HDR-bracket discriminators ($\check{\mathcal{D}}$ and $\hat{\mathcal{D}}$). 
    \textbf{Bold} values indicate category best and \underline{underlined} values indicate overall best. 
    LPIPS\textcolor{red}{\textbf{$^\dagger$}} is included for continuity but \ourModel{} generates random style codes (\rnd-Style). 
    Poor LPIPS performance paired with good FID, MiFID and HDR-VDP performance is indicative of high visual quality with diverse \rnd-Styles.  
    Closely matched EV, $\oiint_I$ and $PL_\Omega$ demonstrates that despite variability in cloud formations, \ourModel{} sustains global and solar illumination.
}
\label{app:tab::AllSky_MGD}
\begin{tabular}{r|ccc|ccccc|}
    \multicolumn{1}{c}{} & \multicolumn{3}{c}{\textbf{$T_{\gamma}$-cLDR}} & \multicolumn{5}{c}{\textbf{HDR}} \\
    \cline{2-9} 
    \multicolumn{1}{c}{} & LPIPS\textcolor{red}{\textbf{$^\dagger$} $\downarrow$}  & FID $\downarrow$  & MiFID $\downarrow$
    & HDR-VDP3 $\uparrow$ & VIF $\uparrow$ &  EV $\leftarrow$ & $\oiint_I \leftarrow$ & $PL_\Omega \leftarrow$ \\ 
\hline
Ground Truth $256^2$                  & -    & -   & -    & -   & -    & 8.58 & 1.22 & 0.374 \\
\hdashline
\ourModel{} \rnd-style $f_{\text{\tiny \rgb}}$ & 0.194 & 20.1 & 340 & \underline{\textbf{8.21}} & \textbf{1.59} & 9.16 & \underline{\textbf{1.28}} & \underline{\textbf{0.425}} \\
\ourModel{} \rnd-style $f_{\text{\tiny \hsv}}$ & 0.194 & 23.7 & 408 & 8.15 & \textbf{1.62} & 9.23 & 1.31 & 0.444 \\
\ourModel{} \rnd-style $f_{\text{\tiny Robertson}}$ & \underline{\textbf{0.19}}  & \underline{\textbf{11}}   & \underline{\textbf{204}} & 8.16 & \textbf{1.52} & \underline{\textbf{9.08}} & \underline{\textbf{1.28}} & \underline{\textbf{0.425}} \\
\ourModel{} \rnd-style $f_{\text{\tiny DNN}}$ & 0.19 & 14.8 & 264 &	8.03 & 0.97 & 7.50 & 0.8426 & 0.1245  \\
\ourModel{} \rnd-style $f_{\text{\tiny Debevec}}$ & 0.266 & 41.3 & 515 & 3.27 & 2.97 & 9.17 & 1.78 & 0.25  \\

\bottomrule
\end{tabular}
\end{table*}
\clearpage

\begin{figure*}
    \centering
    \includegraphics[width=\linewidth]{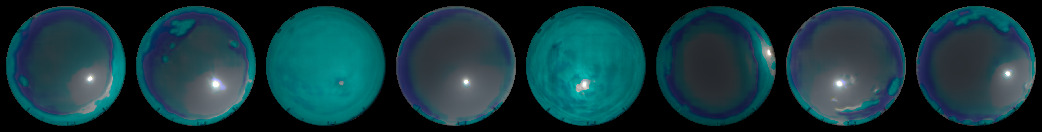}
    \caption{Example of chromatic aberration in LDR bracket fusion}
    \label{app:fig::fusion_chromatic_aberation}
\end{figure*}

\subsection{Fusion}
\label{app:X_experiments::Fusion}

In \cref{app:tab::results_Fusion_128,app:tab::results_Fusion_256} we evaluate the performance of our proposed DNN fusion module ($f_{\text{\tiny DNN}}$) and ablate to determine optimal losses per pre-trained \ourModel{} with \rgb-style encoder.
Initial experimentation demonstrated that predicting separate weights $\left\{W^{\Delta t}_{n}\right\}^N$ and $\left\{W^e_{n}\right\}^N$ for \cref{app:eq::LDR_exposure_fusion_DNN_WtWe} collapsed in training and independent use of Integrated Illumination ($\oiint_I$) and/or Peak Luminance ($PL_\Omega$) as losses is prone to chromatic aberration as illustrated in \cref{app:fig::fusion_chromatic_aberation}.
We therefore only report results for our proposed DNN fusion module ($f_{\text{\tiny DNN}}$) for uniform weights $\left\{W_{n}\right\}^N$ per \cref{app:eq::LDR_exposure_fusion_DNN_W} with adversarial losses.

Results in \cref{app:tab::results_Fusion_128,app:tab::results_Fusion_256} demonstrate that our $f_{\text{\tiny DNN}}$ module is capable of fusion with visual quality similar to $f_{\text{\tiny Robertson}}$, but exhibits difficulty in matching the illumination accuracy of both  $f_{\text{\tiny Robertson}}$ and $f_{\text{\tiny \rgb}}$.
In combination with results at a resolution of $512^2$ in \cref{tab::results}, $f_{\text{\tiny DNN}}$ demonstrates a capacity to model intensity (EV) but with an under-representation of $PL_\Omega$.
This could be a reflection of $f_{\text{\tiny DNN}}$ under-weighting high-exposures to mitigate overexposure, which may be addressable by incorporating Integrated Illumination ($\oiint_I$) and/or Peak Luminance ($PL_\Omega$) to the losses.

\begin{table*}[htb]
\centering
\renewcommand{\arraystretch}{1.3}
\caption{
    Ablation of fusion module $f_{\text{\tiny DNN}}$ adversarial losses, with \ourModel{} trained for exposures $[2^0,2^{-8},2^{-15}]$ and $T_\gamma$ tone mapping.
    Fusion module performance is shown to be closely tied to loss selection, with no discriminator independently offering both visual quality and accurate illumination.
    Combinations of discriminators offer improved performance, with combined $\mathcal{D}+\check{\mathcal{D}}+\hat{\mathcal{D}}$ offering balanced visual quality and accurate illumination.
    Performance with latent inputs ($\xi$) closely matches performance with \rgb{} inputs.
    \textbf{Bold} values indicate category best and \underline{underlined} values indicate overall best.
}
\label{app:tab::results_Fusion_128}
\begin{tabular}{r|c|cc|ccccc|}
    \multicolumn{2}{c}{} & \multicolumn{2}{c}{\textbf{$T_{\gamma}$-cLDR}} & \multicolumn{5}{c}{\textbf{HDR}} \\
    \cline{3-9}
    \multicolumn{1}{c}{} & Loss ($\mathcal{L}$) & FID $\downarrow$  & MiFID $\downarrow$
    & HDR-VDP3 $\uparrow$ & VIF $\uparrow$ & EV $\leftarrow$ &  $PL_\Omega \leftarrow$ & $\oiint_I \leftarrow$ \\

\hline
Ground Truth $128^2$                        & - & - & - & - & - & 7.34 & 0.451 & 1.22 \\

% RGB Input
\midrule
$f_{\text{\tiny \rgb}}$      & - & 23.7 & 359 & \underline{\textbf{8.52}} & \underline{\textbf{1.07}} & 7.64 & 0.373 & \underline{\textbf{1.16}} \\ % 32
\cdashline{2-9}
$f_{\text{\tiny Robertson}}$ & - & \underline{\textbf{15.7}} & \underline{\textbf{264}} & 8.41 & \textbf{0.894} & \textbf{7.49} & \underline{\textbf{0.368}} & 1.13 \\ % 32
\hdashline
$f_{\text{\tiny DNN}}\left(\left\{\Check{I}_n\right\}^N, L\right)$ & $\mathcal{D}+\check{\mathcal{D}}+\hat{\mathcal{D}}$ & 20.4 & 317 & \textbf{8.4} & \textbf{0.56} & \underline{\textbf{7.26}} & \textbf{0.283} & \textbf{1.04} \\ %5
$f_{\text{\tiny DNN}}\left(\left\{\Check{I}_n\right\}^N, L\right)$ & $\check{\mathcal{D}}+\hat{\mathcal{D}}$ & \textbf{16.4} & \textbf{269} & 7.42 & 0.227 & 0.94 & 0.0006 & 0.65 \\ %6
$f_{\text{\tiny DNN}}\left(\left\{\Check{I}_n\right\}^N, L\right)$ & $\mathcal{D}+\hat{\mathcal{D}}$ & 17 & 278 & \textbf{8.4} & 0.468 & \textbf{7.21} & 0.268 & 1.01 \\ %7
$f_{\text{\tiny DNN}}\left(\left\{\Check{I}_n\right\}^N, L\right)$ & $\mathcal{D}+\check{\mathcal{D}}$ & 16.7 & 273 & 8.23 & 0.511 & 6.33 & 0.156 & 0.83 \\ %8
% \hdashline
\cdashline{2-9}
$f_{\text{\tiny DNN}}\left(\left\{\Check{I}_n\right\}^N, L\right)$ & $\mathcal{D}$ & 16.9 & 278 & \textbf{8.29} & \textbf{0.747} & \textbf{6.75} & \textbf{0.163} & \textbf{0.86} \\ %9
$f_{\text{\tiny DNN}}\left(\left\{\Check{I}_n\right\}^N, L\right)$ & $\check{\mathcal{D}}$ & \textbf{15.8} & \textbf{265} & 8.02 & 0.474 & 4.47 & 0.025 & 0.65 \\ %10
$f_{\text{\tiny DNN}}\left(\left\{\Check{I}_n\right\}^N, L\right)$ & $\hat{\mathcal{D}}$ & 21.1 & 326 & 7.41 & 0.204 & 1.02 & 0.0007 & 0.61 \\ %11

% Latent Input
\hdashline
$f_{\text{\tiny DNN}}\left(\left\{\Check{I}_n\right\}^N,{\xi},L\right)$ & $\mathcal{D}+\check{\mathcal{D}}+\hat{\mathcal{D}}$ & 19.7 & 310 & \textbf{8.4} & 0.428 & \textbf{7.15} & 0.267 & \textbf{1.01} \\ %12
$f_{\text{\tiny DNN}}\left(\left\{\Check{I}_n\right\}^N,{\xi},L\right)$ & $\check{\mathcal{D}}+\hat{\mathcal{D}}$ & \underline{\textbf{15.7}} & \underline{\textbf{264}} & 7.41 & 0.202 & 0.95 & 0.0005 & 0.61 \\ %13
$f_{\text{\tiny DNN}}\left(\left\{\Check{I}_n\right\}^N,{\xi},L\right)$ & $\mathcal{D}+\hat{\mathcal{D}}$ & 21.1 & 319 & 8.21 & \textbf{0.679} & 6.14 & 0.111 & 0.78 \\ %14
$f_{\text{\tiny DNN}}\left(\left\{\Check{I}_n\right\}^N,{\xi},L\right)$ & $\mathcal{D}+\check{\mathcal{D}}$ & 18.8 & 298 & \textbf{8.39} & 0.461 & 7.14 & \textbf{0.268} & \textbf{1.01} \\ %15
% \hdashline
\cdashline{2-9}
$f_{\text{\tiny DNN}}\left(\left\{\Check{I}_n\right\}^N,{\xi},L\right)$ & $\mathcal{D}$ & 24.4 & 365 & \textbf{8.41} & \textbf{0.753} & \textbf{7.08} & \textbf{0.284} & \textbf{1.04} \\ %16
$f_{\text{\tiny DNN}}\left(\left\{\Check{I}_n\right\}^N,{\xi},L\right)$ & $\check{\mathcal{D}}$ & \textbf{22.7} & \textbf{334} & 8 & 0.468 & 5.92 & 0.088 & 0.75 \\ %17
$f_{\text{\tiny DNN}}\left(\left\{\Check{I}_n\right\}^N,{\xi},L\right)$ & $\hat{\mathcal{D}}$ & 24.1 & 340 & 8.11 & 0.297 & 5.46 & 0.148 & 0.81 \\ %18
\bottomrule
\end{tabular}
\end{table*}

\begin{table*}[htb]
\centering
\renewcommand{\arraystretch}{1.3}
\caption{
    Ablation of fusion module $f_{\text{\tiny DNN}}$ adversarial losses, with \ourModel{} trained for exposures $[2^0,2^{-8},2^{-15}]$ and $T_\gamma$ tone mapping.
    Fusion module performance is shown to be closely tied to loss selection, with no singular discriminator or combination of discriminators offering both visual quality and accurate illumination.
    Performance with latent inputs ($\xi$) closely matches performance with \rgb{} inputs.
    \textbf{Bold} values indicate category best and \underline{underlined} values indicate overall best.
}
\label{app:tab::results_Fusion_256}
\begin{tabular}{r|c|cc|ccccc|}
    \multicolumn{2}{c}{} & \multicolumn{2}{c}{\textbf{$T_{\gamma}$-cLDR}} & \multicolumn{5}{c}{\textbf{HDR}} \\
    \cline{3-9}
    \multicolumn{1}{c}{} & Loss ($\mathcal{L}$) & FID $\downarrow$  & MiFID $\downarrow$
    & HDR-VDP3 $\uparrow$ & VIF $\uparrow$ & EV $\leftarrow$ &  $PL_\Omega \leftarrow$ & $\oiint_I \leftarrow$ \\

\hline
Ground Truth $256^2$                        & - & - & - & - & - & 8.58 & 0.367 & 1.22 \\
\midrule
$f_{\text{\tiny \rgb}}$     & - & 31.5 & 465 & \underline{\textbf{8.33}} & \textbf{0.894} & 8.96 & \underline{\textbf{0.315}} & \underline{\textbf{1.08}} \\ %36
\cdashline{2-9}
$f_{\text{\tiny Robertson}}$ & - & \textbf{20.5} & \textbf{338} & 8.21 & 0.717 & \textbf{8.81} & 0.314 & \underline{\textbf{1.08}} \\ %36

\hdashline
$f_{\text{\tiny DNN}}\left(\left\{\Check{I}_n\right\}^N, L\right)$ & $\mathcal{D}+\check{\mathcal{D}}+\hat{\mathcal{D}}$ & \textbf{20.4} & \textbf{336} & 7.98 & 0.239 & 5.82 & 0.018 & 0.66 \\ %5

$f_{\text{\tiny DNN}}\left(\left\{\Check{I}_n\right\}^N, L\right)$ & $\check{\mathcal{D}}+\hat{\mathcal{D}}$ & 20.5 & 337 & \textbf{8.24} & 0.304 & 8.31 & \textbf{0.221} & \textbf{0.94}  \\ %6

$f_{\text{\tiny DNN}}\left(\left\{\Check{I}_n\right\}^N, L\right)$ & $\mathcal{D}+\hat{\mathcal{D}}$ & 20.9 & 342 & 8.03 & 0.340 & 7.16 & 0.072 & 0.71  \\ %7
$f_{\text{\tiny DNN}}\left(\left\{\Check{I}_n\right\}^N, L\right)$ & $\mathcal{D}+\check{\mathcal{D}}$ & 20.5 & 337 & \textbf{8.24} & \textbf{0.404} & \textbf{8.43} & \textbf{0.221} & \textbf{0.94}  \\ %8

\cdashline{2-9}
$f_{\text{\tiny DNN}}\left(\left\{\Check{I}_n\right\}^N, L\right)$ & $\mathcal{D}$ & 20.6 & 338 & \textbf{8.24} & \textbf{0.409} & 8.39 & \textbf{0.221} & \textbf{0.94} \\ %9
$f_{\text{\tiny DNN}}\left(\left\{\Check{I}_n\right\}^N, L\right)$ & $\check{\mathcal{D}}$ & \textbf{20.2} & \textbf{333} & \textbf{8.24} & 0.397 & \underline{\textbf{8.45}} & \textbf{0.221} & \textbf{0.94} \\ %10
$f_{\text{\tiny DNN}}\left(\left\{\Check{I}_n\right\}^N, L\right)$ & $\hat{\mathcal{D}}$ & 20.5 & 337 & 7.43 & 0.168 & 0.89 & 0.0001 & 0.59 \\ %11

% Latent Input
\hdashline
$f_{\text{\tiny DNN}}\left(\left\{\Check{I}_n\right\}^N,{\xi},L\right)$ & $\mathcal{D}+\check{\mathcal{D}}+\hat{\mathcal{D}}$ & \textbf{20.7} & \textbf{337} & 8.05 & 0.373 & 7.10 & 0.067 & 0.71 \\ %12

$f_{\text{\tiny DNN}}\left(\left\{\Check{I}_n\right\}^N,{\xi},L\right)$ & $\check{\mathcal{D}}+\hat{\mathcal{D}}$ & 22.5 & 357 & \textbf{8.23} & 0.326 & \textbf{8.25} & \textbf{0.222} & \textbf{0.95} \\ %13

$f_{\text{\tiny DNN}}\left(\left\{\Check{I}_n\right\}^N,{\xi},L\right)$ & $\mathcal{D}+\hat{\mathcal{D}}$ & 25.6 & 407 & 8.05 & \textbf{0.413} & 6.39 & 0.038 & 0.65 \\ %14
$f_{\text{\tiny DNN}}\left(\left\{\Check{I}_n\right\}^N,{\xi},L\right)$ & $\mathcal{D}+\check{\mathcal{D}}$ & \textbf{20.7} & 340 & 8.22 & 0.341 & 8.11 & 0.191 & 0.89 \\ %15

\cdashline{2-9}
$f_{\text{\tiny DNN}}\left(\left\{\Check{I}_n\right\}^N,{\xi},L\right)$ & $\mathcal{D}$ & 27.0 & 435 & 7.97 & \underline{\textbf{2.660}} & 7.53 & 0.108 & 0.77 \\ %16
$f_{\text{\tiny DNN}}\left(\left\{\Check{I}_n\right\}^N,{\xi},L\right)$ & $\check{\mathcal{D}}$ & 20.6 & 337 & \textbf{8.17} & 0.463 & \textbf{8.13} & \textbf{0.155} & \textbf{0.85} \\ %17

$f_{\text{\tiny DNN}}\left(\left\{\Check{I}_n\right\}^N,{\xi},L\right)$ & $\hat{\mathcal{D}}$ & \underline{\textbf{19.8}} & \underline{\textbf{326}} & 7.57 & 0.175 & 1.11 & 0.0002 & 0.60 \\  %18

\bottomrule
\end{tabular}
\end{table*}

%%%%%%%%% END DOCUMENT %%%%%%%%%
\end{document}